\documentclass[manuscript,screen,acmsmall]{acmart}   
\renewcommand\footnotetextcopyrightpermission[1]{} 
\setcopyright{none}
\AtBeginDocument{%
  }

\acmJournal{CSUR}

\usepackage{hyperref}
\usepackage{amsmath}
\usepackage{amsfonts}
\usepackage{graphicx}
\usepackage{nicematrix}
\usepackage{array}
\usepackage{colortbl}
\usepackage{tikz}
\usepackage{float}
\usepackage{comment}
\usepackage{adjustbox}
\usepackage{caption}  
\usepackage{subcaption}
\usepackage{lscape}
\usepackage{longtable}
\usepackage{pgfplots}
\pgfplotsset{compat=1.18}
\usepackage{tabularx}
\usepackage{soul}
\usepackage{multirow} 

\usepackage{amssymb} 

\newcommand{\colorred}[1]{{#1}}
\newcommand{\raju}[1]{\textcolor{black}{#1}}

\definecolor{tabheadcolor}{HTML}{AED6F1} 
\definecolor{tabsubheadcolor}{HTML}{D9EAF7}
\definecolor{fdcolor}{HTML}{E85642}
\definecolor{csacolor}{HTML}{F08705}
\definecolor{ppcolor}{HTML}{736CA8}
\definecolor{pspcolor}{HTML}{12AAB5}
\definecolor{taxlabelcolor}{HTML}{2F5B7C}
\begin{document}


\title{Synthetic Tabular Data Generation: A Comparative Survey for Modern Techniques}

\author{Raju Challagundla}
\email{rchalla5@charlotte.edu}
\orcid{0009-0005-2510-5209}
\affiliation{%
  \institution{University of North Carolina at Charlotte}
  \city{Charlotte}
  \state{North Carolina}
  \country{USA}
}
\author{Mohsen Dorodchi}
\email{Mohsen.Dorodchi@charlotte.edu}
\orcid{0000-0001-7522-1068}
\affiliation{
  \institution{University of North Carolina at Charlotte}
  \city{Charlotte}
  \state{North Carolina}
  \country{USA}
}
\author{Pu Wang}
\email{pu.wang@charlotte.edu}
\orcid{0000-0003-1988-5016}
\affiliation{
  \institution{University of North Carolina at Charlotte}
  \city{Charlotte}
  \state{North Carolina}
  \country{USA}
}
\author{Minwoo Lee}
\email{Minwoo.Lee@charlotte.edu}
\orcid{0000-0002-6860-608X}
\affiliation{
  \institution{University of North Carolina at Charlotte}
  \city{Charlotte}
  \state{North Carolina}
  \country{USA}
}

\begin{abstract}
  As privacy regulations become more stringent and access to real-world data becomes increasingly constrained, synthetic data generation has emerged as a vital solution, especially for tabular datasets, which are central to domains like finance, healthcare and the social sciences. 
  This survey presents a \colorred{focused comparative review} of recent advances in synthetic tabular data generation, emphasizing methods that preserve complex feature relationships, maintain statistical fidelity, and satisfy privacy requirements. A key contribution of this work is the introduction of a novel taxonomy based on practical generation objectives, including intended downstream applications, privacy guarantees, and data utility, directly informing methodological design and evaluation strategies. 
  Therefore, this review prioritizes the actionable goals that drive synthetic data creation, including conditional generation and risk-sensitive modeling. Additionally, the survey proposes a benchmark framework to align technical innovation with real-world demands. 
  By bridging theoretical foundations with practical deployment, this work serves as both a roadmap for future research and a guide for implementing synthetic tabular data in privacy-critical environments.

\end{abstract}

  
\maketitle

\section{Introduction}

The increasing reliance on data-driven technologies across sectors such as finance \cite{tian2021data}, healthcare \cite{rahman2023data}, marketing \cite{arthur2013big}, and social sciences \cite{zhang2020data} highlights the critical role of large, representative datasets in fostering innovation. For instance, big data analytics in finance has transformed investment strategies \cite{vishweswar2024big}, while large datasets in healthcare have advanced personalized medicine and predictive analytics \cite{clim2019big}. 
 However, the collection, storage, and sharing of real-world data face significant challenges, including privacy, regulatory compliance, data scarcity, and logistical difficulties \cite{zhao2019logistics}. These challenges have led to the development of synthetic data generation techniques, which create artificial datasets that replicate the properties of real-world data while protecting sensitive information.
 
The demand for synthetic data \cite{lu2023machine} arises from the limited accessibility of high-quality datasets, particularly due to strict privacy regulations like General Data Protection Regulation (GDPR) \cite{european_union_2016_gdpr} and The Health Insurance Portability and Accountability Act (HIPAA) \cite{HIPAA1996}. Synthetic data enables the creation of anonymized datasets, facilitating collaboration and promoting data democratization \cite{Amplitude2024,lu2023machine}. Additionally, it addresses issues of data imbalance in fields like fraud detection and medical research, enhancing model accuracy and generalizability \cite{he2009learning} 
Synthetic data generation also offers a cost-effective alternative to real-world data collection \cite{babbar2019data}, which is often expensive and time-consuming. Once trained, models can produce vast amounts of data at a fraction of the cost, making them scalable for organizations with limited resources. 

\colorred{When an organization enters a new market, limited target-domain data can delay the initial development of recommender or fraud-detection systems. Synthetic augmentation can support cold-start recommendation and learning from underrepresented classes \cite{wang2024large,assefa2020generating,d2025synthetic}, but its benefit must be evaluated on held-out real data. Because a generator trained only on a source population cannot determine the target population's distribution, expected market differences require target-domain evidence or explicit scenario assumptions, followed by evaluation under distribution shift \cite{gardner2023benchmarking}.}

In the past decade, synthetic data generation has achieved remarkable success across diverse domains, despite several challenges. Capturing complex data distributions, preserving privacy, maintaining feature dependencies, and balancing realism with diversity have all posed significant hurdles. Recent \raju{advances} in generative models, such as Generative Adversarial Networks (GANs) \cite{goodfellow2014generative} and Variational Autoencoders (VAEs) \cite{kingma2013auto}, have been pivotal in overcoming these hurdles, enabling the creation of synthetic data that better mirrors real-world datasets while addressing privacy concerns and enhancing utility. GANs excel in creating high-quality images and text, while VAEs are particularly suited for probabilistic data generation. These innovations have enabled transformative applications across several industries. In finance, synthetic data is leveraged for fraud detection \cite{assefa2020generating}, simulating market conditions, and testing trading algorithms, all without financial risk. In computer vision \cite{karras2019style}, GANs \cite{goodfellow2014generative} generate photorealistic images for data augmentation. In healthcare \cite{hernandez2022synthetic}, synthetic patient records support research in privacy-sensitive scenarios with limited real-world data. Similarly, synthetic data plays a crucial role in training autonomous vehicles \cite{dosovitskiy2017carla} by simulating diverse driving environments, ensuring robust and safe performance.

While synthetic data generation techniques have been successful for various data types such as images, text, audio, video and time series, tabular data presents unique challenges and opportunities that warrant focused attention. Unlike unstructured data (e.g., images or text), tabular data is characterized by heterogeneous feature types (e.g., numerical, categorical, and ordinal), complex inter-feature dependencies, and domain-specific constraints. Particularly in financial domains \cite{assefa2020generating}, for instance, institutions must balance data utility and privacy when developing models for CreditRisk assessment \cite{credit_risk_kaggle}, fraud detection, and loan approvals. Sharing real transaction data can lead to privacy breaches and regulatory violations, thus making synthetic data a viable alternative. However, generating synthetic tabular data requires preserving intricate feature relationships (e.g., correlations between income, credit score, and loan repayment behavior) while ensuring privacy and statistical fidelity. These challenges are less pronounced in other data types, where the focus is often on preserving structural patterns (e.g., pixel relationships in images or semantic coherence in text). 

 \colorred{Prior surveys provide several complementary views of synthetic data generation. Figueira and Vaz \cite{figueira2022survey} review generation and evaluation methods with substantial attention to GAN-based approaches, while Lu et al. \cite{lu2023machine} organize the field from a machine-learning perspective. Bauer et al. \cite{bauer2024comprehensive} develop a multidimensional taxonomy spanning generation goals, data types, and privacy considerations. For tabular data specifically, Fonseca and Bacao \cite{fonseca2023tabular} classify 70 algorithms according to six generation mechanisms and their associated machine-learning purposes, whereas Lautrup et al. \cite{lautrup2024systematic} systematically review fully synthetic tabular generators and utility metrics. More recent work by Cormode et al. \cite{cormode2025synthetic} jointly examines tabular generation methods, privacy attacks, and defenses. Shi et al. \cite{shi2025comprehensive} provide a recent preprint organized around traditional, diffusion-based, and LLM-based methods, including post-processing and evaluation. Alismail and Lanquillon \cite{alismail2025survey} review LLM-based synthetic-data generation and contrast single-model approaches with agentic workflows across data modalities.}
\colorred{Unlike prior surveys that focus primarily on diverse model architectures, the present survey examines synthetic data generation with respect to operational objectives such as maintaining feature dependencies, preserving statistical properties, protecting privacy, and conditioning generation on specified attributes. For each objective, we compare relevant mechanisms, documented limitations, and evaluation criteria; consequently, some methods are discussed across multiple objectives. The review concentrates on representative approaches that provide explicit mechanisms or empirical evidence relevant to these objectives, rather than attempting a comprehensive enumeration of all available generators. The paper also complements the survey with a benchmarking framework that empirically assesses representative methods under each operational objective on the Adult and Credit Risk datasets. }

The survey is structured as follows. Section~\ref{sec:SDG} defines synthetic data generation, providing essential terminology and context.  
Section ~\ref{sec:TSDG} focuses on tabular synthetic data generation, categorizing methods based on key characteristics such as feature dependency, statistical preservation, preserving privacy, conditioning on specific attributes and leveraging specific domain knowledge. Section ~\ref{sec:BAC} presents benchmarking studies that compare different synthetic data generation models across datasets, highlighting their performance with respect to utility, privacy, and statistical fidelity. Section ~\ref{sec:CL} highlights challenges and limitations. Finally, the Appendices provide foundational background, including definitions, data types, \colorred{taxonomy-related tables, and detailed descriptions} of evaluation metrics, \colorred{all of which} support and complement the discussions in the main sections.

\section{Synthetic Data Generation}\label{sec:SDG}

Synthetic data generation refers to the process of creating artificial data that retains the statistical properties and patterns of real-world datasets \cite{figueira2022survey}. 
Let \( \mathbf{X} \) represent the original data, where \( \mathbf{X} \) follows a probability distribution \( P(\mathbf{X}) \). The goal of synthetic data generation is to estimate \( P(\mathbf{X}) \) and sample new data from the estimated distribution: 
\[
\mathbf{X}_{\text{syn}} \sim \hat{P}(\mathbf{X}),
\]
where \( \hat{P}(\mathbf{X}) \) is obtained through statistical estimation or generative modeling techniques.  
The estimation of \( P(\mathbf{X}) \) can be approached using probabilistic models such as Gaussian Mixture Models (GMMs) \cite{reynolds2009gaussian,bishop2006pattern}, Hidden Markov Models (HMMs) \cite{rabiner1989tutorial}, and Bayesian inference \cite{fookes2020bayesian}, as well as deep generative models including Variational Autoencoders (VAEs) and Generative Adversarial Networks (GANs). VAEs \cite{kingma2013auto} structure data through a latent space representation, ensuring that the generated data retains essential patterns, while GANs employ an adversarial framework to refine \( \hat{P}(\mathbf{X}) \) to closely match \( P(\mathbf{X}) \). For cases where synthetic data needs to reflect specific characteristics, conditional distributions are used. Given an attribute \( A \), conditional generation follows
\[
\mathbf{X}_{\text{syn}} \sim \hat{P}(\mathbf{X} \mid A = a),
\]
which ensures that the generated data adheres to predefined conditions such as critical aspects in applications requiring fairness, bias control, or targeted augmentation \cite{ xu2019modeling}. Privacy preservation techniques modify \( \hat{P}(\mathbf{X}) \) to prevent sensitive information leakage. Differentially private synthetic data generation introduces noise either at the distribution estimation stage or during sampling, balancing the trade-off between privacy protection and data utility. By integrating these principles, synthetic data generation enables the creation of high-fidelity artificial datasets while addressing concerns of statistical integrity, controllability, and privacy. The following sections explore how different models approach these challenges and their implications for tabular data generation. \colorred{For completeness, Appendix~\ref{app:definitions} compares the representative definitions synthetic data generation from the literature and presents their corresponding mathematical formulation, while Appendix~\ref{app:data_types} provides additional background on the data types considered in synthetic data generation.}

\begin{figure*}[t]
\centering
\begingroup

\begin{tikzpicture}[
    x=1cm,
    y=0.62cm,
    objective/.style={
        rounded corners=1.5mm,
        minimum width=0.78cm,
        minimum height=0.42cm,
        font=\sffamily\bfseries\scriptsize,
        text=white,
        inner sep=1pt
    },
    inactive/.style={
        circle,
        minimum size=5.6pt,
        inner sep=0pt,
        draw=black!22,
        fill=black!7,
        line width=0.45pt
    },
    active/.style={
        circle,
        minimum size=6.4pt,
        inner sep=0pt,
        line width=0.55pt
    },
    method/.style={
        anchor=west,
        align=left,
        text width=11.25cm,
        font=\fontsize{7.6}{8.5}\selectfont,
        inner sep=0pt
    },
    connector/.style={
        draw=black!42,
        line width=0.75pt,
        line cap=round
    }
]

\path[use as bounding box] (-0.55,-7.35) rectangle (14.25,0.55);

\fill[black!2.5,rounded corners=1mm] (-0.45,-0.52) rectangle (14.15,-1.30);
\fill[black!2.5,rounded corners=1mm] (-0.45,-1.73) rectangle (14.15,-2.27);
\fill[black!2.5,rounded corners=1mm] (-0.45,-2.93) rectangle (14.15,-3.47);
\fill[black!2.5,rounded corners=1mm] (-0.45,-4.13) rectangle (14.15,-4.67);
\fill[black!2.5,rounded corners=1mm] (-0.45,-5.33) rectangle (14.15,-5.87);
\fill[black!2.5,rounded corners=1mm] (-0.45,-6.53) rectangle (14.15,-7.07);

\node[objective,fill=fdcolor]  at (0.00,0) {FDs};
\node[objective,fill=pspcolor] at (0.82,0) {PSPs};
\node[objective,fill=csacolor] at (1.64,0) {CSAs};
\node[objective,fill=ppcolor]  at (2.46,0) {PP};
\node[anchor=west,font=\sffamily\bfseries\footnotesize]
    at (3.15,0) {Representative approaches};

\foreach \x/\c in {0.00/fdcolor,0.82/pspcolor,1.64/csacolor,2.46/ppcolor}{
    \draw[\c!20,line width=0.45pt] (\x,-0.30) -- (\x,-7.12);
}
\draw[black!16,line width=0.45pt] (2.87,0.30) -- (2.87,-7.12);

\foreach \x in {0.00,0.82,1.64,2.46}{\node[inactive] at (\x,-0.90) {};}
\node[active,draw=ppcolor!85!black,fill=ppcolor] at (2.46,-0.90) {};
\node[method] at (3.15,-0.90) {PATE-GAN~\cite{jordon2018pate},
IT-GAN~\cite{lee2021invertible}, DP-SYN~\cite{abay2019privacy},
AIM~\cite{mckenna2022aim}, PEP/GEM~\cite{liu2021iterative}, and
\colorred{Prism~\cite{guan2026prism}}};

\foreach \x in {0.00,0.82,1.64,2.46}{\node[inactive] at (\x,-1.50) {};}
\draw[connector] (1.64,-1.50) -- (2.46,-1.50);
\node[active,draw=csacolor!85!black,fill=csacolor] at (1.64,-1.50) {};
\node[active,draw=ppcolor!85!black,fill=ppcolor] at (2.46,-1.50) {};
\node[method] at (3.15,-1.50) {DataSynthesizer~\cite{ping2017datasynthesizer}};

\foreach \x in {0.00,0.82,1.64,2.46}{\node[inactive] at (\x,-2.00) {};}
\draw[connector] (0.00,-2.00) -- (2.46,-2.00);
\node[active,draw=fdcolor!85!black,fill=fdcolor] at (0.00,-2.00) {};
\node[active,draw=ppcolor!85!black,fill=ppcolor] at (2.46,-2.00) {};
\node[method] at (3.15,-2.00) {EHR-Safe~\cite{yoon2023ehr}};

\foreach \x in {0.00,0.82,1.64,2.46}{\node[inactive] at (\x,-2.60) {};}
\draw[connector] (0.00,-2.60) -- (2.46,-2.60);
\node[active,draw=fdcolor!85!black,fill=fdcolor] at (0.00,-2.60) {};
\node[active,draw=csacolor!85!black,fill=csacolor] at (1.64,-2.60) {};
\node[active,draw=ppcolor!85!black,fill=ppcolor] at (2.46,-2.60) {};
\node[method] at (3.15,-2.60) {DP-CTGAN~\cite{fang2022dp}};

\foreach \x in {0.00,0.82,1.64,2.46}{\node[inactive] at (\x,-3.20) {};}
\draw[connector] (0.82,-3.20) -- (2.46,-3.20);
\node[active,draw=pspcolor!85!black,fill=pspcolor] at (0.82,-3.20) {};
\node[active,draw=ppcolor!85!black,fill=ppcolor] at (2.46,-3.20) {};
\node[method] at (3.15,-3.20) {PrivBayes~\cite{zhang2017privbayes} and
\colorred{HARMONIC~\cite{wang2024harmonic}}};

\foreach \x in {0.00,0.82,1.64,2.46}{\node[inactive] at (\x,-3.80) {};}
\node[active,draw=fdcolor!85!black,fill=fdcolor] at (0.00,-3.80) {};
\node[method] at (3.15,-3.80) {FCT-GAN~\cite{zhao2022fct},
TabDDPM~\cite{kotelnikov2023tabddpm}, TabTransformer~\cite{huang2020tabtransformer},
TabMT~\cite{gulati2023tabmt}, and \colorred{HFGF~\cite{umesh2026dependency}}};

\foreach \x in {0.00,0.82,1.64,2.46}{\node[inactive] at (\x,-4.40) {};}
\draw[connector] (0.00,-4.40) -- (1.64,-4.40);
\node[active,draw=fdcolor!85!black,fill=fdcolor] at (0.00,-4.40) {};
\node[active,draw=csacolor!85!black,fill=csacolor] at (1.64,-4.40) {};
\node[method] at (3.15,-4.40) {CTGAN~\cite{xu2019modeling},
\colorred{GReaT~\cite{borisov2022language}}, and
\colorred{Pred-LLM~\cite{nguyen2024generating}}};

\foreach \x in {0.00,0.82,1.64,2.46}{\node[inactive] at (\x,-5.00) {};}
\draw[connector] (0.00,-5.00) -- (1.64,-5.00);
\node[active,draw=fdcolor!85!black,fill=fdcolor] at (0.00,-5.00) {};
\node[active,draw=pspcolor!85!black,fill=pspcolor] at (0.82,-5.00) {};
\node[active,draw=csacolor!85!black,fill=csacolor] at (1.64,-5.00) {};
\node[method] at (3.15,-5.00) {Binary Diffusion~\cite{kinakh2024tabular},
CTAB-GAN~\cite{zhao2021ctab}, \colorred{MaCoDE~\cite{an2025masked}}, and
\colorred{TabDiff~\cite{shi2025tabdiff}}};

\foreach \x in {0.00,0.82,1.64,2.46}{\node[inactive] at (\x,-5.60) {};}
\draw[connector] (0.00,-5.60) -- (0.82,-5.60);
\node[active,draw=fdcolor!85!black,fill=fdcolor] at (0.00,-5.60) {};
\node[active,draw=pspcolor!85!black,fill=pspcolor] at (0.82,-5.60) {};
\node[method] at (3.15,-5.60) {STaSy~\cite{kim2022stasy},
TABSYN~\cite{zhang2023mixed}, Copula Flows~\cite{kamthe2021copula}, and
Synthsonic~\cite{baak2022synthsonic}};

\foreach \x in {0.00,0.82,1.64,2.46}{\node[inactive] at (\x,-6.20) {};}
\node[active,draw=pspcolor!85!black,fill=pspcolor] at (0.82,-6.20) {};
\node[method] at (3.15,-6.20) {medGAN~\cite{choi2017generating} and
Che et al.~\cite{che2017boosting}};

\node[active,draw=black!55,fill=black!55] at (0.00,-6.88) {};
\node[anchor=west,font=\fontsize{7.2}{8}\selectfont,text=black!70]
    at (0.18,-6.88) {addressed};
\node[inactive] at (1.65,-6.88) {};
\node[anchor=west,font=\fontsize{7.2}{8}\selectfont,text=black!70]
    at (1.73,-6.90) {not assigned};

\end{tikzpicture}

\endgroup

\Description{A membership-matrix taxonomy with four objective columns:
feature dependencies, preservation of statistical properties, conditioning on
specific attributes, and privacy preservation. Filled, connected markers show
the objectives explicitly addressed by each listed group of representative
tabular-data generators.}

\caption{\colorred{Objective-membership taxonomy of representative
tabular-data generators. Filled markers indicate addressed objectives;
connections indicate combinations. FDs: feature dependencies;
PSPs: preservation of statistical properties; CSAs: conditioning on
specific attributes; PP: privacy preservation. PP denotes privacy
treatment or evaluation, not necessarily a formal differential-privacy
guarantee.}}
\label{fig:taxnomy_venn}
\vspace{-0.8em}
\end{figure*}

\section{Tabular Synthetic Data Generation}
\label{sec:TSDG}

Tabular data, characterized by its structured format of rows and columns, is one of the most common and versatile forms of data used across numerous domains. The generation of tabular synthetic data presents unique challenges: (1) accurately capturing feature dependencies, which is crucial for maintaining the relationships between variables to ensure that the synthetic data reflect real-world patterns; (2) preserving statistical properties, ensuring that distributions, correlations, and higher-order moments are retained for meaningful analysis; (3) ensuring privacy, protecting sensitive information by preventing re-identification risks while maintaining data utility; (4) conditioning on specific attributes, allowing for the targeted generation of synthetic data based on predefined characteristics; and (5) addressing domain-specific requirements, ensuring that generated data adhere to the constraints and nuances of particular fields. 
For example, the Credit Risk dataset \cite{credit_risk_kaggle}, widely used in financial applications, exhibits complex relationships between variables such as loan amounts, credit scores, and default risks, which are essential for evaluating creditworthiness and risk assessment. Similarly, the Adult dataset \cite{adult_2}, derived from U.S. Census data and commonly used in social science research, contains intricate patterns between education level, occupation, and salary levels, reflecting demographic and economic trends. These patterns highlight the importance of preserving feature dependencies when generating synthetic data. Both the Credit Risk dataset and the Adult dataset will serve as reference points for evaluating and comparing synthetic data generation methods throughout this discussion. However, these challenges are often competing in nature, and it is difficult for a single model to simultaneously satisfy all of them effectively.

\raju{Existing synthetic data generation models differ significantly in how they address these challenges, with some approaches focusing on joint end-to-end modeling while others decompose the problem into multiple stages.} 
\raju{To systematically categorize these challenges and the corresponding synthesis approaches, we refer to the taxonomy depicted in (Fig.~\ref{fig:taxnomy_venn}), organizing methods based on their primary focus: Feature Dependencies (FDs), Preservation of Statistical
Properties (PSPs), Conditioning on Specific Attributes (CSAs) and Privacy Preservation (PP). This categorization provides a structured framework for analyzing and comparing different synthetic data generation models according to which aspects of data fidelity and privacy they prioritize.}
\subsection{Maintaining Feature Dependency}
\label{sec:mfdcy}
\paragraph{Significance and Challenges}
Maintaining feature dependency preservation is crucial in generating synthetic tabular data because real-world datasets often exhibit complex relationships between features. For example, in the credit risk dataset \cite{credit_risk_kaggle}, a customer’s creditworthiness is influenced by a combination of factors such as income, debt-to-income ratio, employment history, and credit score. Accurately replicating these dependencies is essential to ensure the synthetic data remains useful for downstream applications like training machine learning models or conducting risk analysis \cite{figueira2022survey}.
However, preserving these feature dependencies presents significant challenges. One issue arises from the non-linear relationships between features. In many datasets, including the credit risk dataset, the relationship between certain features is not straightforward. For instance, while higher income generally reduces the likelihood of default, this relationship may weaken or even reverse at extremely high income levels. Capturing such subtle, non-linear dependencies is critical for synthetic data generation models since any oversimplification or failure to account for these complexities could result in inaccurate representations of real-world data \cite{xu2019modeling}. Additionally, the high dimensionality and mixed data types in tabular datasets further complicate the process. Datasets like Credit Risk typically include both categorical (e.g., ``employment length\colorred{''}) and continuous (e.g., ``annual income\colorred{''}) features. Modeling these different feature types together while preserving the intricate relationships between them presents a significant challenge. The more dimensions a dataset has, the more complex the relationships between features become, requiring sophisticated techniques to ensure that all dependencies are maintained accurately across the entire dataset \cite{xu2018synthesizing}. Another significant challenge lies in imbalanced data distributions. In many real-world datasets, such as those used in credit risk modeling, rare but critical events like loan defaults are often underrepresented. Despite their infrequency, these events play a crucial role in accurate risk assessment and decision-making \cite{chawla2002smote}. Generating synthetic data that effectively captures these rare events while preserving the intricate relationships between features is a complex task \cite{douzas2019geometric}. If these rare dependencies are not adequately represented, the resulting synthetic data may fail to provide meaningful insights or, worse, lead to biased models that perform poorly on underrepresented but critical events. Moreover, some dependencies only emerge under specific conditions, adding another layer of complexity. For instance, the relationship between ``loan amount\colorred{''} and ``income\colorred{''} may vary depending on factors like ``employment history\colorred{''} \cite{xu2018synthesizing}. In such cases, the synthetic data generation model must be able to adjust for these conditional relationships, ensuring that the generated data accurately reflects the varied interactions between features under different conditions.
\colorred{Feature dependencies also include hard record-level constraints rather than only statistical associations. For example, a post date must follow its corresponding sign-in date, so a derived elapsed-time feature must be strictly positive. Such requirements can be expressed as functional dependencies, where one set of attributes determines another, or as logical dependencies that restrict valid combinations of values \cite{umesh2025preserving}. Matching marginal distributions or pairwise correlations does not guarantee either form of dependency. Beyazit et al. \cite{beyazit2023inductive} show that tabular prediction functions can exhibit high-frequency irregularity that conflicts with the smooth-function bias of neural networks. Although their analysis concerns prediction rather than synthetic-data generation, it provides a plausible explanation for difficulty learning sharp rule boundaries; therefore, hard-constraint satisfaction must be evaluated directly.}
Finally, in domains such as finance or healthcare, features often exhibit temporal dependencies. For example, a person’s past credit history can influence their current creditworthiness, and these relationships may evolve over time \cite{yoon2019time}. Generating synthetic data that captures both the static and dynamic aspects of these temporal dependencies requires advanced techniques capable of modeling the changing nature of feature interactions.

\begin{table*}[htpb]
\centering
\caption{Maintaining feature-dependency approaches and the challenges they explicitly address.}
\label{tab:feature_dependency}

\footnotesize
\setlength{\tabcolsep}{2.2pt}

\begin{tabular}{@{}l*{11}{c}@{}}
\hline
\textbf{Challenges}
& \rotatebox{65}{\textbf{CTGAN}}
& \rotatebox{65}{\textbf{FCT-GAN}}
& \rotatebox{65}{\textbf{CTAB-GAN}}
& \rotatebox{65}{\textbf{Synthsonic}}
& \rotatebox{65}{\textbf{TabDDPM}}
& \rotatebox{65}{\textbf{TabTransformer}}
& \rotatebox{65}{\textbf{TabMT}}
& \rotatebox{65}{\colorred{\textbf{Pred-LLM}}}
& \rotatebox{65}{\colorred{\textbf{MaCoDE}}}
& \rotatebox{65}{\colorred{\textbf{TabDiff}}}
& \rotatebox{65}{\colorred{\textbf{HFGF}}} \\
\hline

Non-linear Relationships
& \checkmark & \checkmark & \checkmark & & \checkmark & & &
& \colorred{\checkmark} & \colorred{\checkmark} & \\
\hline

\shortstack[l]{High Dimensionality and \\ Mixed Data Types}
& \checkmark & & \checkmark & \checkmark & \checkmark & & &
& & \colorred{\checkmark} & \\
\hline

Imbalanced Data Distributions
& \checkmark & & & & & & &
& & & \\
\hline

Conditional Feature Dependencies
& \checkmark & & \checkmark & & & & &
\colorred{\checkmark}
& \colorred{\checkmark}
& \colorred{\checkmark}
& \colorred{\checkmark} \\
\hline

Temporal Dependencies
& & & & & & \checkmark & \checkmark
& & & & \\
\hline

\shortstack[l]{Complex Multi-way \\ Feature Dependencies}
& & & \checkmark & & & & &
& \colorred{\checkmark}
& \colorred{\checkmark}
& \\
\hline

\colorred{\shortstack[l]{Functional and Logical Dependencies}}
& & & & & & & &
& & & \colorred{\checkmark} \\
\hline
\end{tabular}

\vspace{-0.8em}
\end{table*}
\paragraph{Approaches}

Several synthetic data generation models have attempted to address the aforementioned challenges and thus preserve feature dependency effectively.

\textbf{Conditional Tabular GAN (CTGAN)} \cite{xu2019modeling} introduces a conditional generator that models the data distribution conditioned on a sampled categorical column and value. To effectively handle imbalanced categorical data, CTGAN employs \raju{a mode-specific normalization technique, which transforms numerical features based on their underlying mixture distribution rather than conditioning on categorical attributes. This normalization reduces bias toward dominant regions of continuous features and improves representation of multimodal distributions.}  

\raju{Given a numerical feature \( x \), CTGAN first models its distribution using a Gaussian mixture model. 
For each observed value \( x \), the most probable mixture component \( k^* \) is identified, and normalization is performed relative to that component as 
\begin{equation}
\tilde{x} = \frac{x - \mu_{k^*}}{\sigma_{k^*}},
\label{eq:modenorm}  
\end{equation}
where \( \mu_{k^*} \) and \( \sigma_{k^*} \) denote the mean and standard deviation of the selected mixture component.
}  
\colorred{During training, CTGAN randomly samples a categorical column \(x_c\) 
and one of its values \(v_c\) from the data distribution and conditions the 
generator on a corresponding one-hot vector \(\mathbf{c}\). This vector is 
defined over all categorical values across all columns, with only the entry 
corresponding to \(v_c\) set to one. The conditioned generator produces an 
encoded synthetic record as}
\[
\colorred{
G(\mathbf{z} \mid \mathbf{c}) \rightarrow \tilde{\mathbf{x}}',
}
\]
\colorred{where \(\mathbf{z} \sim P_{\mathbf{z}}\) is the latent noise vector 
and \(\tilde{\mathbf{x}}'\) is the encoded generated record. The discriminator 
receives an encoded real or generated record together with the same condition 
vector and outputs the probability that the record is real:}
\[
\colorred{
D(\tilde{\mathbf{x}} \mid \mathbf{c}) \rightarrow [0,1].
}
\]
\colorred{The CTGAN training objective is given in Eq.~\ref{eq:ctgan_loss}:}
\begin{equation}
\colorred{
\min_G \max_D
\mathbb{E}_{\tilde{\mathbf{x}} \sim P_{\mathrm{data}}}
\left[\log D(\tilde{\mathbf{x}} \mid \mathbf{c})\right]
+
\mathbb{E}_{\mathbf{z} \sim P_{\mathbf{z}},\,\mathbf{c} \sim P_{\mathbf{c}}}
\left[
\log\!\left(
1-D(G(\mathbf{z}\mid\mathbf{c})\mid\mathbf{c})
\right)
\right],
}
\label{eq:ctgan_loss}
\end{equation}
\colorred{where \(P_{\mathbf{c}}\) denotes the sampling distribution over 
condition vectors.}
\raju{This formulation ensures consistency between conditional variables in both real and generated data.}  

CTGAN effectively preserves inter-feature relationships through conditional modeling—for instance, when conditioned on employment type, the generated income and loan default probabilities remain consistent with realistic correlations. However, despite its strengths, CTGAN may struggle to capture complex multi-way dependencies that require conditioning on multiple categorical attributes simultaneously.

\textbf{CTAB-GAN \cite{zhao2021ctab}} extends the CTGAN framework by 
\raju{introducing additional architectural components and auxiliary loss terms to better preserve complex feature dependencies in mixed-type tabular data}. 
\raju{Unlike CTGAN, which primarily relies on conditional sampling for handling categorical variables, CTAB-GAN incorporates auxiliary objectives that operate on a joint mixed-type feature representation learned by an encoder. This enables more effective modeling of complex feature relationships.} 
\colorred{The CTAB-GAN generator objective is given in
Eq.~\ref{eq:ctab_generator_loss}:}
\begin{equation}
\colorred{
L_G =
L_G^{\mathrm{orig}}
+ L_G^{\mathrm{info}}
+ L_G^{\mathrm{class}}
+ L_G^{\mathrm{generator}},
}
\label{eq:ctab_generator_loss}
\end{equation}
where $L_G^{\mathrm{orig}}$ is the standard adversarial loss used to train the generator and discriminator. 
The information loss $L_G^{\mathrm{info}}$ 
\raju{encourages the generated data to match the statistical properties of the real data by aligning first- and second-order moments}:
\begin{equation}
L_G^{\mathrm{info}} =
\left\| \mathbb{E}[\mathbf{x}_{\text{real}}] - \mathbb{E}[\mathbf{x}_{\text{gen}}] \right\|_2^2
+
\left\| \text{SD}(\mathbf{x}_{\text{real}}) - \text{SD}(\mathbf{x}_{\text{gen}}) \right\|_2^2,
\label{eq:ctab_infoloss}
\end{equation}
where $\mathbf{x}_{\text{real}}$ and $\mathbf{x}_{\text{gen}}$ denote real and generated samples respectively, and 
\raju{$\text{SD}(\cdot)$ represents the standard deviation operator}. 
Class loss $L_G^{\mathrm{class}}$ enforces label consistency between generated samples and their intended class labels:
\begin{equation*}
L_G^{\mathrm{class}} =
\mathbb{E}\left[
\left| l(G(z)) - C(f_e(G(z))) \right|
\right]_{z \sim p(z)},
\end{equation*}
where 
\raju{$l(G(z))$ denotes the target label associated with the generated sample}, and 
\raju{$C(f_e(G(z)))$ denotes the classifier prediction obtained by applying the classifier $C$ to the encoded generated sample with an encoder \( f_e \).} 
\colorred{Lastly, the generator loss}
\begin{equation*}
\colorred{
L_G^{\mathrm{generator}} = H(m_i,\hat{m}_i),
}
\end{equation*}
\colorred{is the cross-entropy between the given and generated
conditional-vector bits, where \(m_i\) and \(\hat{m}_i\) denote those
bits for the selected column \(i\), and \(H(\cdot,\cdot)\) denotes
cross-entropy. This term encourages the generator to reproduce the
sampled condition.}
\raju{By combining adversarial learning with statistical alignment and auxiliary supervision, CTAB-GAN improves the preservation of feature dependencies compared to CTGAN. For example, in datasets that contain both continuous variables (e.g., ``credit score\colorred{''}, ``income\colorred{''}) and categorical attributes (e.g., ``loan purpose\colorred{''}), feature relationships are often conditional on combinations of variables. Higher income may increase the likelihood of loan approval only for certain loan purposes and within a certain credit score ranges. Such cross-type, higher-order interactions can be difficult for CTGAN to capture since its conditional sampling conditions on a single categorical attribute at a time. In contrast, CTAB-GAN models these dependencies more effectively by learning a joint representation of all features and using auxiliary losses.}

\textbf{Fourier Conditional Tabular GAN (FCT-GAN) \cite{zhao2022fct}} uses two key innovations, feature tokenization, \raju{which maps heterogeneous tabular features into learnable vector representations} and Fourier transformations, to improve the modeling of tabular data. As illustrated in Figure~\ref{fig:fct_gan}, the feature tokenizer first preprocesses mixed data types (e.g., categorical and numerical features) into a structured format suitable for deep learning. 
\raju{The tokenized feature representations are then transformed into the frequency domain using Fourier transformations, enabling the model to capture complex non-linear dependencies more effectively}. By representing data as combinations of sinusoidal components, this approach can reveal underlying periodic patterns and global feature interactions that are difficult to detect in the original domain. For instance, relationships such as those between ``loan terms\colorred{''}, ``interest rates\colorred{''}, and ``default probabilities\colorred{''} can be better modeled using this frequency-based approach.
 FCT-GAN excels in capturing intricate, nonlinear interactions that traditional GANs often miss, making it particularly effective for datasets with complex feature dependencies. However, it requires careful tuning of hyperparameters, such as the number of Fourier components, to balance the accuracy and computational complexity.

\raju{\textbf{Synthsonic \cite{baak2022synthsonic}} addresses feature dependency modeling through an explicit probabilistic factorization using a tree-structured Bayesian Network (BN). After transforming the data into a copula space, where marginal effects are removed, the model learns dependencies by factorizing the joint distribution as:
\begin{equation*}
P(\mathbf{U}) = \prod_{i=1}^{d} P(U_i \mid \text{Pa}(U_i)),
\end{equation*}
where $\text{Pa}(U_i)$ denotes the parent set of feature $U_i$ determined using a Chow--Liu \cite{chow1968approximating} or Tree-Augmented Naive Bayes (TAN) \cite{friedman1997bayesian} structure. This formulation enables Synthsonic to capture dominant pairwise dependencies efficiently. However, since the structure is restricted to a tree, it cannot directly represent higher-order interactions among multiple features. To address this, Synthsonic applies a second stage where a calibrated discriminative model is used to adjust the generated samples, helping recover some of the dependencies that are not captured by the BN. Overall, Synthsonic models feature dependencies explicitly through the Bayesian Network for pairwise relationships, and uses the correction stage to improve the quality of more complex interactions.
}

\textbf{Tabular Denoising Diffusion Probabilistic Models (TabDDPM) \cite{kotelnikov2023tabddpm} }iteratively denoise a data distribution, capturing intricate dependencies by modeling the data generation process as a stochastic
transformation. TabDDPM employs a combination of Gaussian diffusion for numerical features
and multinomial diffusion for categorical and binary features, allowing it to handle mixed data
types effectively. For a tabular data sample
$\mathbf{x} = [\mathbf{x}_{\text{num}}, \mathbf{x}_{\text{cat}}^{1}, \ldots, \mathbf{x}_{\text{cat}}^{C}]$,
where $\mathbf{x}_{\text{num}} \in \mathbb{R}^{N_{\text{num}}}$ represents numerical features and
$\mathbf{x}_{\text{cat}}^{i}$ represents categorical features with $K_i$ categories each, TabDDPM
processes one-hot encoded categorical features and normalized numerical features. The model
uses a multi-layer neural network to predict the noise $\epsilon$ for Gaussian diffusion and the
categorical probabilities for multinomial diffusion during the reverse process. This enables
TabDDPM to capture subtle correlations, such as how ``debt-to-income ratio'' and ``employment
history'' jointly influence the ``loan default'' variable.
However, TabDDPM is computationally intensive, requiring a large number of iterations to achieve
convergence. 

\raju{\textbf{TabTransformer \cite{huang2020tabtransformer}} is a Transformer-based model
designed for tabular data that 
captures feature dependencies via self-attention \cite{vaswani2017attention}.} It leverages contextual embeddings to model the complex interactions between features, even in cases where those dependencies evolve over time or are not immediately apparent. The model achieves this by using the self-attention mechanism, which helps it to focus on the most relevant features for each record, taking into account both direct and indirect relationships. The self-attention mechanism calculates dependencies between features  \raju{by computing attention weights between their contextual embeddings,as defined in Section~\ref{subsec:generative_models} (Eq.~\ref{eq:selfattention})}. This enables the model to learn and maintain dependencies between features even when their relationships are non-linear or complex. For instance, in a financial dataset, TabTransformer can capture the way features like payment history, loan amounts, and credit scores influence one another over time, maintaining these inter-feature dependencies across the entire dataset. This ability to focus on important features while preserving their interdependencies makes TabTransformer a powerful tool for generating synthetic tabular data that respects the underlying structure of the original data.

\textbf{Tabular Masked Transformer (TabMT) \cite{gulati2023tabmt}} is a tabular data generation model that leverages masked transformer mechanisms. Its structure is tailored to jointly model relationships among diverse features, providing several key advantages.
First, TabMT accounts for patterns bidirectionally between features. Unlike sequential data, tabular data lacks a natural ordering, meaning bidirectional learning enables the model to better understand and embed relationships between features, ensuring that dependencies between features are preserved. Second, TabMT's masking procedure allows for arbitrary prompts during generation, making it uniquely flexible compared to other generators with limited conditioning capabilities. This flexibility ensures that the model can generate data while maintaining the relational structure of the original dataset, even when prompted with partial or non-sequential inputs. Third, TabMT handles missing data effectively by setting the masking probability for missing values to 1, whereas other generators often require separate imputation steps. This approach allows TabMT to learn and preserve dependencies even in the presence of incomplete data, ensuring high-quality synthetic samples. These structural advantages stem from TabMT's novel masked generation procedure, which builds on the original BERT \cite{devlin2018bert} masking approach but introduces key modifications to improve generative performance. As illustrated in Figure~\ref{fig:tabmt}, TabMT processes categorical and numerical fields separately. For categorical fields, a standard embedding matrix is used, while numerical fields are quantized and represented using ordered embeddings. The masking probability \( p_i \) for the \( i \)-th row is sampled from a uniform distribution \( U(0, 1) \), ensuring \raju{that the training distribution of masked set sizes $|F_i^m|$ matches the uniform distribution encountered during generation, where $F_i^m$ denotes the set of masked fields in row $i$. This is achieved by:}
\[
P(|F^m_i| = k) = \int_0^1 \binom{l}{k} p^k (1-p)^{l-k} \, dp = \frac{1}{l + 1}.
\]
\raju{Here, $l$ is the total number of fields in a row, and $k$ is the number of masked fields.}
Additionally, TabMT predicts masked values in a random order during generation, addressing the mismatch between training and inference caused by fixed-order generation. This ensures that the model encounters each subset size uniformly, as:
\[
P(F^m_i = s) = \frac{t! \cdot (l - t)!}{l!} = \frac{1}{\binom{l}{t}},
\]
\raju{where $s$ denotes a particular subset of masked fields and $t$ is the number of fields generated at that step}. These modifications, combined with the transformer architecture and dynamic linear layer, enable TabMT to generate high-quality synthetic data that preserves the relational structure and dependencies of the original dataset, even in the presence of missing values or heterogeneous data types.
\colorred{
Pred-LLM \cite{nguyen2024generating} targets feature--target dependence through three mechanisms: permutation of serialized features during fine-tuning, feature-conditional sampling, and label generation by prompting the fine-tuned language model with the generated features. MaCoDE \cite{an2025masked} instead interprets masked language modeling as non-parametric conditional-density estimation, allowing arbitrary subsets of variables to be treated as observed conditions or generation targets. These approaches learn conditional statistical relationships; neither mechanism alone guarantees satisfaction of deterministic functional dependencies or domain rules.} \colorred{This distinction is supported by the evaluation of Umesh et al. \cite{umesh2025preserving}, in which none of the seven examined generators simultaneously preserved all tested functional and logical dependencies. Their subsequent Hierarchical Feature Generation Framework (HFGF) \cite{umesh2026dependency} separates independent and dependent features: a base synthesizer first generates the independent features, after which dependent features are reconstructed using predefined functional- and logical-dependency rules. These rules can include probabilistic one-to-many conditional relationships, but HFGF relies on dependencies that are predefined or identified separately and does not explicitly learn arbitrary multi-way interactions from the joint distribution.}

\paragraph{Success and Limitations}
The discussed models demonstrate varying strengths and limitations in preserving feature dependency for synthetic data generation. CTGAN effectively handles imbalanced data and captures simple feature dependencies but struggles with complex multi-way relationships \cite{zhao2021ctab}. CTAB-GAN balances the handling of mixed data types and captures both linear and non-linear interactions. However, due to its complex architecture, especially when learning dependencies across diverse data types and feature interactions, it can require significant computational resources. This is particularly true during the training phase, where the conditional GAN structure and the need for hyperparameter tuning can lead to increased training time and resource consumption. Additionally, the use of CNNs combined with one-hot encoding makes CTAB-GAN more vulnerable to column permutations, which can affect its robustness and generalizability in scenarios where the order of features varies \cite{zhao2022fct}. FCT-GAN excels in modeling non-linear dependencies through Fourier transformations, though it requires careful hyperparameter tuning to achieve optimal performance \cite{zhao2022fct}. Additionally, the use of Fourier transformations introduces significant computational complexity, making the model resource-intensive and less scalable for large datasets \cite{li2020fourier}. TabDDPM is highly effective at preserving intricate and subtle dependencies, as demonstrated by its ability to model complex data distributions through iterative denoising \cite{ho2020denoising}. However, this comes at the cost of being resource-intensive, requiring significant computational power for both training and sampling \cite{dhariwal2021diffusion}. Additionally, TabDDPM demands a large number of iterations to converge, which can lead to longer training times compared to other generative models \cite{nichol2021improved}. Lastly, Transformer-based models TabTransformer, TabMT are well-suited for temporal or sequential dependencies, leveraging their self-attention mechanisms to capture complex patterns in such data. However, they have limited utility for datasets lacking these structures, as their design is optimized for relational and sequential tasks \cite{vaswani2017attention}. Each model addresses specific challenges, making their selection dependent on the dataset characteristics and application requirements.

\subsection{Preserving Statistical Properties}
\label{sec:psp}
\paragraph{Significance and Challenges}Preserving statistical properties such as distributions, correlations, and higher-order moments is essential for generating realistic synthetic data. In datasets like Credit Risk, maintaining these properties ensures that key insights, such as default rates and income distributions, remain valid for downstream applications like risk modeling and financial analysis \cite{figueira2022survey}.
Distributions define the overall spread of data values, such as how annual incomes are distributed across different income brackets. If a real dataset shows that most individuals earn between \$40,000 and \$80,000, synthetic data should preserve this trend rather than generating incomes uniformly across all possible values. Correlations capture relationships between different features, such as how loan amounts typically increase with income. Ignoring these relationships in synthetic data may lead to misleading conclusions in financial modeling \cite{xu2019modeling}. Higher-order moments, such as skewness and kurtosis, further describe the shape of data distributions \cite{hoogeboom2021argmax}. For instance, if loan defaults are heavily skewed toward lower-income groups, synthetic data should replicate this pattern to ensure accurate risk assessment.

One major challenge in preserving statistical properties is marginal distribution mismatch, where synthetic data fails to replicate the real distribution of individual features, such as ``annual income\colorred{''} or ``credit score\colorred{,''} leading to unrealistic results. A distorted income distribution, for example, could misrepresent credit risk levels \cite{patki2016synthetic}. Another challenge is joint distribution representation, where capturing complex correlations between features, such as ``loan amount\colorred{''} and ``income\colorred{,''} becomes difficult as dimensionality increases \cite{nelsen2006introduction}. Additionally, rare event simulation poses difficulties, as rare but critical occurrences, like loan defaults, must be accurately reproduced for effective risk modeling\cite{figueira2022survey}.
\begin{table*}[htpb]
    \centering
    \caption{Preserving Statistical Properties Challenges and  Models that address}
    \resizebox{\textwidth}{!}{
    \begin{tabular}{lccccccc}
        \hline
        \textbf{Challenges}
& \textbf{CTAB-GAN}
& \textbf{Copula Flows}
& \textbf{PrivBayes}
& \textbf{STaSy}
& \textbf{TABSYN}
& \textbf{medGAN}
& \colorred{\textbf{TabDiff}} \\
\hline

Marginal Distribution Mismatch
& \checkmark & & \checkmark & \checkmark & \checkmark & &
\colorred{\checkmark} \\
\hline

Joint Distribution Representation
& \checkmark & \checkmark & & \checkmark & \checkmark & \checkmark &
\colorred{\checkmark} \\
\hline

\colorred{Rare Event Simulation}
& & \checkmark & & & & & \\
\hline
    \end{tabular}
    }
    \label{tab:statistical_properties}
    \vspace{-0.8em}
\end{table*}

\paragraph{Approaches} Several models have been developed to address the challenges of preserving statistical properties, particularly the distributional characteristics, correlations, and higher-order moments, in synthetic data generation. Each model offers unique methods for tackling these challenges, making them suitable for different types of data and application contexts. Table \ref{tab:statistical_properties} summarizes how these models address specific challenges related to preserving statistical properties, highlighting their strengths and limitations. \raju{As shown in Table \ref{tab:statistical_properties}, among prior category methods, CTAB-GAN is the only approach that explicitly incorporates mechanisms to preserve such statistical properties.}

\textbf{CTAB-GAN }preserves statistical properties through its Information Loss \colorred{(\(L_G^{\mathrm{info}}\))} Eq.(\ref{eq:ctab_infoloss}), which ensures that the synthetic data matches the first-order (mean) and second-order (standard deviation) statistics of the real data. This loss is critical for maintaining the statistical integrity of the generated data. In addition, CTAB-GAN employs a variational Gaussian mixture model (VGM) \cite{bishop2006pattern} to encode continuous features (e.g., credit scores, loan amounts) and one-hot encoding for categorical features (e.g., loan purpose, employment status). The VGM fits a mixture of Gaussian distributions to the data, ensuring accurate modeling of continuous feature distributions. However, the preservation of statistical properties is primarily enforced through the Information Loss, which explicitly matches the mean and variance of real and synthetic data. For example, in a credit risk dataset, CTAB-GAN uses \colorred{\(L_G^{\mathrm{info}}\)} to ensure that the synthetic data preserves the mean and variance of features such as credit scores and loan amounts. This is crucial for maintaining the statistical fidelity of the dataset, as it ensures that the synthetic data reflects the same central tendency and variability as the real data. However, CTAB-GAN may struggle with rare event simulation, such as generating high-risk borrowers with extremely low credit scores, due to the GAN's tendency to underrepresent minority classes.

\textbf{Copula Flows \cite{kamthe2021copula}} preserve statistical properties by decomposing the joint distribution of the data into marginal distributions and a copula function. The joint density \( f_X \) is expressed as:
\[
f_X(X) = c_X(F_{X_1}, \dots, F_{X_d}) \prod_{k=1}^d f_{X_k}(X_k),
\]
where \( c_X \) is the copula density obtained as the derivative of the copula CDF (cumulative distribution function) \cite{kamthe2021copula, sklar1959fonctions}, \raju{which captures the dependency structure among variables independently of their marginal distributions. In other words, \( c_X \) models how features interact once they have been transformed to the uniform space via their marginals.} \( F_{X_k} \) are the marginal CDFs, and \( f_{X_k} \) are the marginal densities. The marginal flows \( F_{X_k} \) are trained using neural spline flows \cite{durkan2019neural} to approximate the true CDFs, ensuring that the synthetic data matches the marginal distributions of the original data. For example, in a credit risk dataset, the marginal flow for ``credit scores\colorred{''} ensures that the synthetic credit scores have the same distribution as the real data, while the marginal flow for ``loan amounts\colorred{''} preserves the distribution of loan values. The copula function \( c_X \) captures the dependencies between features, preserving the joint distribution. For instance, in a credit risk dataset, the copula function ensures that the relationship between ``credit scores\colorred{''} and ``default rates\colorred{''} is maintained in the synthetic data. This is achieved by modeling the joint CDF \( C_X(F_{X_1}, \dots, F_{X_d}) \), which describes how the marginal distributions interact. Copula Flows are particularly effective at capturing rare events, as they model tail dependencies in the data. However, their performance in rare event simulation can be affected by the representation of such events in the training data; if the rare events are too sparse or underrepresented, the model may struggle to generate them accurately.

\textbf{Privacy-Preserving Bayesian Network (PrivBayes) \cite{zhang2017privbayes}} preserves statistical properties by modeling the joint distribution of the data using a Bayesian network, which decomposes the joint distribution into a product of conditional distributions, as shown in Eq.(~\ref{eq:bayesian}).
PrivBayes learns the structure of the Bayesian network and the conditional distributions from the data, ensuring that the synthetic data matches the marginal and conditional distributions of the original data. By doing so, PrivBayes helps mitigate Marginal Distribution Mismatch, ensuring that the generated data accurately reflects the statistical properties of the original dataset at the feature level. For example, in a credit risk dataset, PrivBayes can model the conditional distribution of ``default rates\colorred{''} given ``credit scores\colorred{''} and ``income levels\colorred{''}, ensuring that the synthetic data preserves these relationships. \raju{While} it captures local dependencies, PrivBayes may not fully represent complex joint distributions across multiple features, limiting its ability to address joint distribution representation comprehensively. \colorred{In addition,} PrivBayes may struggle with rare event simulation, such as generating high-risk borrowers with extremely low credit scores, due to the limitations of Bayesian networks in capturing rare occurrences in the data.  

\textbf{Score-based Tabular Data Synthesis (STaSy) \cite{kim2022stasy}} is a model designed to preserve the statistical properties of tabular data by leveraging score-based generative models \cite{song2020score}. Unlike traditional generative models, STaSy incorporates self-paced learning to ensure training stability. This allows for the generation of synthetic tabular data that accurately reflects the statistical properties of the original dataset. One key aspect of STaSy is its ability to preserve both the marginal distributions and the dependencies between features. The model iteratively refines noisy data by utilizing a denoising score matching objective, which approximates the gradient of the log-likelihood to ensure that the generated data aligns with the original data’s statistical characteristics. The denoising score matching objective for the \(i\)-th record \colorred{\(\mathbf{x}_i\)} is defined by a function that measures the discrepancy between the predicted score and the gradient of the log-likelihood of the data. The denoising score matching loss for the \(i\)-th record is defined as follows:

\[
l_i = \mathbb{E}_{t} \mathbb{E}_{\mathbf{x}_i(t)} \left[ \lambda(t) \left\| S_\theta(\mathbf{x}_i(t), t) - \nabla_{\mathbf{x}_i(t)} \log p(\mathbf{x}_i(t) \mid \mathbf{x}_i(0)) \right\|^2 \right],
\]
where \(S_\theta(\mathbf{x}_i(t), t)\) is the score function, which estimates the gradient of the log-likelihood of the noisy data \(\mathbf{x}_i(t)\), and \(\lambda(t)\) is a scaling function that adjusts the emphasis on different noise scales during training. The function \(\nabla_{\mathbf{x}_i(t)} \log p(\mathbf{x}_i(t) \mid \mathbf{x}_i(0))\) represents the true gradient of the log-likelihood, and \(t\) refers to the noise level applied to the data. The STaSy objective function combines this score matching loss with a self-paced regularizer to effectively balance the model’s learning process, adjusting the contribution of each record during training. The overall objective is:

\[
\min_{\theta, v} \sum_{i=1}^{N} v_i l_i + r(v; \alpha, \beta),
\]
where \(0 \leq v_i \leq 1\) for all \(i\), and \(r(v; \alpha, \beta)\) is the self-paced regularizer. The parameters \(\alpha\) and \(\beta\), which belong to the interval \([0, 1]\), control the thresholds for engagement in the training process as it progresses. The variable \(v_i\) controls the participation of the \(i\)-th record, with a value of 1 indicating full participation and 0 indicating no participation. 

\colorred{These mechanisms were evaluated in terms of overall sampling quality, diversity, and downstream performance. However, STaSy does not report a frequency-specific evaluation establishing reliable reproduction of extremely rare events, such as an event occurring once in 10,000 records; overall distributional similarity should therefore not be interpreted as evidence of coverage at that frequency \cite{kim2022stasy}.}

\textbf{TABSYN \cite{zhang2023mixed}}, a principled approach for tabular data synthesis preserves statistical properties through a two-stage process, as illustrated in Figure \ref{fig:tabsyn}. In the first stage, each row of tabular data \(\mathbf{x} = [\mathbf{x}^{\text{num}}, \mathbf{x}^{\text{cat}}]\) is mapped to a latent space \(z\) using a column-wise tokenizer and an encoder. The tokenizer converts each column (both numerical and categorical) into a \(d\)-dimensional vector. For numerical columns, the tokenizer applies a linear transformation:
\[
e^{\text{num}}_i = x^{\text{num}}_i \cdot w^{\text{num}}_i + b^{\text{num}}_i,
\]
where \(w^{\text{num}}_i\) and \(b^{\text{num}}_i\) are learnable parameters. For categorical columns, the tokenizer uses an embedding lookup:
\[
e^{\text{cat}}_i = \mathbf{x}^{\text{oh}}_i \cdot W^{\text{cat}}_i + b^{\text{cat}}_i,
\]
where \(\mathbf{x}^{\text{oh}}_i\) is the one-hot encoded categorical feature, and \(W^{\text{cat}}_i\) and \(b^{\text{cat}}_i\) are learnable parameters. The tokenized representations \(E = [e^{\text{num}}_1, \dots, e^{\text{num}}_{M_{\text{num}}}, e^{\text{cat}}_1, \dots, e^{\text{cat}}_{M_{\text{cat}}}]\) are fed into a Transformer-based encoder to produce latent embeddings \(z\). This ensures that the synthetic data retains the marginal distributions of individual features, such as credit scores and loan amounts in a creditrisk dataset.

In the second stage, a diffusion process is applied in the latent space. The forward process gradually adds noise to the latent embeddings \(z_0 \rightarrow z_T\):
\[
z_t = z_0 + \sigma(t) \epsilon, \quad \epsilon \sim \mathcal{N}(0, I),
\]
where \(\sigma(t)\) is the noise level at time \(t\). The reverse process denoises the embeddings \(z_T \rightarrow z_0\), starting from a base distribution \(p(z_T)\). This ensures that the synthetic data preserves the joint distribution of features, such as the relationship between credit scores and default rates. Finally, the latent embeddings \(z_0\) are mapped back to the data space \(\tilde{\mathbf{x}}\) using a decoder and detokenizer:
\[
\tilde{x}^{\text{num}}_i = \tilde{e}^{\text{num}}_i \cdot \tilde{w}^{\text{num}}_i + \tilde{b}^{\text{num}}_i, \quad \tilde{\mathbf{x}}^{\text{cat}}_i = \text{Softmax}(\tilde{e}^{\text{cat}}_i \cdot \tilde{W}^{\text{cat}}_i + \tilde{b}^{\text{cat}}_i),
\]
where \(\tilde{w}^{\text{num}}_i\), \(\tilde{b}^{\text{num}}_i\), \(\tilde{W}^{\text{cat}}_i\), and \(\tilde{b}^{\text{cat}}_i\) are learnable parameters of the detokenizer. TABSYN effectively addresses marginal distribution mismatch and joint distribution representation by leveraging the tokenizer and encoder to model marginal distributions and the diffusion process to capture dependencies between features. However, it may struggle with Rare Event Simulation, such as generating high-risk borrowers with extremely low credit scores, due to the diffusion process's focus on the bulk of the data distribution rather than the tails.

\colorred{\textbf{TabDiff \cite{shi2025tabdiff}} jointly models numerical and categorical columns through feature-wise learnable diffusion processes and a mixed-type sampler. It evaluates marginal and pairwise-correlation fidelity and uses classifier-free guidance for conditional missing-column imputation; the method does not provide a formal privacy guarantee.}

\textbf{Medical Generative Adversarial Network (medGAN) \cite{choi2017generating}} preserves the statistical properties of the original electronic health records (EHR) data through a combination of an autoencoder and generative adversarial networks (GANs). The autoencoder, comprising an encoder $\text{Enc}$ and a decoder $\text{Dec}$, is pre-trained to learn the salient features of the input data $\mathbf{x} \in \mathbb{Z}_+^{|C|}$. The encoder compresses $\mathbf{x}$ into a latent representation $\text{Enc}(\mathbf{x}) \in \mathbb{R}^h$, and the decoder reconstructs it, minimizing the reconstruction loss, $L_R$:
\[
L_R = 
\begin{cases}
    \frac{1}{m} \sum_{i=0}^m \|\mathbf{x}_i - \mathbf{x}'_i\|_2^2  & \quad \text{if } \mathbf{x}_i \text{ is a count variable}\\ 
    \frac{1}{m} \sum_{i=0}^m \mathbf{x}_i \log \mathbf{x}'_i + (1 - \mathbf{x}_i) \log(1 - \mathbf{x}'_i)  & \quad \text{if } \mathbf{x}_i \text{ is a binary variable}
  \end{cases}
\]
where $\mathbf{x}'_i = \text{Dec}(\text{Enc}(\mathbf{x}_i))$. As shown in the architecture diagram (Figure ~\ref{fig:medGAN}), the generator $G$ takes a random prior $\mathbf{z} \in \mathbb{R}^r$ and produces synthetic data $G(\mathbf{z})$, which is then mapped to discrete values using the pre-trained decoder $\text{Dec}(G(\mathbf{z}))$. The discriminator $D$ is trained to distinguish between real data $\mathbf{x}$ and synthetic data $\text{Dec}(G(\mathbf{z}))$, ensuring that the synthetic data matches the distribution of the real data through the adversarial objective: 

\[
\min_G \max_D V(G, D) = \mathbb{E}_{\mathbf{x} \sim p_{\text{data}}} [\log D(\mathbf{x})] + \mathbb{E}_{\mathbf{z} \sim p_z} [\log(1 - D(\text{Dec}(G(\mathbf{z}))))].
\]
To prevent mode collapse, medGAN introduces minibatch averaging, where the discriminator is provided with the average of minibatch samples:

\[
\theta_d \leftarrow \theta_d + \alpha \nabla_{\theta_d} \frac{1}{m} \sum_{i=1}^m \log D(\mathbf{x}_i, \bar{\mathbf{x}}) + \log(1 - D(\mathbf{x}_{z_i}, \bar{\mathbf{x}}_z)),
\]
where $\bar{\mathbf{x}} = \frac{1}{m} \sum_{i=1}^m \mathbf{x}_i$ and $\bar{\mathbf{x}}_z = \frac{1}{m} \sum_{i=1}^m \mathbf{x}_{z_i}$. Additionally, batch normalization and shortcut connections are used to stabilize training and improve learning efficiency. Despite these strengths, medGAN has limitations. \raju{medGAN captures correlations implicitly through the autoencoder, but it lacks explicit mechanisms for enforcing feature dependencies across features.} 
\raju{In contrast, other GAN-based models introduced in Section~\ref{sec:mfdcy}, such as CTGAN and CTAB-GAN, explicitly model feature dependencies via conditional generators, loss functions, or conditional vectors. As a result, medGAN may generate synthetic data with unrealistic relationships among variables.} 
The model also lacks mechanisms for conditioning on specific attributes, limiting its ability to generate targeted synthetic data for specific subsets of patients. Furthermore, while minibatch averaging helps prevent mode collapse, the model may still struggle with highly imbalanced datasets or rare categories. Finally, the reliance on GANs makes training computationally expensive and sensitive to hyperparameter tuning, which could hinder scalability and practicality in real-world applications.

\paragraph{Success and Limitations}
Preserving statistical properties in synthetic data generation has yielded significant progress, enhancing data utility and enabling robust applications across various domains. A key success lies in ensuring synthetic datasets mirror the statistical characteristics of real data, making them viable for downstream tasks like model training, fairness evaluation, and bias mitigation. Additionally, these techniques have empowered scenario analysis and stress testing, particularly in high-stakes domains where understanding rare events and tail risks is critical. By capturing complex feature dependencies, models that prioritize statistical fidelity improve the reliability of synthetic data for predictive analytics and decision-making.

However, preserving statistical properties introduces several challenges. Ensuring accurate reproduction of both marginal and joint distributions is non-trivial, especially with heterogeneous data types and intricate feature relationships. One notable limitation is the risk of overfitting, where synthetic data closely mimics the original, inadvertently compromising privacy. Moreover, balancing privacy and utility poses a constant trade-off, models that focus on statistical accuracy may reveal subtle patterns that adversaries could exploit. Another persistent challenge is generalizability: techniques optimized for one domain may struggle to adapt to others due to differing data structures and contextual dependencies. Computational complexity is also a concern, as capturing nuanced statistical relationships demands more sophisticated architectures and prolonged training times, limiting scalability. These successes and limitations reflect the ongoing effort to balance statistical fidelity, privacy preservation, and computational efficiency in synthetic data generation.

\subsection{Preserving Data Privacy}
\label{sec:pdp}
\paragraph{Significance and Challenges}Privacy preservation is essential in synthetic data generation, particularly for sensitive datasets such as Credit Risk, which contain personal financial details \cite{dwork2014algorithmic}. The challenge lies in ensuring that synthetic data maintains statistical utility for tasks like credit risk assessment while preventing adversaries from inferring sensitive attributes. Synthetic data should protect individuals’ information while preserving patterns that make it useful for downstream tasks. Techniques such as differential privacy have emerged as a key solution, as highlighted in foundational works like \raju{Dwork et al. \cite{dwork2014algorithmic}}. One of the primary risks is reconstruction attacks, where adversaries attempt to reverse-engineer real data from synthetic datasets. If synthetic records closely resemble actual individuals' data, attackers can infer attributes such as income, loan amount, or credit score. A notable form of reconstruction attack is the model inversion attack \cite{fredrikson2014privacy}, where an attacker exploits a trained model to infer missing attributes. For example, if a credit approval model is trained on features such as age, employment status, and credit score, an attacker with partial knowledge of an applicant’s details can repeatedly query the model to uncover sensitive information. \raju{Another vulnerability arises from attribute inference attacks \cite{nasr2019comprehensive}, where statistical correlations (i.e., feature dependencies) in synthetic data may reveal sensitive relationships, reflecting a trade-off between preserving data utility and ensuring privacy}. If synthetic data preserves a strong link between high-income individuals and loan approvals, an attacker could deduce an applicant’s financial status with high probability. Preventing such attacks requires techniques that prevent direct memorization, introduce privacy constraints, and reduce overfitting to real data. These risks are extensively discussed in \raju{Yeom et al. \cite{yeom2018privacy}}. Another major challenge is balancing the privacy-utility trade-off. Methods such as differential privacy inject noise into data or model training to obscure individual records, but excessive noise can distort statistical relationships, making the synthetic data less useful. Conversely, prioritizing utility without sufficient privacy safeguards increases vulnerability to attacks. A well-designed synthetic data model must ensure that data remains useful while mitigating privacy risks, particularly in regulated sectors like finance, where data breaches have legal and ethical implications. The trade-offs between privacy and utility are explored in \raju{Bindschaedler et al. \cite{bindschaedler2017plausible}}.

\begin{table*}[htpb] 
\centering 
\caption{Privacy-preserving approaches address the following key challenges in synthetic data generation} 
\resizebox{\textwidth}{!}{
\begin{tabular}{lccccccccc} 
\hline 
\textbf{Challenges} & \textbf{DP-SYN} & \textbf{PATE-GAN} & \textbf{PrivBayes} & \textbf{DP-CTGAN}  & \textbf{IT-GAN} & \textbf{EHR-Safe} & \raju{\textbf{AIM}} & \raju{\textbf{PEP}} & \raju{\textbf{GEM}}  \\ 
\hline 
Prevents Direct Memorization & \checkmark & \checkmark & \checkmark & \checkmark & \checkmark & &\checkmark & \checkmark & \checkmark\\ 
\hline
Protects Against Model Inversion Attacks & \checkmark & \checkmark & \checkmark &  &  & &\checkmark & \checkmark & \checkmark \\
\hline
Handles Attribute Inference Risks & \checkmark & \checkmark & \checkmark &  &  & \checkmark & \checkmark & \checkmark & \checkmark\\
\hline
Preserves Feature Dependencies &  & \checkmark & \checkmark & \checkmark & \checkmark & \checkmark & \checkmark & \checkmark & \checkmark\\
\hline
\end{tabular} 
}
\label{tab:privacy_preservation} 
\vspace{-0.8em}
\end{table*}

\paragraph{Approaches}
Several privacy-preserving synthetic data generation models have been developed to address these challenges. Table~\ref{tab:privacy_preservation} summarizes how different privacy-preserving synthetic data generation models handle common challenges, such as preventing direct memorization, protecting against model inversion attacks, and preserving feature dependencies. 

\textbf{Differentially private synthetic data generation (DP-SYN) \cite{abay2019privacy}} preserves privacy by combining a differentially private auto-encoder (DP-Auto) with a differentially private expectation maximization (DP-EM) algorithm. As shown in Figure~\ref{fig:dpsyn}, the process begins by partitioning the credit risk data set \( D \) into \( k \) groups based on labels (e.g., ``Credit Risk\colorred{''} and ``Non-Credit Risk\colorred{''}). The privacy budget \( \epsilon \) and \( \delta \) are then allocated across these groups, ensuring that privacy is maintained within each subset. The allocation is proportional to the size of each group, so larger groups receive a slightly higher budget. Next, a private auto-encoder is trained for each group. During training, Gaussian noise \( z_{it} \sim N(0, \sigma^2 C^2) \) is added to the gradients of the model to ensure differential privacy. For example, in the case of the Credit Risk dataset, features such as `income`, `age`, and `loan amount` are encoded into a latent representation, preserving their general patterns but obscuring any individual data points. The model parameters \( w_t \) are updated with this noisy gradient, which helps prevent the model from memorizing sensitive information about individuals. The update rule for the model is:

\[
w_{t+1} = w_t - \eta \cdot \left( \frac{1}{|B_t|} \sum_{i_t} \left( \nabla \ell(w_t; \mathbf{x}_{it}) + z_{it} \right) \right),
\]
where \( \eta \) is the learning rate and \( |B_t| \) is the batch size. This process is repeated until the model converges or the privacy budget \( \epsilon_A \) is exhausted. The encoded data is then processed by the DP-EM algorithm, which identifies latent patterns and generates synthetic data with statistical properties similar to the original dataset. For example, the DP-EM algorithm synthesizes data that reflects the distribution of features like ``loan amount\colorred{''} and ``credit score\colorred{''}, ensuring that feature dependencies are preserved without revealing individual data. This step is critical to maintaining the relationships between different attributes while preventing the inference of private information. Finally, the synthetic data is decoded and aggregated to form a complete synthetic dataset, as shown in Figure~\ref{fig:dpsyn}. Throughout this process, privacy is guaranteed by tracking the cumulative privacy loss using the moments accountant method, ensuring that the overall privacy budget is not exceeded. The result is a synthetic dataset that maintains the statistical properties of the original data while providing privacy guarantees.

\textbf{Private Aggregation of Teacher Ensembles GAN (PATE-GAN) \cite{jordon2018pate}} preserves privacy by incorporating the PATE mechanism into the GAN framework, ensuring differential privacy during training. \colorred{In the standard GAN framework, \(G\) maps a noise vector \(z\) to a generated sample \(G(z)\), while \(D\) distinguishes generated samples from real data; their adversarial objective is given in Eq.~(\ref{eq:gan_loss}). PATE-GAN replaces \(D\) with a student discriminator \(S\), which is trained using noisy labels aggregated from teacher discriminators \(T_1,T_2,\dots,T_k\).} The teacher-discriminators are trained on disjoint subsets of the data, with their empirical loss defined as:
\[
L_i^T(\theta_i^T) = - \left( \sum_{u \in D_i} \log T_i(u; \theta_i^T) + \sum_{j=1}^{n} \log(1 - T_i(G(z_j); \theta_i^T)) \right),
\]
where \( D_i \) is the partitioned dataset for teacher \( T_i \), and each teacher classifies real and fake samples independently. The student-discriminator \( S \) is trained using the noisy labels from the teachers. The loss function for the student-discriminator is:
\[
L_S(\theta_S) = \sum_{j=1}^{n} r_j \log S(u_j; \theta_S) + (1 - r_j) \log(1 - S(u_j; \theta_S)),
\]
\raju{where $r_j$ are the noisy labels obtained by aggregating teacher votes with added noise through the PATE mechanism}. Finally, the generator \( G \) is trained to minimize the loss with respect to the student-discriminator:
\[
L_G(\theta_G; S) = \sum_{j=1}^{n} \log(1 - S(G(z_j; \theta_G))).
\]
This loss is similar to the original GAN generator loss Eq.(\ref{eq:gan_g_logpbly})
, but the discriminator has been replaced by the student, which provides differential privacy guarantees. By training the student with noisy teacher labels and ensuring it is exposed only to generated samples, PATE-GAN effectively prevents direct memorization, protects against model inversion attacks, handles attribute inference risks, and preserves feature dependencies in the synthetic data. \raju{Differential privacy is ensured through the PATE mechanism, and the moments accountant method is used to calculate the cumulative privacy guarantee over the full training process.}

\textbf{Privacy-Preserving Bayesian Network (PrivBayes) \cite{zhang2017privbayes} }is a synthetic data generation technique that leverages Bayesian networks to model feature dependencies while ensuring differential privacy. The method builds a Bayesian network to capture the joint distribution of features, where each node represents a feature and edges represent conditional dependencies between them. To preserve privacy, PrivBayes introduces Laplace noise to the parameters of the Bayesian network using the Laplace mechanism. The general loss function for the Bayesian network parameter estimation, including the Laplace noise, is expressed as: \[
L(\theta) = \sum_{i=1}^{n} \log P(\mathbf{x}_i | \theta) + \text{Laplace noise},
\]
\raju{where $P(\mathbf{x}_i|\theta)$ is the conditional probability for feature set $\mathbf{x}_i$, $\theta$ are the parameters of the network, and the added Laplace noise denotes a perturbation term drawn from the Laplace mechanism, calibrated to the query sensitivity and the privacy parameter $\epsilon$}. The Laplace noise is calibrated based on the sensitivity of queries made on the data and the privacy parameter \(\epsilon\), ensuring differential privacy by adding noise that prevents direct memorization and reconstruction of individual data points. This approach mitigates risks such as model inversion attacks and attribute inference, as the noisy parameters make it difficult for attackers to accurately infer sensitive attributes. The method also preserves feature dependencies, generating synthetic data that mimics the statistical properties of the original data as discussed in Section~\ref{sec:psp}. The differential privacy mechanism ensures privacy by calibrating the scale of the Laplace noise, which is inversely proportional to the sensitivity of the query and the privacy parameter \(\epsilon\), maintaining a balance between privacy protection and data utility.

\textbf{DP-CTGAN \cite{fang2022dp}} integrates differential privacy (DP) into CTGAN to generate synthetic tabular data while ensuring privacy preservation (Figure~\ref{fig:dpctgan}). The model employs a privacy accountant to track privacy loss and incorporates differential privacy into the training process. The core privacy mechanism follows the Gaussian mechanism, where noise is added to the gradients of the critic (discriminator) rather than the generator to maintain convergence stability. Given a privacy budget \((\epsilon_0, \delta_0)\), the model ensures that the total privacy loss remains within acceptable bounds. The loss function of the critic is defined as:
\begin{equation*}
    L_C = \frac{1}{s} \sum_{k=1}^{s} \left( C(\hat{r}_j^k, \text{cond}_j^k) - C(r_j^k, \text{cond}_j^k) \right) + L_{\mathrm{GP}},
\end{equation*}
where \( C(\cdot) \) represents the critic's evaluation, \( \hat{r}_j^k \) and \( r_j^k \) denote synthetic and real samples conditioned on \(\text{cond}_j^k\), and \( L_{\mathrm{GP}} \) is the gradient penalty term ensuring Lipschitz continuity. Differential privacy is enforced by adding Gaussian noise \( \xi \sim \mathcal{N}(0, \sigma_C^2 I) \) to the critic’s gradient updates:
\begin{equation*}
    \Phi_C \leftarrow \Phi_C - 0.0002 \times \text{Adam}(\nabla_{\Phi_C} (L_C + 10 L_{\mathrm{GP}} + \xi)).
\end{equation*}
The generator loss is defined as:
\begin{equation*}
    L_G = \frac{1}{m} \sum_{j=1}^{m} \text{CrossEntropy}(\hat{d}_{i^*,j}, m_{i^*}) - \frac{1}{m/s} \sum_{k=1}^{m/s} C(\hat{r}_s^k, \text{cond}_s^k),
\end{equation*}
where the first term ensures accurate generation of categorical variables using cross-entropy loss, and the second term minimizes the critic’s ability to distinguish real and synthetic samples. The generator parameters are updated as:
\begin{equation*}
    \Phi_G \leftarrow \Phi_G - 0.0002 \times \text{Adam}(\nabla_{\Phi_G} L_G).
\end{equation*}
To prevent direct memorization and model inversion attacks, DP-CTGAN ensures that only the critic interacts with real data, and the added noise limits adversaries from inferring individual data points. The privacy budget \((\epsilon, \delta)\) is updated iteratively:
\begin{equation*}
    \epsilon \leftarrow \text{query } A \text{ with } \delta_0,
\end{equation*}
where \( A \) represents the privacy accountant tracking cumulative privacy loss. Feature dependencies are preserved through mode-specific normalization and conditional sampling, ensuring balanced data generation for categorical attributes. By integrating differential privacy and leveraging CTGAN’s ability to model complex tabular data distributions, DP-CTGAN effectively generates privacy-preserving synthetic data while mitigating attribute inference risks and maintaining statistical fidelity.

CTGAN with differential privacy a potential approach for privacy-preserving synthetic data generation is adding differential privacy to the CTGAN framework. This technique helps mitigate privacy risks by ensuring that the synthetic data does not reveal private information about individuals. Although this approach offers some degree of privacy protection, its guarantees are not as robust as those provided by models like DP-SYN or PATE-GAN. Additionally, the introduced noise may hinder the model's ability to learn complex data distributions. While this method has been demonstrated with medical data, it could be extended to other domains such as credit risk modeling to protect sensitive financial information.

\textbf{The Invertible Tabular GAN (IT-GAN) \cite{lee2021invertible}} preserves privacy by integrating adversarial training from GANs with the log-density training of invertible neural networks, enabling a trade-off between synthesis quality and privacy protection. The architecture, depicted in Figure ~\ref{fig:itgan}, consists of four key data paths: the AE-path, log-density-path, GAN-path, and synthesis-path. The AE-path uses an autoencoder (AE) to transform real records \( \mathbf{x}_{\text{real}} \) into hidden representations \( h_{\text{real}} \) and reconstructs them as \( \hat{\mathbf{x}}_{\text{real}} \). The encoder \( E \) and decoder \( R \) are defined as: \[
h_{\text{real}} = E(\mathbf{x}_{\text{real}}) = \text{FC}_{1n_e}(\dots \phi(\text{FC}_{11}(\colorred{\mathbf{x}_{\text{real}}}))\dots), 
\]
\[
\hat{\mathbf{x}}_{\text{real}} = R(h_{\text{real}}) = \text{FC}_{2n_r}(\dots \phi(\text{FC}_{21}(h_{\text{real}}))\dots),
\]
where \( \phi \) is a ReLU activation, \( \text{FC} \) denotes fully connected layers, and \( n_e \) and \( n_r \) represent the number of layers in the encoder and decoder, respectively. The log-density-path leverages an invertible neural network to compute the log-density of \( h_{\text{real}} \), denoted as \( \log \hat{p}_g(h_{\text{real}}) \), which approximates the log-density of the real data. This ensures that the generated data distribution aligns closely with the real data distribution, enhancing privacy. The generator \( G \), based on Neural Ordinary Differential Equations (NODEs), is defined as:
\[
h_{\text{fake}} = z(0) + \int_0^1 f(z(t), t; \theta_g) \, dt,
\]
where \( f(z, t; \theta_g) \) is a learnable function, and \( z(0) \sim \mathcal{N}(0, 1) \) is a latent vector sampled from a unit Gaussian. The log-probability \( \log p(h_{\text{fake}}) \) is estimated using the Hutchinson estimator: \[
\log p(h_{\text{fake}}) = \log p(z(0)) - \mathbb{E}_{p(\epsilon)}\left[ \int_0^1 \epsilon^\top \frac{\partial f}{\partial z(t)} \epsilon \, dt \right].
\]

The GAN-path employs adversarial training, where the discriminator \( D \) distinguishes between \( h_{\text{real}} \) and \( h_{\text{fake}} \). The synthesis-path generates synthetic records post-training using the generator and decoder. Privacy is further reinforced by the AE's mode-specific normalization, which discretizes continuous data into Gaussian mixtures and one-hot vectors, reducing the risk of re-identification. The training algorithm optimizes the model using a combination of reconstruction loss \( L_{\text{AE}} \), WGAN-GP loss, and a log-density regularizer \( R_{\text{density}} \), defined as: \[
L_{\text{AE}} = L_{\text{Reconstruct}} + \frac{1}{2} \|h_{\text{real}}\|^2 + \|h_{\text{fake}} - \hat{h}_{\text{fake}}\|^2, 
\]
\[
R_{\text{density}} = \gamma \mathbb{E}_{x \sim p_{\text{data}}} \left[ -\log \hat{p}(E(\mathbf{x})) \right],
\]
where \( \gamma \) controls the regularization strength. 
Despite its advancements, IT-GAN has several limitations that impact its practical application. First, the loss of fine-grained details during mode-specific normalization, while effective for preserving privacy, may compromise the synthesis quality for certain applications by discarding subtle but important information in the data. Second, the scalability of the model is constrained by the computational overhead introduced by the Hutchinson estimator, which, although unbiased, increases the complexity and resource requirements, making it less suitable for very large datasets. Finally, the training stability of IT-GAN is challenged by the integration of multiple loss functions, such as \( L_{\text{AE}} \), WGAN-GP, and \( R_{\text{density}} \), which can lead to instability and necessitate careful hyperparameter tuning to achieve optimal performance. These limitations highlight the trade-offs involved in balancing privacy preservation, synthesis quality, and computational efficiency in IT-GAN.

\textbf{EHR-Safe \cite{yoon2023ehr}} is a privacy-preserving synthetic data generation model designed to generate synthetic electronic health records (EHR) while preserving the privacy of sensitive information. The methodology includes encoding and decoding features, normalizing complex distributions, conditioning adversarial training, and handling missing data. This is achieved through a deep encoder-decoder architecture, which is also adaptable to other datasets. For example, categorical features such as ``loan status\colorred{''} or ``employment type\colorred{''} are encoded into abstract representations, which do not directly expose sensitive information. The encoder-decoder mechanism is optimized to transform categorical features through the following loss function for static categorical data:
\colorred{
\[
\min_{CE^{s},CF_{1}^{s},\ldots,CF_{K}^{s}}
\sum_{k=1}^{K}
L_c\left(
CF_{k}^{s}\left[CE^{s}[s_{1}^{\mathrm{co}},\ldots,s_{K}^{\mathrm{co}}]\right],
s_{k}^{\mathrm{co}}
\right),\]}
where \( C{E}^{s} \) is the encoder for the static categorical features, \( C{F}_{k}^{s} \) is the decoder for each categorical feature, and \( {s}_{\mathrm{co}} \) represents the one-hot encoded categorical feature values (e.g., ``loan status\colorred{''} encoded as 1 for ``approved\colorred{''} or 0 for ``denied\colorred{''} in credit risk data). For temporal categorical features (such as monthly payment status over time), the following loss function is used:
\colorred{
\[\min_{CE^{t},CF_{1}^{t},\ldots,CF_{K}^{t}}
\sum_{k=1}^{K}
L_c\left(
CF_{k}^{t}\left[CE^{t}[t_{1}^{\mathrm{co}},\ldots,t_{K}^{\mathrm{co}}]\right],
t_{k}^{\mathrm{co}}
\right),\]}
where \( C{E}^{t} \) and \( C{F}_{k}^{t} \) correspond to the temporal categorical encoder and decoder, respectively, and \( {t}_{\mathrm{co}} \) represents the one-hot encoded temporal feature values. This model also uses stochastic normalization for numerical features, such as ``annual income\colorred{''} and ``loan amount\colorred{''}. This normalization step ensures that the distribution of numerical values is consistent and maintains privacy. The stochastic normalization function is defined as follows:
\[
\mathbf{X}^\text{normalized} = \text{StochasticNormalization}(\mathbf{X}),
\]
where \( \mathbf{X} \) represents a numerical feature, such as ``income\colorred{''} or ``loan amount\colorred{''}. The stochastic normalization process adjusts the feature distribution while preserving the underlying statistical properties. A key innovation of EHR-Safe is its use of adversarial training with a Wasserstein GAN with Gradient Penalty (WGAN-GP) to train the generator and discriminator. The objective of adversarial training is to generate synthetic data that is indistinguishable from real data. The adversarial loss function is:
\[
\mathbf{\max}_G \mathbf{\min}_D \frac{1}{N} \sum_{i=1}^{N} D(e[i]) - \frac{1}{N} \sum_{i=1}^{N} D(\hat{e}[i]) + \eta \left( \left|\left|\nabla_D(\hat{e}[i])\right|\right| - 1 \right)^2,
\]
where \( e[i] \) represents the real encoder state, \( \hat{e}[i] \) is the synthetic encoder state generated by the model, and \( \eta \) is the gradient penalty factor. The generator \( G \) aims to create synthetic data that matches the real data distribution, while the discriminator \( D \) tries to distinguish between real and synthetic data. In addition, EHR-Safe generates realistic missing patterns for missing data (e.g. missing ``credit score\colorred{''} values). The missing patterns are represented as:
\[
D_M = \{m_n(i), m_c(i), m_{\tau n}(i), m_{\tau c}(i)\}_{i=1}^N,
\]
where \( m_n(i) \) and \( m_c(i) \) represent missing patterns for numerical and categorical features, and \( m_{\tau n}(i) \) and \( m_{\tau c}(i) \) are the missing patterns for temporal features. These missing patterns help maintain the structural integrity of the synthetic data and avoid information leakage. Finally, in the inference phase, synthetic data is generated by sampling random vectors from a Gaussian distribution and passing them through the trained generator \( G \). The synthetic encoder states \( \hat{e} \) are then decoded to synthetic data, including categorical and numerical features and missing patterns. The final synthetic dataset is generated as follows: \[
\hat{e} = G(z), \quad \text{where } z \sim N(0, I),
\]
\[
(\hat{s}_n, \hat{s}_{\mathrm{ce}}, \hat{t}_n, \hat{t}_{\mathrm{ce}}, \hat{u}, \hat{m}_n, \hat{m}_c, \hat{m}_{\tau n}, \hat{m}_{\tau c}) = F(\hat{e}).
\] The synthetic data includes features like ``loan amount\colorred{''} (\( \hat{s}_n \)), ``loan status\colorred{''} (\( \hat{s}_c \)), ``monthly payment\colorred{''} (\( \hat{t}_n \)), and missing patterns, such as \( \hat{m}_n \) for numerical features or \( \hat{m}_c \) for categorical features. The final synthetic data output is: \[
\hat{D} = \{ \hat{s}_n(i), \hat{s}_c(i), \hat{u}, \hat{m}_n(i), \hat{m}_c(i), \hat{m}_{\tau n}(i), \hat{m}_{\tau c}(i) \}_{i=1}^{M},
\]
where \( M \) is the number of synthetic samples generated. This synthetic data is useful for tasks like credit scoring or risk assessment, while maintaining privacy and ensuring data utility. Through these mechanisms, EHR-Safe provides a robust solution for generating privacy-preserving synthetic data from different datasets, ensuring both the confidentiality of sensitive data and the utility of the generated data for analysis and modeling purposes. While the model is effective in generating privacy-preserving synthetic data, it assumes that feature distributions can be normalized to fully preserve privacy while maintaining data utility, which may not always hold true for datasets with complex dependencies or rare categorical values. Furthermore, the approach to missing data generation assumes a uniform missingness mechanism, which may not fully capture the complexities of real-world missing data patterns. Finally, the model may struggle to generate high-quality synthetic data in domains with highly heterogeneous features or extreme feature interactions, such as in healthcare or finance.

\raju{ \textbf{Query release based approaches: }
Several privacy preserving synthetic data generation methods \cite{mckenna2022aim,liu2021iterative, hardt2012simple} formulate synthesis as a query release problem, where the goal is to construct a sanitized dataset that approximately preserves the answers to a large collection of statistical queries while satisfying differential privacy. Let the sensitive dataset be denoted by $D \in \mathcal{X}^n$, and let $\mathcal{Q}$ represent a class of statistical queries, where each $q \in \mathcal{Q}$ computes a marginal or low-order statistic over $D$. To ensure differential privacy, query answers are released using a randomized mechanism $\mathcal{M}$ satisfying
\begin{equation*}
\Pr[\mathcal{M}(D) \in S] \le e^{\varepsilon} \Pr[\mathcal{M}(D') \in S],
\end{equation*}
for any neighboring datasets $D, D'$ differing in a single record and any measurable set $S$. Within this formulation, \textbf{the Adaptive Iterative Mechanism (AIM) \cite{mckenna2022aim}} improves utility by adaptively selecting a subset of informative queries rather than measuring all queries uniformly. At iteration $t$, AIM maintains a synthetic distribution $\hat{P}_t$ and selects the query that maximizes the discrepancy between the synthetic and real data,
\begin{equation*}
q_t = \arg\max_{q \in \mathcal{Q}} \left| q(\hat{P}_t) - \tilde{q}(D) \right|,
\end{equation*}
where $\tilde{q}(D)$ denotes the noisy measurement obtained via a differentially private noise mechanism. The synthetic distribution is then updated by solving
\begin{equation*}
\hat{P}_{t+1} = \arg\min_{P \in \Delta} \mathrm{KL}(P \| \hat{P}_t) \quad \text{s.t.} \quad q_t(P) = \tilde{q}_t(D),
\end{equation*}
with $\Delta$ denoting the probability simplex. By focusing the privacy budget on workload-relevant queries, AIM achieves improved utility under fixed privacy constraints. Related iterative approaches further generalize this framework. \textbf{Private Entropy Projection (PEP)\cite{liu2021iterative}} extends Multiplicative Weights Exponential Mechanism (MWEM) \cite{hardt2012simple} style algorithms by adaptively reusing previously measured noisy queries, reducing redundant privacy expenditure while enforcing multiple constraints simultaneously, and updates the synthetic distribution via entropy-regularized projection,
\begin{equation*}
\hat{P}_{t+1} = \arg\min_{P \in \Delta} \mathrm{KL}(P \| \hat{P}_t) \quad \text{s.t.} \quad q_i(P) = \tilde{q}_i(D), \;\; \forall i \le t.
\end{equation*}
Generative Networks with the Exponential Mechanism (GEM) \cite{liu2021iterative} further improve scalability by parameterizing the synthetic distribution $P_\theta$ using a neural generative model and selecting queries through the exponential mechanism,
\begin{equation*}
\Pr(q_t = q) \propto \exp\left( \frac{\varepsilon \cdot u(q, D, P_\theta)}{2 \Delta u} \right),
\end{equation*}
where $u(\cdot)$ measures the discrepancy between real and synthetic query answers. Gradient-based optimization over $\theta$ enables GEM to capture richer distributions while preserving formal differential privacy guarantees. Collectively, AIM, PEP, and GEM address the privacy challenges summarized in Table \ref{tab:privacy_preservation} by preventing direct memorization, protecting against model inversion and attribute inference attacks, and preserving feature dependencies through iterative constraint enforcement. However, these methods suffer from scalability limitations in high-dimensional settings, as the number of candidate marginal queries grows combinatorially with the number of attributes,
\begin{equation*}
|\mathcal{Q}| = \mathcal{O}\left(\sum_{k=1}^{d} \binom{d}{k}\right),
\end{equation*}
and from privacy budget fragmentation, since the total budget must be divided across iterations,
\begin{equation*}
\varepsilon = \sum_{t=1}^{T} \varepsilon_t,
\end{equation*}
which increases noise variance and leads to utility degradation under strict privacy regimes. While AIM mitigates this issue through adaptive query selection and GEM improves expressiveness via neural parameterization, performance remains sensitive to workload definition and privacy budget allocation, highlighting the inherent trade-off between strong, formal privacy guarantees and expressive modeling capacity.
}

\paragraph{Success and Limitations} 

Privacy-preserving models have demonstrated their ability to mitigate risks such as model inversion and attribute inference attacks, making synthetic data safer for sharing and analysis. Techniques such as differential privacy provide formal guaranties that limit the leakage of sensitive information, even when adversaries attempt to reverse-engineer the original data \cite{abay2019privacy}. Additionally, these models show promise in balancing privacy and utility by preserving key statistical properties and feature dependencies, enabling synthetic data to remain useful in downstream tasks like machine learning training and predictive modeling \cite{jordon2018pate}. However, these models face notable challenges. The privacy--utility trade-off remains a persistent issue, \raju{as stronger privacy guarantees typically require increased noise injection or aggregation, which can degrade data fidelity and reduce the usefulness of synthetic datasets}. High-dimensional datasets such as credit risk \cite{credit_risk_kaggle} and adult \cite{adult_2} datasets \raju{further amplify this challenge}, where achieving both strong privacy protection and high utility becomes increasingly complex \cite{papernot2016semi}.Scalability is another limitation, as models tailored to privacy preservation \raju{ particularly those relying on multiple training partitions, teacher ensembles, or iterative noise addition} sometimes struggle with large datasets or require computationally expensive mechanisms, making them less practical for real-time applications \cite{yoon2023ehr}. Furthermore, the effectiveness of some privacy-preserving techniques is highly dependent on the quality and diversity of the training data. \raju{When datasets are biased or highly imbalanced, these models can exhibit unstable training behavior, reduced utility, or unintended leakage of privacy through overfitting to dominant patterns}. Finally, many approaches lack robustness against evolving attack strategies~\cite{guan2024zero,annamalai2024linear}, which \raju{have demonstrated increasing effectiveness in extracting sensitive information from synthetic or aggregated data}, highlighting the need for stronger and more adaptive privacy guaranties.

\subsection{Conditioning on Specific Attributes}
\label{sec:csa}
\paragraph{Significance and Challenges}Conditioning in synthetic data generation enables targeted sampling \cite{goodfellow2014generative}, allowing models to generate specific subpopulations while preserving statistical properties. In Credit Risk, this is crucial for simulating borrower segments based on factors such as loan grade, interest rate, or annual income. Effective conditioning ensures that generated records remain realistic, meaning synthetic applicants conditioned on high-interest loans should exhibit higher default rates, consistent with real-world trends. This capability is essential for financial institutions aiming to assess risk under different economic conditions, model rare borrower segments, or conduct fairness-aware analyses. By generating synthetic data with controlled attributes, models can help mitigate bias \cite{chawla2002smote}, ensure equitable representation \cite{zemel2013learning}, and improve the robustness of predictive analytics \cite{xu2019modeling}.

Despite its importance, conditioning in synthetic data generation presents several challenges. One of the primary issues is ensuring statistical realism: Synthetic data conditioned on specific attributes must align with natural distributions \cite{xu2019modeling}. If a model conditions on borrowers with low annual income but generates unrealistic financial attributes, such as low debt-to-income ratios, it distorts the statistical integrity of the data. This can mislead financial risk assessment models that rely on synthetic data for training. Ensuring that conditioned samples reflect genuine feature dependencies is particularly difficult in high-dimensional datasets with complex relationships.
Another major challenge is mode collapse \cite{goodfellow2014generative}, where models repeatedly generate a limited set of similar records instead of maintaining the full diversity of the original data distribution. This is particularly common in GAN-based models, where the discriminator may favor a subset of outputs, leading to poor generalization. In a Credit Risk scenario, if a model is conditioned on high-risk borrowers, but mode collapse occurs, it may produce only a narrow subset of borrower profiles, failing to reflect the full spectrum of financial behaviors. \colorred{Fairness constraints require care because aggregate parity can obscure group-specific conditional relationships. When clinically or causally relevant feature--outcome relationships differ across groups, forcing identical distributions or outcome rates may remove information needed for valid prediction; conversely, preserving every observed difference can reproduce historical bias. The selected fairness definition must therefore be justified by the intended use and domain evidence. Evaluation should report group- and intersection-specific fidelity and utility together with the chosen fairness metric, because improving one fairness measure may reduce utility or privacy and does not establish fairness under other definitions \cite{liu2025can}.} 
\begin{table*}[htpb] 
\centering 
\caption{Key Challenges in Conditioning on attributes and the models that addresses.}
\resizebox{\textwidth}{!}{
\begin{tabular}{lcccccc} 
\hline 
\textbf{Challenges}
& \textbf{CTGAN}
& \textbf{DataSynthesizer}
& \textbf{TVAE}
& \textbf{GReaT}
& \textbf{TabFairGAN}
& \colorred{\textbf{Pred-LLM}} \\
\hline

Maintains Feature Dependencies
& \checkmark & \checkmark & \checkmark & \checkmark & &
\colorred{\checkmark} \\
\hline

Handles Mode Collapse
& & \checkmark & & & & \\
\hline

Supports Fairness Constraints
& & & & & \checkmark & \\
\hline

\colorred{Targets Statistical Fidelity}
& \checkmark & \checkmark & \checkmark & \checkmark & &
\colorred{\checkmark} \\
\hline
\end{tabular} 
}
\label{tab:cond_attrs} 
\vspace{-0.8em}
\end{table*}
\paragraph{Approaches}Several synthetic data generation models address these challenges using different conditioning mechanisms, as summarized in Table~\ref{tab:cond_attrs}. Techniques such as improved latent space sampling \cite{arjovsky2017wasserstein}, hierarchical conditioning \cite{hu2020hierarchical}, and fairness-aware adversarial training \cite{rajabi2022tabfairgan} have been proposed to enhance conditioning accuracy. Fairness-aware models introduce adversarial regularization to enforce balanced distributions across demographic attributes, reducing bias in generated data. However, the trade-off between conditioning accuracy and efficiency remains a key challenge, especially when handling large-scale financial datasets.

\textbf{CTGAN (Conditional Tabular GAN) }\cite{xu2019modeling} extends GANs for tabular data by incorporating conditional vectors and mode-specific normalization to handle categorical variables effectively. Conditional vectors enable the model to generate data conditioned on specific categories (e.g., loan grade ‘A’ in a Credit Risk dataset), while mode-specific normalization prevents overfitting to frequent categories and ensures rare categories are represented Eq.(\ref{eq:modenorm}). For example, conditioning on loan grade ‘A’ allows CTGAN to generate synthetic samples of high-credit borrowers while preserving correlations between features like income and credit utilization. However, CTGAN can struggle with mode collapse when conditioned on imbalanced categories and often requires careful hyperparameter tuning to maintain diversity and avoid training instability.

\textbf{DataSynthesizer} \cite{ping2017datasynthesizer} is a privacy-preserving tool designed to generate synthetic datasets by conditioning on specific attributes to ensure structural and statistical similarity to the original dataset. The DataDescriber module analyzes the input dataset to infer attribute distributions and correlations, adding noise for privacy. It outputs a summary \( S \), which includes marginal distributions \(\mathbf{ P(X) }\) for independent attributes and conditional distributions \(\mathbf{ P(X \mid \text{Pa}(X))} \) encoded in a Bayesian network \( N \) for correlated attributes. The DataGenerator module then samples from this summary, conditioning on attributes differently based on the mode. In independent attribute mode, attributes are sampled independently using \( \mathbf{X \sim P(X)} \) or uniformly for uniform attributes. In correlated attribute mode, attributes are sampled based on their conditional dependencies \(\mathbf{ X \sim P(X \mid \text{Pa}(X))} \). This ensures that synthetic data preserves correlations and dependencies present in the original dataset. However, the tool has limitations: the independent attribute mode cannot capture conditional relationships, while the correlated attribute mode relies heavily on the accuracy of the Bayesian network \( N \). Additionally, the tool does not explicitly handle missing data or noisy inputs, and its privacy guarantees depend on the noise added during the DataDescriber phase, which may not suffice for highly sensitive datasets. Overall, while DataSynthesizer effectively conditions on specific attributes to generate useful synthetic data, its usability comes at the cost of limited customization and potential trade-offs in accuracy and privacy.

\textbf{Tabular Variational Autoencoder (TVAE) \cite{xu2019modeling}} enables the generation of synthetic data by conditioning on specific attributes, ensuring that the generated samples align with desired characteristics. For instance, in a Credit Risk dataset, conditioning on attributes such as annual income \( > \$75,\!000 \) can guide the model to generate samples that reflect high-income borrowers. This conditioning is achieved by incorporating the conditioning signal directly into the encoder and decoder networks. Specifically, \raju{ the encoder \( q_\phi(\mathbf{z} \mid \mathbf{x}, \mathbf{c}) \) maps the input data \( \mathbf{x} \), together with condition \( \mathbf{c} \), to a latent Gaussian distribution over \( \mathbf{z} \), parameterized by mean and standard deviation.}The latent variable is then sampled using the reparameterization trick:
\[
\mathbf{z} = \mu_\phi(\mathbf{x}, \mathbf{c}) + \sigma_\phi(\mathbf{x}, \mathbf{c}) \cdot \boldsymbol{\epsilon}, \quad \boldsymbol{\epsilon} \sim \mathcal{N}(0, \mathbf{I}).
\]
Here, \( \mu_\phi(\mathbf{x}, \mathbf{c}) \) and \( \sigma_\phi(\mathbf{x}, \mathbf{c}) \) reflect how both the input and the conditioning attributes influence the structure of the latent space. By conditioning the latent representation in this way, TVAE allows the decoder \( p_\theta(\mathbf{x} \mid \mathbf{z}, \mathbf{c}) \) to generate synthetic data \( \mathbf{x}_{\text{syn}} \) that reflects the specified conditions while preserving realistic joint distributions of correlated features—such as credit utilization and debt-to-income ratio, ensuring consistency with the original dataset. However, TVAE relies on a Gaussian prior \( p(\mathbf{z}) = \mathcal{N}(0, \mathbf{I}) \) in the latent space, which can limit its ability to model highly non-linear or multi-modal distributions, particularly when the conditioning attributes are imbalanced or rare. While TVAE is computationally efficient compared to GANs, it may struggle to fully capture complex dependencies in the data, whereas GANs, though harder to train, often excel at modeling such intricacies.

\textbf{Generation of Realistic Tabular data (GReaT) \cite{borisov2022language}} leverages the power of auto-regressive generative large language models (LLMs) to generate synthetic tabular data while enabling conditioning on specific attributes. Given a tabular data row \( \mathbf{x = (x_1, x_2, \dots, x_n)} \), GReaT can condition on a subset of features \( \mathbf{x_{\text{cond}} = (x_{c_1}, x_{c_2}, \dots, x_{c_k}) }\) by encoding them as natural language prompts (e.g., ``Loan grade is A\colorred{''}) that prime the transformer's self-attention mechanism \cite{vaswani2017attention} to properly weight these known features when generating the remaining features \( \mathbf{x_{\text{gen}} = (x_{g_1}, x_{g_2}, \dots, x_{g_{n-k}}) }\) by modeling the conditional distribution:
\[
\mathbf{x_{\text{gen}} \sim p(x_{\text{gen}} | x_{\text{cond}})}.
\]
This allows GReaT to produce realistic synthetic data that preserves the relationships between conditioned and generated features. For example, in a Credit Risk dataset, conditioning on loan grade 'A' enables GReaT to generate synthetic samples with realistic distributions of income and credit utilization for high-credit borrowers. The auto-regressive nature of the LLM ensures that the generated data adheres to the statistical properties of the original dataset. However, GReaT has limitations. First, its reliance on LLMs makes it computationally expensive, particularly for large datasets or high-dimensional feature spaces. Second, while it excels at capturing feature dependencies, it may struggle with highly imbalanced datasets or rare categories, as LLMs tend to prioritize frequent patterns. Finally, GReaT requires careful tuning of hyperparameters, such as sequence length and learning rate, to achieve optimal performance, which can be time-consuming and resource-intensive. Despite these limitations, GReaT represents a significant advancement in tabular data generation, particularly for scenarios requiring fine-grained conditioning.

\colorred{A more recent LLM-based method, Pred-LLM \cite{nguyen2024generating}, extends this conditioning strategy through permutation-based fine-tuning and feature-conditional sampling. It generates feature values under the specified conditions and then queries the fine-tuned model for the target label, explicitly targeting preservation of the feature--target relationship.}

\textbf{TabFairGAN \cite{rajabi2022tabfairgan}} incorporates fairness constraints to ensure equitable outcomes across protected attributes, such as race or gender. This is achieved by enforcing demographic parity, which ensures that the probability of a positive decision $y = 1$ is equal across groups defined by the protected attribute $s$:
\[
P(y = 1 | s = 1) = P(y = 1 | s = 0),
\]
where $s = 1$ and $s = 0$ represent privileged and underprivileged groups, respectively. The model uses a discrimination score (DS) to measure fairness:
\[
DS = P(y = 1 | s = 1) - P(y = 1 | s = 0).
\]
\colorred{By minimizing \(DS\), TabFairGAN targets demographic parity for the selected protected attribute rather than guaranteeing fairness under all definitions.}. However, this fairness mechanism is not equivalent to general conditioning on specific attributes. While fairness constraints condition on protected attributes to enforce equitable treatment, they do not provide fine-grained control over the generation process for arbitrary attributes, such as income or credit score. For example, TabFairGAN cannot explicitly generate synthetic data for individuals with a specific income level or loan grade unless those attributes are incorporated into the fairness constraints.

\paragraph{Success and Limitations}
Privacy-preserving synthetic data models have shown promising results in balancing data utility and privacy, particularly in scenarios where safeguarding sensitive information is crucial. Models incorporating differential privacy mechanisms, such as DataSynthesizer, provide formal privacy guarantees, ensuring that individual records cannot be easily inferred \cite{choi2017generating}. These methods excel at mitigating privacy risks, making synthetic data more secure for public release or analysis. Additionally, models leveraging auto-regressive architectures, like GReaT, effectively capture complex feature dependencies and conditional distributions, enhancing data realism \cite{borisov2022language}. Fairness-aware techniques, such as TabFairGAN, have also contributed to reducing biases across demographic attributes, supporting more equitable data generation processes \cite{chouldechova2018frontiers}. However, these advancements come with trade-offs. Strong privacy mechanisms often degrade the quality of feature dependencies, impacting the overall utility of synthetic datasets. High-dimensional datasets amplify this challenge, as preserving privacy while maintaining statistical fidelity becomes increasingly complex. Furthermore, techniques relying on large models, like GReaT, introduce considerable computational overhead, limiting their practicality for large-scale applications. Fairness-aware approaches, while effective at bias mitigation, may reduce the accuracy of feature relationships, leading to potential loss of nuance in the synthetic data. Ultimately, the choice of privacy-preserving model depends heavily on the specific use case, balancing the need for privacy, fairness, and data utility according to the domain's unique requirements.
Subsequent to the comprehensive analysis of diverse methodologies for synthetic tabular data generation, Table \ref{tab:taxonomy} presents a succinct summary of the principal characteristics and approaches. This comparative overview elucidates the strengths and limitations of each method, predicated on their capacity to address the challenges associated with feature dependencies, statistical preservation, privacy considerations, and domain-specific requirements. The table also includes methods such as Borderline- SMOTE \cite{han2005borderline}, ADASYN \cite{he2008adasyn} and FinGAN \cite{kate2022fingan} which are explicitly designed for or commonly applied to tabular data but may not fully address all the aforementioned characteristics.
\subsection{Applications: Leveraging Domain-Specific Knowledge}
\label{sec:ldsk}
Synthetic data generation techniques are increasingly tailored to meet the needs of specific domains by incorporating domain knowledge to refine general methods. This ensures that generated datasets are not only statistically sound but also practically relevant and privacy-conscious. Domains like finance, healthcare, marketing, and social sciences each present unique challenges that demand specialized solutions.
In finance, synthetic data supports tasks like credit risk modeling, where it is essential to preserve dependencies among features like income, loan amount, and default likelihood. Models such as CTGAN \cite{xu2019modeling} effectively capture these relationships, and when integrated with Differential Privacy (DP) \cite{dwork2006differential}, protect sensitive financial information. FinGAN \cite{kate2022fingan} further addresses sector-specific needs, like customer segmentation, generating synthetic profiles that reflect real behaviors while complying with regulatory standards \cite{ghatak2022survey}.In healthcare, where data access is limited by privacy laws like HIPAA \cite{HIPAA1996}, models like medGAN \cite{choi2017generating} generate synthetic medical records by learning dependencies among clinical attributes. EHR-safe \cite{yoon2023ehr} and DP-CTGAN \cite{fang2022dp} further enhance realism and privacy, producing high-fidelity electronic health records (EHRs) while ensuring confidentiality \cite{dwork2006differential}.
In marketing, synthetic data simulates consumer behavior, aiding in personalized marketing and customer segmentation. WGAN \cite{arjovsky2017wasserstein} captures complex dependencies between demographics and purchasing behavior. Privacy-preserving methods such as DP ensure compliance while maintaining utility \cite{dwork2016calibrating}.
In the social sciences, synthetic data enables research on sensitive topics like voting behavior or education outcomes. SMOTE and VAEs are used to correct class imbalance and retain statistical properties. For instance, SMOTE has been applied in education to predict student dropouts \cite{flores2022comparison}, enabling proactive policy responses. These methods offer a balance between realism and privacy, supporting broader evidence-based research.
In sum, domain-specific adaptations of synthetic data generation models—such as FinGAN, medGAN, EHR-safe, and DP-CTGAN illustrate how general techniques are fine-tuned to preserve statistical fidelity, address class imbalance, and meet privacy requirements. These adaptations make synthetic data a viable and valuable tool across industries.

\section{Benchmarking}
\label{sec:BAC}
\subsection{Setup}
To systematically evaluate the performance of various synthetic data generation models, we implemented a comprehensive Python pipeline integrating data preprocessing, model training, synthetic data generation, and metric-based evaluation. The pipeline supports two real-world datasets: Adult \cite{adult_2} and CreditRisk \cite{credit_risk_kaggle}, each with predefined continuous, categorical, mixed-type, and target columns stored in a dictionary for modular access. We began by loading and cleaning each dataset, handling missing values, and applying \textit{LabelEncoder} to categorical features. \colorred{For JSD and membership-inference evaluation, real and synthetic categorical values were mapped to the same decoded labels.}

For synthetic data generation, our framework includes multiple models: CTGAN, FCT-GAN, CTAB-GAN, TABSYN, DP-CTGAN, and PATE-GAN, each initialized with its corresponding configuration (e.g., epochs, batch size, and differential privacy parameters like $\epsilon$ and $\delta$). These models were chosen based on their open source availability, compatibility with mixed-type tabular data, and adoption in previous work, ensuring a balanced comparison of utility, privacy, and scalability. DP-CTGAN and PATE-GAN were initialized via the \textit{snsynth} library using \textit{Synthesizer.create}, while CTGAN uses SDV's \textit{SingleTableMetadata}, and custom initializers were employed for FCT-GAN\footnote{\url{https://github.com/ethan-keller/FCT-GAN}} and CTAB-GAN\footnote{\url{https://github.com/Team-TUD/CTAB-GAN}} to pass parameters like \textit{mixed\_columns} and \textit{problem\_type}. TABSYN\footnote{\url{https://github.com/amazon-science/tabsyn}} was evaluated separately due to its distinct dependencies, with its code adapted to support the Credit Risk dataset.

Each model was trained once per dataset, and after training, 10 independent synthetic datasets of 10,000 samples each were generated per model. Evaluation statistics, mean and standard deviation, were computed across these 10 generations to ensure robustness and account for randomness in sampling. \colorred{Additionally, statistical significance for the reported comparisons was assessed using permutation-based p-value testing}. For privacy-preserving models, differential privacy budgets were fixed: PATE-GAN was trained with $\epsilon=1.0$, and DP-CTGAN with $\epsilon=2.0$. \raju{ This choice reflects model-specific stability considerations rather than an attempt to favor any method: PATE-GAN benefits from privacy amplification through teacher aggregation, allowing it to operate under tighter privacy budgets, whereas DP-CTGAN applies noise directly to gradient updates and was observed to suffer from unstable training and degenerate outputs at $\epsilon=1.0$. A slightly relaxed budget was therefore used for DP-CTGAN to ensure meaningful convergence and non-trivial synthetic data generation.} Once trained, the synthesizers generated synthetic datasets, which were evaluated using four criteria as mentioned in Table~\ref{tab:eval_metrics_comparison}: (1) \textit{feature dependency}, (2) \textit{statistical similarity metrics}, \colorred{(3) \textit{privacy assessment} through model inversion and membership-inference attacks, measured using Accuracy and AUROC}, and (4) \textit{machine learning utility}, computed via a custom \textit{SyntheticDataMetrics} module where a \textit{RandomForestClassifier} (with hyperparameters \textit{n\_estimators=100}, \textit{max\_depth=10}, \textit{random\_state=42}, and \textit{n\_jobs=-1}) or \textit{RandomForestRegressor} (with hyperparameters \textit{n\_estimators=100}, \textit{max\_depth=10}, \textit{random\_state=42}, and \textit{n\_jobs=-1}) trained on synthetic data was tested on real-data holdout splits. Each trained model was serialized using \textit{joblib} and saved for reproducibility, with the framework designed to default to GPU (CUDA) and support future extensions.

\subsection{Results}

\begin{figure}[t!]
    \centering
    
    \begin{subfigure}[t]{0.5\textwidth}
        \centering
        \includegraphics[width=0.8\linewidth]{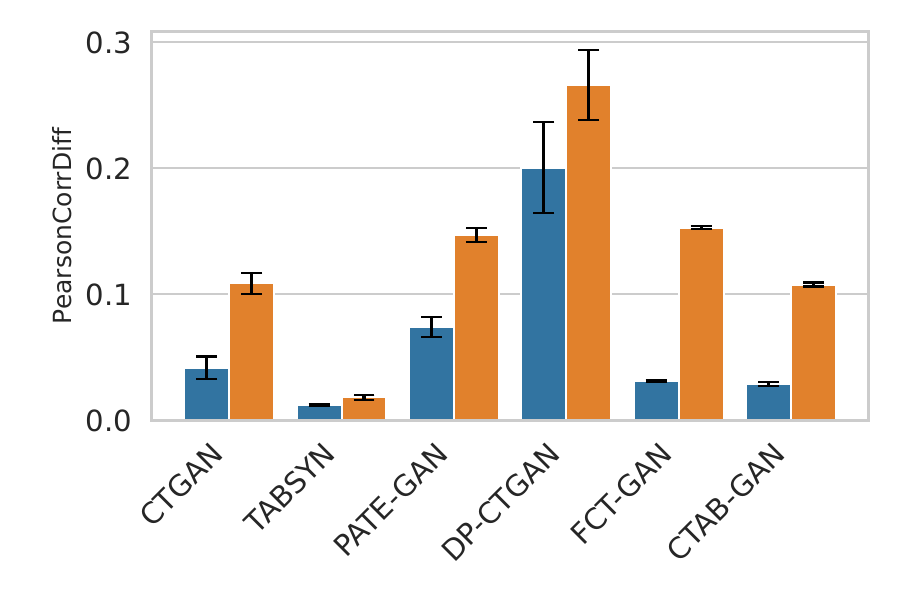}
        \vspace{-0.2cm}
        \caption{Pearson Correlation Difference}
        \label{fig:pearson}
    \end{subfigure}
    \hfill
    \begin{subfigure}[t]{0.49\textwidth}
        \centering
        \includegraphics[width=0.8\linewidth]{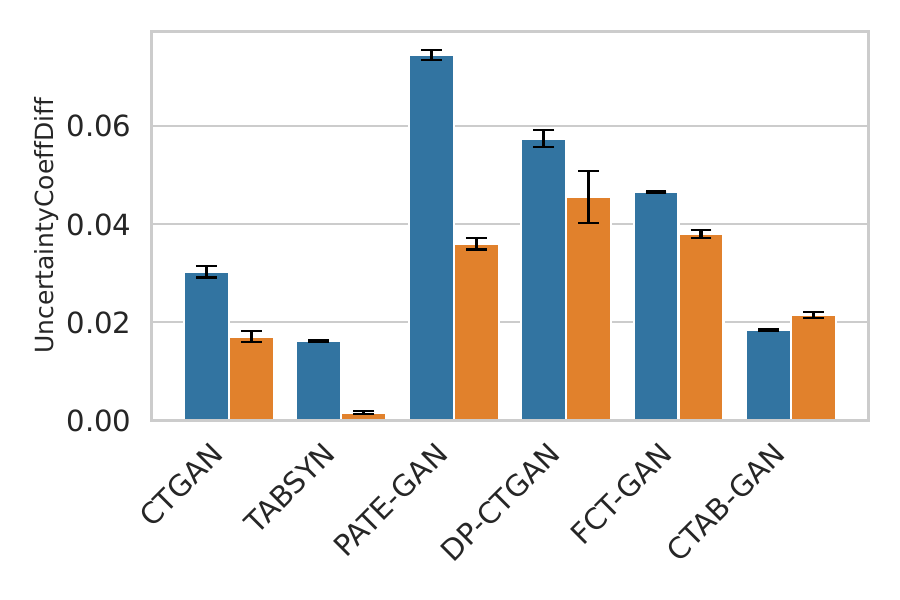}
        \vspace{-0.2cm}
        \caption{Uncertainty Coefficient Difference}
        \label{fig:uncert}
    \end{subfigure}
    
    \vspace{0.1cm}
    \begin{subfigure}[t]{0.5\textwidth}
        \centering
        \includegraphics[width=0.8\linewidth]{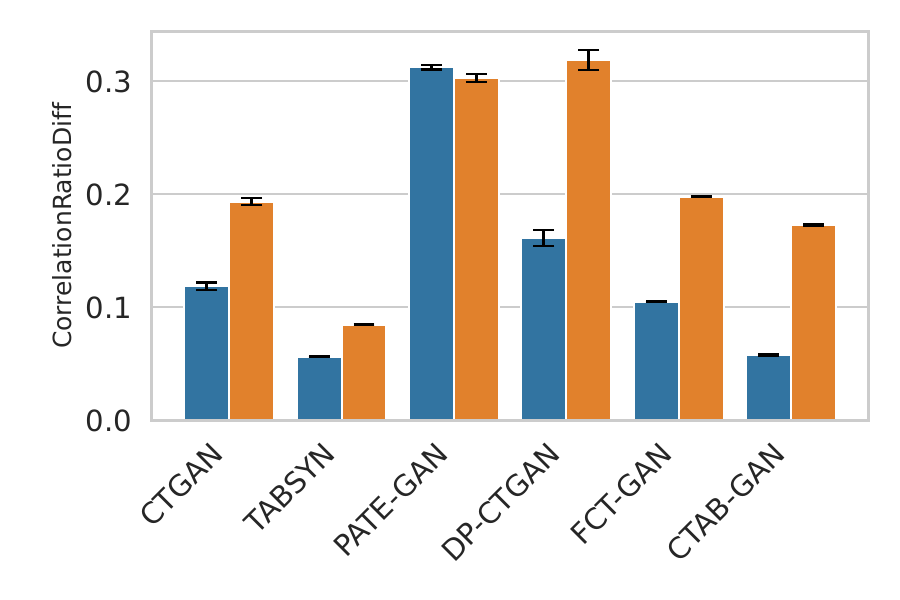}
        \vspace{-0.2cm}
        \caption{Correlation Ratio Difference}
        \label{fig:corrratio}
    \end{subfigure}
    \begin{subfigure}{0.49\textwidth}
        \centering
        \vspace{-2.5cm}
        \includegraphics[width=0.8\linewidth]{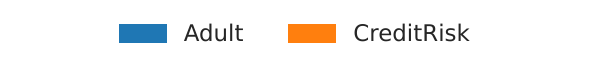}
    \end{subfigure}
    
    \vspace{-0.2cm}
    \caption{Comparison of feature dependency metrics.}
    \label{fig:fd_metrics}
    \vspace{-0.8em}
\end{figure}

\begin{figure}[t!]
    \centering
    
    \begin{subfigure}[b]{0.5\textwidth}
        \centering
        \includegraphics[width=0.8\linewidth]{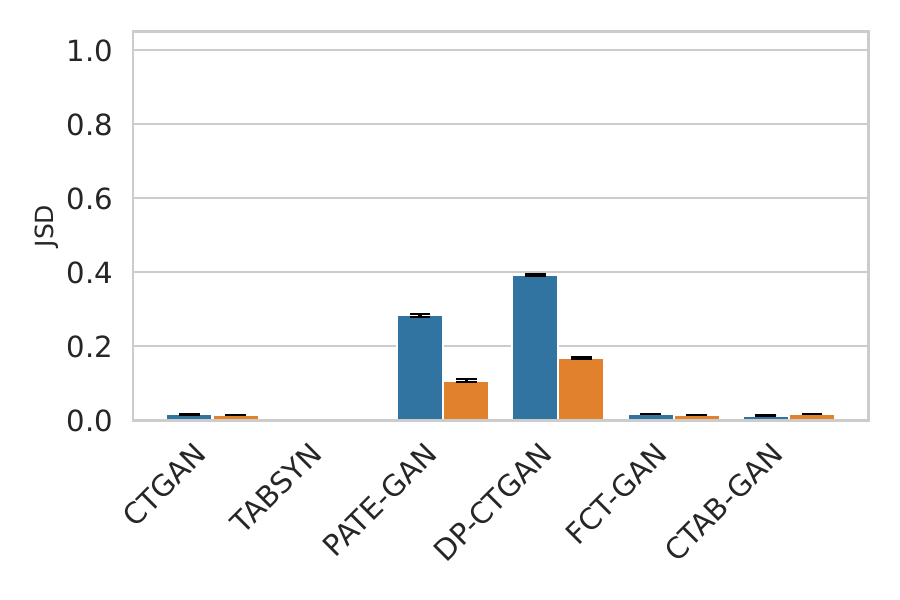}
        \vspace{-0.2cm}
        \caption{Jensen-Shannon Divergence}
        \label{fig:jsd}
    \end{subfigure}
    \hfill
    \begin{subfigure}[b]{0.49\textwidth}
        \centering
 \includegraphics[width=0.8\linewidth]{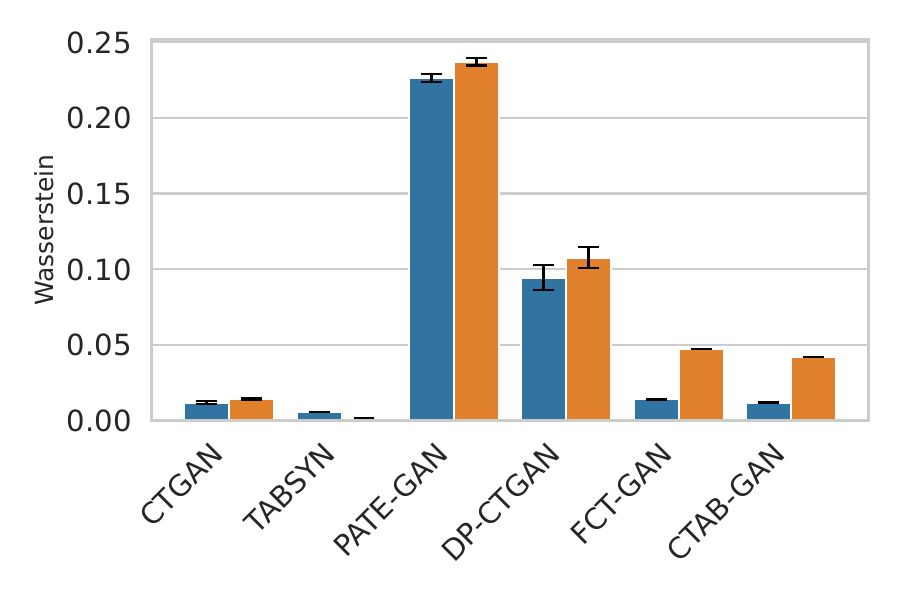}
        \vspace{-0.2cm}
        \caption{Wasserstein Distance}
        \label{fig:Wasserstein}
    \end{subfigure}
    
    \vspace{0.1cm}
    \begin{subfigure}{\textwidth}
        \centering        \includegraphics[width=0.4\textwidth]{plots/dataset_legend.pdf}
    \end{subfigure}  
    \vspace{-0.7cm}
    \caption{Comparison of statistical similarity by model and dataset.}
    \label{fig:stats_metrics}
    \vspace{-0.8em}
\end{figure}

To assess how well synthetic data retains feature dependencies, we compare real and synthetic data using three metrics: \textit{Pearson Correlation Difference} (for continuous features), \textit{Uncertainty Coefficient Difference} (for categorical features), and \textit{Correlation Ratio Difference} (for mixed feature pairs). Smaller differences indicate higher structural fidelity. Across both the Adult and CreditRisk datasets, Figure~\ref{fig:fd_metrics} shows that \textit{TABSYN} achieves the lowest deviations across all three metrics, particularly excelling on the Credit dataset with correlation differences as low as 0.012–0.056. Moreover, the associated p-values for TABSYN are above 0.09, indicating that these differences are not statistically significant and thus the synthetic data closely preserves the real data relationships. In contrast, models such as \textit{PATE-GAN} and \textit{DP-CTGAN} exhibit substantially larger correlation differences up to 0.27 for Pearson's correlation and 0.32 for correlation ratio differences with highly significant p-values ($p=0.002$). This demonstrates statistically robust distortions likely introduced by the noise added for privacy guarantees. These models also show greater variability (standard deviation), particularly DP-CTGAN, indicating inconsistent synthetic data quality across runs. Models like \textit{CTGAN}, \textit{FCT-GAN}, and \textit{CTAB-GAN} generally achieve moderate fidelity with smaller but statistically significant distortions, reflected by correlation differences around 0.03–0.11 and low variance.


We further evaluate distributional similarity using \textit{Jensen-Shannon Divergence (JSD)} for categorical features and \textit{Wasserstein Distance} for continuous features. \colorred{Figure~\ref{fig:stats_metrics} shows that \textit{TABSYN} attains the lowest JSD values, with $0.0019\pm0.0002$ on Adult and $0.000172\pm0.000051$ on CreditRisk, while its Wasserstein distance remains below 0.006. The corresponding differences are non-significant ($p>0.09$), indicating close marginal alignment between the real and synthetic distributions. \textit{FCT-GAN} obtains JSD values of $0.0178\pm0.0006$ on Adult and $0.0149\pm0.0003$ on CreditRisk, while \textit{CTAB-GAN} obtains $0.0135\pm0.0004$ and $0.0182\pm0.0003$, respectively. Although these differences are statistically significant ($p=0.002$), their magnitudes remain below 0.02 and therefore do not indicate completely disjoint categorical distributions. Overall, TABSYN provides the closest distributional match, while FCT-GAN and CTAB-GAN exhibit small but statistically detectable marginal differences.}

\begin{table*}[htbp]
\centering
\captionsetup{justification=centering}
\caption{\colorred{Privacy evaluation results. Higher values correspond to greater attack success.}}
\footnotesize
\setlength{\tabcolsep}{3pt}

\begin{tabular}{l|cc|cc|cc}
\hline
\multirow{2}{*}{Model} 
& \multicolumn{4}{c|}{Model Inversion Attack} 
& \multicolumn{2}{c}{Membership Inference Attack} \\
\cline{2-7}
& \multicolumn{2}{c|}{Accuracy} 
& \multicolumn{2}{c|}{AUROC} 
& \multicolumn{2}{c}{AUROC} \\
\cline{2-7}
& Adult & Credit 
& Adult & Credit 
& Adult & Credit \\
\hline

CTGAN
& \raju{0.810$\pm$0.006} & \raju{0.834$\pm$0.006}
& \raju{0.851$\pm$0.007} & \raju{0.833$\pm$0.010}
& \colorred{0.503$\pm$0.002} & \colorred{0.507$\pm$0.002} \\

TABSYN
& \raju{0.840$\pm$0.003} & \raju{0.911$\pm$0.001}
& \raju{0.891$\pm$0.001} & \raju{0.903$\pm$0.001}
& \colorred{0.502$\pm$0.003} & \colorred{0.509$\pm$0.005} \\

PATE-GAN
& 0.433$\pm$0.113 & 0.602$\pm$0.065
& 0.516$\pm$0.097 & 0.511$\pm$0.071
& 0.499$\pm$0.002 & 0.506$\pm$0.003 \\

DP-CTGAN
& \raju{0.761$\pm$0.003} & \raju{0.772$\pm$0.025}
& 0.542$\pm$0.090 & \raju{0.705$\pm$0.043}
& 0.495$\pm$0.002 & 0.500$\pm$0.002 \\

FCT-GAN
& \raju{0.792$\pm$0.005} & \raju{0.786$\pm$0.002}
& \raju{0.745$\pm$0.032} & \raju{0.690$\pm$0.012}
& \colorred{0.502$\pm$0.003} & \colorred{0.502$\pm$0.003} \\

CTAB-GAN
& \raju{0.795$\pm$0.006} & \raju{0.786$\pm$0.002}
& \raju{0.814$\pm$0.082} & \raju{0.771$\pm$0.009}
& \colorred{0.503$\pm$0.003} & \colorred{0.505$\pm$0.004} \\

\hline
\end{tabular}
\label{tab:prmt} 
\end{table*}

\colorred{Privacy risk is evaluated through model inversion and membership inference, which measure different types of disclosure. Table~\ref{tab:prmt} presents the numerical results, and Figure~\ref{fig:pmc} shows the corresponding trends. In this benchmark, model inversion evaluates whether an attack model trained on synthetic data can infer the target attribute of real holdout records from their remaining attributes. TABSYN obtains the highest model-inversion AUROC, reaching 0.891 on Adult and 0.903 on Credit Risk, while CTGAN obtains 0.851 and 0.833, respectively ($p=0.002$). These values indicate that the target attribute remains predictable from relationships represented in the synthetic data; they do not show whether a specific record was used to train the generator. Membership inference addresses that separate question. Its mean AUROC ranges from 0.495 to 0.509 across the evaluated models, with all results close to the chance-level value of 0.5. Therefore, the difference between the two attacks is not contradictory: model inversion measures target-attribute predictability, whereas membership inference measures the exposure of individual training records. These empirical results do not constitute formal privacy guarantees. We report AUROC as the primary MIA metric because it evaluates attack ranking across thresholds, whereas accuracy depends on a single threshold.}

\begin{table*}[htbp]
\centering
\caption{Machine Learning Utility Metrics by model and dataset.}
\footnotesize
\setlength{\tabcolsep}{2.5pt}

\begin{tabular}{l|cc|cc|cc|cc}
\hline
\multirow{2}{*}{Model}
& \multicolumn{2}{c|}{F1-Score}
& \multicolumn{2}{c|}{AUROC}
& \multicolumn{2}{c|}{AUPRC}
& \multicolumn{2}{c}{Accuracy} \\
\cline{2-9}
& Adult & Credit
& Adult & Credit
& Adult & Credit
& Adult & Credit \\
\hline

CTGAN
& 0.801$\pm$0.009 & 0.823$\pm$0.007
& 0.856$\pm$0.007 & 0.828$\pm$0.010
& 0.631$\pm$0.020 & 0.621$\pm$0.018
& 0.816$\pm$0.006 & 0.829$\pm$0.006 \\

TABSYN
& 0.838$\pm$0.002 & 0.905$\pm$0.002
& 0.890$\pm$0.001 & 0.903$\pm$0.001
& 0.742$\pm$0.003 & 0.838$\pm$0.002
& 0.840$\pm$0.002 & 0.910$\pm$0.002 \\

PATE-GAN
& 0.434$\pm$0.151 & 0.625$\pm$0.049
& 0.514$\pm$0.097 & 0.515$\pm$0.071
& 0.278$\pm$0.071 & 0.231$\pm$0.041
& 0.429$\pm$0.113 & 0.604$\pm$0.065 \\

DP-CTGAN
& 0.666$\pm$0.001 & 0.756$\pm$0.027
& 0.543$\pm$0.090 & 0.704$\pm$0.043
& 0.295$\pm$0.059 & 0.422$\pm$0.068
& 0.767$\pm$0.003 & 0.769$\pm$0.025 \\

FCT-GAN
& 0.743$\pm$0.013 & 0.693$\pm$0.002
& 0.709$\pm$0.032 & 0.682$\pm$0.012
& 0.489$\pm$0.024 & 0.380$\pm$0.017
& 0.799$\pm$0.005 & 0.783$\pm$0.002 \\

CTAB-GAN
& 0.748$\pm$0.017 & 0.693$\pm$0.005
& 0.685$\pm$0.082 & 0.772$\pm$0.009
& 0.494$\pm$0.040 & 0.497$\pm$0.013
& 0.801$\pm$0.006 & 0.784$\pm$0.002 \\

\hline
\end{tabular}
\label{tab:mlut}
\end{table*}
To quantify downstream utility, classifiers trained on synthetic data are evaluated on real data using \textit{F1-score}, \textit{AUROC}, \textit{AUPRC}, and \textit{Accuracy}. The results are reported in Table~\ref{tab:mlut}, while Figure ~\ref{fig:mlu} provides a visual comparison across models and datasets. \textit{CTAB-GAN} consistently delivers strong performance across all metrics and datasets, achieving F1-scores around 0.75 and Accuracy near 0.80, with statistically significant differences ($p=0.002$) relative to baseline. \textit{FCT-GAN} also performs well on the Adult dataset but shows moderate utility decline on CreditRisk. Although \textit{TABSYN} attains the highest F1-scores (up to 0.90), its AUROC and AUPRC metrics decline notably on Adult, with p-values near 0.09 indicating these differences are not significant but suggest a tendency toward overfitting or reduced ranking robustness. On the other hand, \textit{DP-CTGAN} and \textit{PATE-GAN} suffer from substantially degraded utility across both datasets (e.g., Accuracy as low as 0.43 and AUROC near 0.51 on Adult), consistently with significant p-values ($p=0.002$). This underscores the classical privacy-utility tradeoff, where privacy preservation reduces the usefulness of synthetic data for predictive modeling. Additionally, these models show higher performance variability, reflecting unstable training dynamics under differential privacy constraints.

Overall, model performance varies markedly between datasets. The CreditRisk dataset, being more complex with higher feature dimensionality and imbalance, presents greater challenges for synthetic data generation. This results in larger fidelity and utility losses, particularly for privacy-preserving models. Therefore, model selection should be informed by dataset characteristics and the desired balance between privacy, fidelity, and utility. The full experimental pipeline and code are publicly available at \url{https://github.com/Raju-Challagundla/syndataeval}.

\section{Challenges \& Limitations}
\label{sec:CL}
While significant progress has been made in the generation of synthetic data for tabular data, several challenges remain unresolved. A primary concern is balancing privacy guarantees with data utility. Techniques such as differential privacy and secure multi-party computation aim to preserve privacy but do not entirely eliminate the risk of model inversion attacks. These attacks enable adversaries to infer sensitive information, particularly when synthetic data closely approximates the original distribution. In regulated industries like finance and healthcare, where privacy laws such as GDPR and HIPAA mandate strict data handling protocols, achieving an optimal trade-off between privacy and utility is especially challenging. Frequent policy changes and data deletion requests further complicate dataset stability, leading to incomplete or inconsistent data. This instability degrades machine learning model performance, often resulting in biased or unrealistic predictions.

Another critical limitation is the generalizability of synthetic data models across diverse datasets. Real-world data spans a wide range of applications, each with unique statistical properties, feature interactions, and domain-specific characteristics. Models that perform well on one dataset often struggle with others, raising concerns about the scalability of existing methods. As industries evolve and incorporate new data types, ensuring that synthetic data models remain adaptable is an ongoing challenge.

A more complex issue is the accurate preservation of feature dependencies. Many domains involve intricate relationships between variables, such as correlations, temporal patterns, or hierarchical structures. High-dimensional or sequential data further complicates dependency modeling, as inaccuracies can lead to unrealistic synthetic outputs. This undermines the reliability of synthetic data for downstream tasks such as predictive modeling and decision making. In addition, imbalanced data sets, particularly those where rare events are critical, pose significant difficulties. The failure to capture these rare but important occurrences can render synthetic data ineffective in real-world applications.
\colorred{Omitted stratifying variables can change or reverse aggregate associations, as illustrated by Simpson's paradox \cite{bickel1975sex}. In the Berkeley admissions data, department-level application patterns explained much of the aggregate association with sex. A generator cannot recover an unrecorded variable or its conditional structure without additional information or assumptions and may therefore reproduce a misleading association involving a protected attribute. Dependency and fairness should be evaluated across relevant observed subgroups, with claims limited to recorded variables.}

\colorred{Rare-event synthesis is also affected by label noise: some recorded fraud negatives may be undetected positives. A generator cannot identify these cases from observed labels alone, and increasing label noise reduces synthetic-data quality and downstream utility \cite{kim2025empirical}. Verified labels should be corrected when available; if the objective is to reproduce an annotation process or test robustness, label uncertainty should be modeled explicitly and evaluated using audit labels or controlled noise. Evaluation should report rare-class coverage, class-conditional utility, and sensitivity to label noise rather than aggregate fidelity alone.} Ensuring statistical fidelity and realism remains a central challenge. \colorred{Statistical fidelity and downstream utility are distinct because reproducing all observed properties may preserve measurement errors, label noise, or spurious associations. Task-guided generation can prioritize relationships useful for a specified task \cite{jia2024tabular}, but may reduce distributional fidelity or subgroup performance and cannot identify erroneous patterns without external evidence. It should therefore be evaluated on real data excluded from generator training and tuning, with utility, fidelity, fairness, and privacy reported separately.}

While advanced generative models, including GANs, VAEs, and diffusion models, have made notable progress, they still struggle to fully replicate the nuanced distributions and dependencies found in real-world data. Achieving realism requires models to capture complex variable relationships without overfitting to specific datasets. \colorred{Population distributions can also change over time, independently of within-record temporal dependence. Seasonal behavior and longer-term drift may alter features and their relationships with outcomes. A generator trained on a static snapshot represents only that period and can reproduce recurring variation only when time is explicit and the training data cover it; otherwise, target-period observations or scenario assumptions are required. Because static similarity may conceal failures in cross-sectional and trajectory dynamics \cite{kwon2026generative}, deployment requires time-aware evaluation, drift monitoring, and updating when new observations arrive.} The evaluation of synthetic data also presents a significant limitation. Currently, there is no consensus on comprehensive, domain-specific metrics for assessing synthetic data quality. Traditional measures, such as distributional similarity or diversity, may not fully reflect the utility of synthetic data in high-stakes applications. The absence of standardized evaluation frameworks makes it difficult to determine whether synthetic data can reliably replace real data in critical tasks. 

Lastly, the interpretability of generative models remains a barrier to adoption. Many state-of-the-art models, such as GANs and VAEs, operate as black boxes, making it difficult for practitioners to understand the underlying generation process. In industries where decisions have substantial consequences, stakeholders may be hesitant to trust synthetic data produced by opaque systems.
\bibliographystyle{ACM-Reference-Format}
\bibliography{main}

@String{Computing = "Computing" }

@String{Computer = "{IEEE} Computer" }

@String{Springer = "Springer-Verlag" }

@book{bishop2006pattern,
  title={Pattern recognition and machine learning},
  author={Bishop, Christopher M and Nasrabadi, Nasser M},
  volume={4},
  number={4},
  year={2006},
  publisher={Springer}
}

@article{goodfellow2014generative,
  title={Generative adversarial nets},
  author={Goodfellow, Ian and Pouget-Abadie, Jean and Mirza, Mehdi and Xu, Bing and Warde-Farley, David and Ozair, Sherjil and Courville, Aaron and Bengio, Yoshua},
  journal={Advances in neural information processing systems},
  volume={27},
  year={2014}
}

@article{reynolds2009gaussian,
  title={Gaussian mixture models.},
  author={Reynolds, Douglas A and others},
  journal={Encyclopedia of biometrics},
  volume={741},
  number={659-663},
  year={2009},
  publisher={Berlin, Springer}
}

@article{rabiner1989tutorial,
  title={A tutorial on hidden Markov models and selected applications in speech recognition},
  author={Rabiner, Lawrence R},
  journal={Proceedings of the IEEE},
  volume={77},
  number={2},
  pages={257--286},
  year={1989},
  publisher={Ieee}
}

@article{kingma2013auto,
  title={Auto-encoding variational bayes},
  author={Kingma, Diederik P and Welling, Max},
  journal={arXiv preprint arXiv:1312.6114},
  year={2013}
}

@inproceedings{zhao2021ctab,
  title={Ctab-gan: Effective table data synthesizing},
  author={Zhao, Zilong and Kunar, Aditya and Birke, Robert and Chen, Lydia Y},
  booktitle={Asian Conference on Machine Learning},
  pages={97--112},
  year={2021},
  organization={PMLR}
}

@article{zhao2022fct,
  title={FCT-GAN: Enhancing Table Synthesis via Fourier Transform},
  author={Zhao, Zilong and Birke, Robert and Chen, Lydia Y},
  journal={arXiv preprint arXiv:2210.06239},
  year={2022}
}

@inproceedings{kotelnikov2023tabddpm,
  title={Tabddpm: Modelling tabular data with diffusion models},
  author={Kotelnikov, Akim and Baranchuk, Dmitry and Rubachev, Ivan and Babenko, Artem},
  booktitle={International Conference on Machine Learning},
  pages={17564--17579},
  year={2023},
  organization={PMLR}
}

@article{vaswani2017attention,
  title={Attention is all you need},
  author={Vaswani, Ashish and Shazeer, Noam and Parmar, Niki and Uszkoreit, Jakob and Jones, Llion and Gomez, Aidan N and Kaiser, {\L}ukasz and Polosukhin, Illia},
  journal={Advances in neural information processing systems},
  volume={30},
  year={2017}
}

@article{xu2019modeling,
  title={Modeling tabular data using conditional gan},
  author={Xu, Lei and Skoularidou, Maria and Cuesta-Infante, Alfredo and Veeramachaneni, Kalyan},
  journal={Advances in neural information processing systems},
  volume={32},
  year={2019}
}

@inproceedings{ping2017datasynthesizer,
  title={Datasynthesizer: Privacy-preserving synthetic datasets},
  author={Ping, Haoyue and Stoyanovich, Julia and Howe, Bill},
  booktitle={Proceedings of the 29th International Conference on Scientific and Statistical Database Management},
  pages={1--5},
  year={2017}
}

@article{xu2018synthesizing,
  title={Synthesizing tabular data using generative adversarial networks},
  author={Xu, Lei and Veeramachaneni, Kalyan},
  journal={arXiv preprint arXiv:1811.11264},
  year={2018}
}

@article{lee2021invertible,
  title={Invertible tabular GANs: Killing two birds with one stone for tabular data synthesis},
  author={Lee, Jaehoon and Hyeong, Jihyeon and Jeon, Jinsung and Park, Noseong and Cho, Jihoon},
  journal={Advances in Neural Information Processing Systems},
  volume={34},
  pages={4263--4273},
  year={2021}
}

@article{zhang2017privbayes,
  title={Privbayes: Private data release via bayesian networks},
  author={Zhang, Jun and Cormode, Graham and Procopiuc, Cecilia M and Srivastava, Divesh and Xiao, Xiaokui},
  journal={ACM Transactions on Database Systems (TODS)},
  volume={42},
  number={4},
  pages={1--41},
  year={2017},
  publisher={ACM New York, NY, USA}
}

@inproceedings{jordon2018pate,
  title={PATE-GAN: Generating synthetic data with differential privacy guarantees},
  author={Jordon, James and Yoon, Jinsung and Van Der Schaar, Mihaela},
  booktitle={International conference on learning representations},
  year={2018}
}

@article{avino2018generating,
  title={Generating synthetic but plausible healthcare record datasets},
  author={Avi{\~n}{\'o}, Laura and Ruffini, Matteo and Gavald{\`a}, Ricard},
  journal={arXiv preprint arXiv:1807.01514},
  year={2018}
}

@article{park2018data,
  title={Data synthesis based on generative adversarial networks},
  author={Park, Noseong and Mohammadi, Mahmoud and Gorde, Kshitij and Jajodia, Sushil and Park, Hongkyu and Kim, Youngmin},
  journal={arXiv preprint arXiv:1806.03384},
  year={2018}
}

@article{figueira2022survey,
  title={Survey on synthetic data generation, evaluation methods and GANs},
  author={Figueira, Alvaro and Vaz, Bruno},
  journal={Mathematics},
  volume={10},
  number={15},
  pages={2733},
  year={2022},
  publisher={MDPI}
}

@article{kamthe2021copula,
  title={Copula flows for synthetic data generation},
  author={Kamthe, Sanket and Assefa, Samuel and Deisenroth, Marc},
  journal={arXiv preprint arXiv:2101.00598},
  year={2021}
}

@inproceedings{karras2019style,
  title={A style-based generator architecture for generative adversarial networks},
  author={Karras, Tero and Laine, Samuli and Aila, Timo},
  booktitle={Proceedings of the IEEE/CVF conference on computer vision and pattern recognition},
  pages={4401--4410},
  year={2019}
}

@article{devlin2018bert,
  title={Bert: Pre-training of deep bidirectional transformers for language understanding},
  author={Devlin, Jacob},
  journal={arXiv preprint arXiv:1810.04805},
  year={2018}
}

@article{yoon2019time,
  title={Time-Series Generative Adversarial Networks},
  author={Yoon, Jinsung and Zame, William and van der Schaar, Mihaela},
  journal={Proceedings of the 36th International Conference on Machine Learning},
  volume={97},
  pages={5614--5623},
  year={2019},
  publisher={PMLR}
}

@article{ho2020denoising,
  title={Denoising diffusion probabilistic models},
  author={Ho, Jonathan and Jain, Ajay and Abbeel, Pieter},
  journal={Advances in neural information processing systems},
  volume={33},
  pages={6840--6851},
  year={2020}
}

@inproceedings{patki2016synthetic,
  title={The synthetic data vault},
  author={Patki, Neha and Wedge, Roy and Veeramachaneni, Kalyan},
  booktitle={2016 IEEE international conference on data science and advanced analytics (DSAA)},
  pages={399--410},
  year={2016},
  organization={IEEE}
}

@article{zhao2019logistics,
  title={Logistical challenges in data collection for machine learning},
  author={Zhao, Wei and Zhang, Li},
  journal={IEEE Transactions on Big Data},
  volume={5},
  number={3},
  pages={345--356},
  year={2019},
  publisher={IEEE}
}

@article{lu2023machine,
  title={Machine learning for synthetic data generation: a review},
  author={Lu, Yingzhou and Shen, Minjie and Wang, Huazheng and Wang, Xiao and van Rechem, Capucine and Wei, Wenqi},
  journal={arXiv preprint arXiv:2302.04062},
  year={2023}
}

@online{Amplitude2024,
  title={What Is Data Democratization? Definition and Principles},
  author={Amplitude},
  year={2024},
  journal={Amplitude},
  url={https://www.amplitude.com/blog/data-democratization}
}

@article{he2009learning,
  title={Learning from imbalanced data},
  author={He, Haibo and Garcia, Edward A},
  journal={IEEE Transactions on Knowledge and Data Engineering},
  volume={21},
  number={9},
  pages={1263--1284},
  year={2009},
  publisher={IEEE}
}

@article{chawla2002smote,
  title={SMOTE: Synthetic Minority Over-sampling Technique},
  author={Chawla, Nitesh V and Bowyer, Kevin W and Hall, Louis O and Kegelmeyer, Wayne P},
  journal={Journal of Artificial Intelligence Research},
  volume={16},
  pages={321--357},
  year={2002},
  publisher={AI Access Foundation}
}

@article{fookes2020bayesian,
  title={Bayesian Neural Networks: An Introduction and Survey},
  author={Goan, Ethan and Fookes, Clinton},
  journal={arXiv preprint arXiv:2006.12024},
  year={2020}
}

@article{hernandez2022synthetic,
  title={Synthetic data generation for tabular health records: A systematic review},
  author={Hernandez, Mikel and Epelde, Gorka and Alberdi, Ane and Cilla, Rodrigo and Rankin, Debbie},
  journal={Neurocomputing},
  volume={493},
  pages={28--45},
  year={2022},
  publisher={Elsevier}
}

@inproceedings{dosovitskiy2017carla,
  title={CARLA: An open urban driving simulator},
  author={Dosovitskiy, Alexey and Ros, German and Codevilla, Felipe and Lopez, Antonio and Koltun, Vladlen},
  booktitle={Conference on robot learning},
  pages={1--16},
  year={2017},
  organization={PMLR}
}

@inproceedings{assefa2020generating,
  title={Generating synthetic data in finance: opportunities, challenges and pitfalls},
  author={Assefa, Samuel A and Dervovic, Danial and Mahfouz, Mahmoud and Tillman, Robert E and Reddy, Prashant and Veloso, Manuela},
  booktitle={Proceedings of the First ACM International Conference on AI in Finance},
  pages={1--8},
  year={2020}
}

@article{vishweswar2024big,
  title={Big Data Analytics in Finance: Predictive Modeling for Investment Strategies},
  author={Vishweswar Sastry, V. N. and Guruprasad Desai, D. R. and Hemanth Kumar and Manjushree M.},
  journal={European Economic Letters (EEL)},
  volume={14},
  number={3},
  pages={572--581},
  year={2024},
  month={Aug.},
  url={https://eelet.org.uk/index.php/journal/article/view/1803},
  abstractNote={The purpose of this paper is to examine the use of big data analytics in finance based on case studies, focusing on predictive modeling for financial investment strategies. It addresses the importance of big data in revolutionizing established financial processes and explores predictive modeling methods and their implications for investment decisions. The paper also discusses challenges of big data analytics in finance and provides scope for future research.}
}

@article{clim2019big,
  title={Big Data in home healthcare: A new frontier in personalized medicine. Medical emergency services and prediction of hypertension risks},
  author={Clim, Antonio and Zota, Razvan Daniel and Tinica, Grigore},
  journal={International Journal of Healthcare Management},
  volume={12},
  number={3},
  pages={241--249},
  year={2019},
  publisher={Taylor \& Francis}
}

@misc{HIPAA1996,
  title = {Health Insurance Portability and Accountability Act of 1996 (HIPAA)},
  author = {U.S. Department of Health and Human Services},
  year = {1996},
  journal = {Public Law 104-191},
  url = {https://www.hhs.gov/hipaa}
}

@article{babbar2019data,
  title={Data scarcity, robustness and extreme multi-label classification},
  author={Babbar, Rohit and Sch{\"o}lkopf, Bernhard},
  journal={Machine Learning},
  volume={108},
  number={8},
  pages={1329--1351},
  year={2019},
  publisher={Springer}
}

@misc{credit_risk_kaggle,
  author       = {Kaggle},
  title        = {Credit Risk Dataset},
  year         = {n.d.},
  url          = {https://www.kaggle.com/datasets/laotse/credit-risk-dataset},
  note         = {Accessed: 2025-01-28}
}

@misc{european_union_2016_gdpr,
  title        = {General Data Protection Regulation},
  author       = {{European Union}},
  year         = {2016},
  journal      = {Official Journal of the European Union},
  volume       = {49},
  pages        = {L119},
  url          = {https://gdpr-info.eu},
  note         = {Retrieved from https://gdpr-info.eu}
}

@article{rahman2023data,
  title={Editorial: Data-driven technologies for future healthcare systems},
  author={Rahman, Md. Arafatur and Moayedikia, Alireza and Wiil, Uffe Kock},
  journal={Frontiers in Medical Technology},
  volume={5},
  pages={1183687},
  year={2023},
  doi={10.3389/fmedt.2023.1183687},
  url={https://doi.org/10.3389/fmedt.2023.1183687}
}

@article{tian2021data,
  title={Data-driven approaches in FinTech: a survey},
  author={Tian, Xin and He, Jing Selena and Han, Meng},
  journal={Information Discovery and Delivery},
  volume={49},
  number={2},
  pages={123--135},
  year={2021},
  publisher={Emerald Publishing Limited}
}

@book{arthur2013big,
  title={Big data marketing: engage your customers more effectively and drive value},
  author={Arthur, Lisa},
  year={2013},
  publisher={John Wiley \& Sons}
}

@article{zhang2020data,
  title={Data-driven computational social science: A survey},
  author={Zhang, Jun and Wang, Wei and Xia, Feng and Lin, Yu-Ru and Tong, Hanghang},
  journal={Big Data Research},
  volume={21},
  pages={100145},
  year={2020},
  publisher={Elsevier}
}

@article{hoogeboom2021autoregressive,
  title={Autoregressive diffusion models},
  author={Hoogeboom, Emiel and Gritsenko, Alexey A and Bastings, Jasmijn and Poole, Ben and Berg, Rianne van den and Salimans, Tim},
  journal={arXiv preprint arXiv:2110.02037},
  year={2021}
}

@book{nelsen2006introduction,
  title={An introduction to copulas},
  author={Nelsen, Roger B},
  year={2006},
  publisher={Springer}
}

@article{hoogeboom2021argmax,
  title={Argmax flows and multinomial diffusion: Learning categorical distributions},
  author={Hoogeboom, Emiel and Nielsen, Didrik and Jaini, Priyank and Forr{\'e}, Patrick and Welling, Max},
  journal={Advances in Neural Information Processing Systems},
  volume={34},
  pages={12454--12465},
  year={2021}
}

@article{douzas2019geometric,
  title={Geometric SMOTE a geometrically enhanced drop-in replacement for SMOTE},
  author={Douzas, Georgios and Bacao, Fernando},
  journal={Information sciences},
  volume={501},
  pages={118--135},
  year={2019},
  publisher={Elsevier}
}

@inproceedings{choi2017generating,
  title={Generating multi-label discrete patient records using generative adversarial networks},
  author={Choi, Edward and Biswal, Siddharth and Malin, Bradley and Duke, Jon and Stewart, Walter F and Sun, Jimeng},
  booktitle={Machine learning for healthcare conference},
  pages={286--305},
  year={2017},
  organization={PMLR}
}

@article{dwork2016calibrating,
  title={Calibrating noise to sensitivity in private data analysis},
  author={Dwork, Cynthia and McSherry, Frank and Nissim, Kobbi and Smith, Adam},
  journal={Journal of Privacy and Confidentiality},
  volume={7},
  number={3},
  pages={17--51},
  year={2016}
}

@inproceedings{dwork2006differential,
  title={Differential privacy},
  author={Dwork, Cynthia},
  booktitle={International colloquium on automata, languages, and programming},
  pages={1--12},
  year={2006},
  organization={Springer}
}

@article{dwork2014algorithmic,
  title={The algorithmic foundations of differential privacy},
  author={Dwork, Cynthia and Roth, Aaron and others},
  journal={Foundations and Trends{\textregistered} in Theoretical Computer Science},
  volume={9},
  number={3--4},
  pages={211--407},
  year={2014},
  publisher={Now Publishers, Inc.}
}

@inproceedings{fredrikson2014privacy,
  title={Privacy in pharmacogenetics: An $\{$End-to-End$\}$ case study of personalized warfarin dosing},
  author={Fredrikson, Matthew and Lantz, Eric and Jha, Somesh and Lin, Simon and Page, David and Ristenpart, Thomas},
  booktitle={23rd USENIX security symposium (USENIX Security 14)},
  pages={17--32},
  year={2014}
}

@inproceedings{yeom2018privacy,
  title={Privacy risk in machine learning: Analyzing the connection to overfitting},
  author={Yeom, Samuel and Giacomelli, Irene and Fredrikson, Matt and Jha, Somesh},
  booktitle={2018 IEEE 31st computer security foundations symposium (CSF)},
  pages={268--282},
  year={2018},
  organization={IEEE}
}

@article{bindschaedler2017plausible,
  title={Plausible deniability for privacy-preserving data synthesis},
  author={Bindschaedler, Vincent and Shokri, Reza and Gunter, Carl A},
  journal={arXiv preprint arXiv:1708.07975},
  year={2017}
}

@inproceedings{abay2019privacy,
  title={Privacy preserving synthetic data release using deep learning},
  author={Abay, Nazmiye Ceren and Zhou, Yan and Kantarcioglu, Murat and Thuraisingham, Bhavani and Sweeney, Latanya},
  booktitle={Machine Learning and Knowledge Discovery in Databases: European Conference, ECML PKDD 2018, Dublin, Ireland, September 10--14, 2018, Proceedings, Part I 18},
  pages={510--526},
  year={2019},
  organization={Springer}
}

@inproceedings{fang2022dp,
  title={Dp-ctgan: Differentially private medical data generation using ctgans},
  author={Fang, Mei Ling and Dhami, Devendra Singh and Kersting, Kristian},
  booktitle={International Conference on Artificial Intelligence in Medicine},
  pages={178--188},
  year={2022},
  organization={Springer}
}

@article{rajabi2022tabfairgan,
  title={Tabfairgan: Fair tabular data generation with generative adversarial networks},
  author={Rajabi, Amirarsalan and Garibay, Ozlem Ozmen},
  journal={Machine Learning and Knowledge Extraction},
  volume={4},
  number={2},
  pages={488--501},
  year={2022},
  publisher={MDPI}
}

@inproceedings{ghatak2022survey,
  title={A survey on privacy preserving synthetic data generation and a discussion on a privacy-utility trade-off problem},
  author={Ghatak, Debolina and Sakurai, Kouichi},
  booktitle={International Conference on Science of Cyber Security},
  pages={167--180},
  year={2022},
  organization={Springer}
}

@inproceedings{arjovsky2017wasserstein,
  title={Wasserstein generative adversarial networks},
  author={Arjovsky, Martin and Chintala, Soumith and Bottou, L{\'e}on},
  booktitle={International conference on machine learning},
  pages={214--223},
  year={2017},
  organization={PMLR}
}

@inproceedings{hu2020hierarchical,
  title={Hierarchical modes exploring in generative adversarial networks},
  author={Hu, Mengxiao and Li, Jinlong and Hu, Maolin and Hu, Tao},
  booktitle={Proceedings of the AAAI Conference on Artificial Intelligence},
  volume={34},
  number={07},
  pages={10981--10988},
  year={2020}
}

@article{lin1991divergence,
  title={Divergence measures based on the Shannon entropy},
  author={Lin, Jianhua},
  journal={IEEE Transactions on Information theory},
  volume={37},
  number={1},
  pages={145--151},
  year={1991},
  publisher={IEEE}
}

@article{ramdas2017wasserstein,
  title={On wasserstein two-sample testing and related families of nonparametric tests},
  author={Ramdas, Aaditya and Garc{\'\i}a Trillos, Nicol{\'a}s and Cuturi, Marco},
  journal={Entropy},
  volume={19},
  number={2},
  pages={47},
  year={2017},
  publisher={MDPI}
}

@book{box2015time,
  title={Time series analysis: forecasting and control},
  author={Box, George EP and Jenkins, Gwilym M and Reinsel, Gregory C and Ljung, Greta M},
  year={2015},
  publisher={John Wiley \& Sons}
}

@article{radford2018improving,
  title={Improving language understanding by generative pre-training},
  author={Radford, Alec},
  year={2018}
}

@article{scarselli2008graph,
  title={The graph neural network model},
  author={Scarselli, Franco and Gori, Marco and Tsoi, Ah Chung and Hagenbuchner, Markus and Monfardini, Gabriele},
  journal={IEEE transactions on neural networks},
  volume={20},
  number={1},
  pages={61--80},
  year={2008},
  publisher={IEEE}
}

@article{song2020denoising,
  title={Denoising diffusion implicit models},
  author={Song, Jiaming and Meng, Chenlin and Ermon, Stefano},
  journal={arXiv preprint arXiv:2010.02502},
  year={2020}
}

@article{huang2020tabtransformer,
  title={Tabtransformer: Tabular data modeling using contextual embeddings},
  author={Huang, Xin and Khetan, Ashish and Cvitkovic, Milan and Karnin, Zohar},
  journal={arXiv preprint arXiv:2012.06678},
  year={2020}
}

@misc{adult_2,
  author       = {Becker, Barry and Kohavi, Ronny},
  title        = {{Adult}},
  year         = {1996},
  howpublished = {UCI Machine Learning Repository},
  note         = {{DOI}: https://doi.org/10.24432/C5XW20}
}

@article{yoon2023ehr,
  title={EHR-Safe: generating high-fidelity and privacy-preserving synthetic electronic health records},
  author={Yoon, Jinsung and Mizrahi, Michel and Ghalaty, Nahid Farhady and Jarvinen, Thomas and Ravi, Ashwin S and Brune, Peter and Kong, Fanyu and Anderson, Dave and Lee, George and Meir, Arie and others},
  journal={NPJ Digital Medicine},
  volume={6},
  number={1},
  pages={141},
  year={2023},
  publisher={Nature Publishing Group UK London}
}

@inproceedings{che2017boosting,
  title={Boosting deep learning risk prediction with generative adversarial networks for electronic health records},
  author={Che, Zhengping and Cheng, Yu and Zhai, Shuangfei and Sun, Zhaonan and Liu, Yan},
  booktitle={2017 IEEE International Conference on Data Mining (ICDM)},
  pages={787--792},
  year={2017},
  organization={IEEE}
}

@article{kate2022fingan,
  title={Fingan: Generative adversarial network for analytical customer relationship management in banking and insurance},
  author={Kate, Prateek and Ravi, Vadlamani and Gangwar, Akhilesh},
  journal={arXiv preprint arXiv:2201.11486},
  year={2022}
}

@book{pearl2014probabilistic,
  title={Probabilistic reasoning in intelligent systems: networks of plausible inference},
  author={Pearl, Judea},
  year={2014},
  publisher={Elsevier}
}

@article{dempster1977maximum,
  title={Maximum likelihood from incomplete data via the EM algorithm},
  author={Dempster, Arthur P and Laird, Nan M and Rubin, Donald B},
  journal={Journal of the royal statistical society: series B (methodological)},
  volume={39},
  number={1},
  pages={1--22},
  year={1977},
  publisher={Wiley Online Library}
}

@article{baum1970maximization,
  title={A maximization technique occurring in the statistical analysis of probabilistic functions of Markov chains},
  author={Baum, Leonard E and Petrie, Ted and Soules, George and Weiss, Norman},
  journal={The annals of mathematical statistics},
  volume={41},
  number={1},
  pages={164--171},
  year={1970},
  publisher={JSTOR}
}

@article{zhang2023mixed,
  title={Mixed-type tabular data synthesis with score-based diffusion in latent space},
  author={Zhang, Hengrui and Zhang, Jiani and Srinivasan, Balasubramaniam and Shen, Zhengyuan and Qin, Xiao and Faloutsos, Christos and Rangwala, Huzefa and Karypis, George},
  journal={arXiv preprint arXiv:2310.09656},
  year={2023}
}

@inproceedings{jolicoeur2024generating,
  title={Generating and imputing tabular data via diffusion and flow-based gradient-boosted trees},
  author={Jolicoeur-Martineau, Alexia and Fatras, Kilian and Kachman, Tal},
  booktitle={International Conference on Artificial Intelligence and Statistics},
  pages={1288--1296},
  year={2024},
  organization={PMLR}
}

@article{gulati2023tabmt,
  title={TabMT: Generating tabular data with masked transformers},
  author={Gulati, Manbir and Roysdon, Paul},
  journal={Advances in Neural Information Processing Systems},
  volume={36},
  pages={46245--46254},
  year={2023}
}

@article{borisov2022language,
  title={Language models are realistic tabular data generators},
  author={Borisov, Vadim and Se{\ss}ler, Kathrin and Leemann, Tobias and Pawelczyk, Martin and Kasneci, Gjergji},
  journal={arXiv preprint arXiv:2210.06280},
  year={2022}
}

@inproceedings{han2005borderline,
  title={Borderline-SMOTE: a new over-sampling method in imbalanced data sets learning},
  author={Han, Hui and Wang, Wen-Yuan and Mao, Bing-Huan},
  booktitle={International conference on intelligent computing},
  pages={878--887},
  year={2005},
  organization={Springer}
}

@article{kim2022stasy,
  title={Stasy: Score-based tabular data synthesis},
  author={Kim, Jayoung and Lee, Chaejeong and Park, Noseong},
  journal={arXiv preprint arXiv:2210.04018},
  year={2022}
}

@article{li2020fourier,
  title={Fourier neural operator for parametric partial differential equations},
  author={Li, Zongyi and Kovachki, Nikola and Azizzadenesheli, Kamyar and Liu, Burigede and Bhattacharya, Kaushik and Stuart, Andrew and Anandkumar, Anima},
  journal={arXiv preprint arXiv:2010.08895},
  year={2020}
}

@article{dhariwal2021diffusion,
  title={Diffusion models beat gans on image synthesis},
  author={Dhariwal, Prafulla and Nichol, Alexander},
  journal={Advances in neural information processing systems},
  volume={34},
  pages={8780--8794},
  year={2021}
}

@inproceedings{nichol2021improved,
  title={Improved denoising diffusion probabilistic models},
  author={Nichol, Alexander Quinn and Dhariwal, Prafulla},
  booktitle={International conference on machine learning},
  pages={8162--8171},
  year={2021},
  organization={PMLR}
}

@inproceedings{he2008adasyn,
  title={ADASYN: Adaptive synthetic sampling approach for imbalanced learning},
  author={He, Haibo and Bai, Yang and Garcia, Edwardo A and Li, Shutao},
  booktitle={2008 IEEE international joint conference on neural networks (IEEE world congress on computational intelligence)},
  pages={1322--1328},
  year={2008},
  organization={Ieee}
}

@article{durkan2019neural,
  title={Neural spline flows},
  author={Durkan, Conor and Bekasov, Artur and Murray, Iain and Papamakarios, George},
  journal={Advances in neural information processing systems},
  volume={32},
  year={2019}
}

@article{song2020score,
  title={Score-based generative modeling through stochastic differential equations},
  author={Song, Yang and Sohl-Dickstein, Jascha and Kingma, Diederik P and Kumar, Abhishek and Ermon, Stefano and Poole, Ben},
  journal={arXiv preprint arXiv:2011.13456},
  year={2020}
}

@article{acs2018differentially,
  title={Differentially private mixture of generative neural networks},
  author={Acs, Gergely and Melis, Luca and Castelluccia, Claude and De Cristofaro, Emiliano},
  journal={IEEE Transactions on Knowledge and Data Engineering},
  volume={31},
  number={6},
  pages={1109--1121},
  year={2018},
  publisher={IEEE}
}

@inproceedings{zemel2013learning,
  title={Learning fair representations},
  author={Zemel, Rich and Wu, Yu and Swersky, Kevin and Pitassi, Toni and Dwork, Cynthia},
  booktitle={International conference on machine learning},
  pages={325--333},
  year={2013},
  organization={PMLR}
}

@article{papernot2016semi,
  title={Semi-supervised knowledge transfer for deep learning from private training data},
  author={Papernot, Nicolas and Abadi, Mart{\'\i}n and Erlingsson, Ulfar and Goodfellow, Ian and Talwar, Kunal},
  journal={arXiv preprint arXiv:1610.05755},
  year={2016}
}

@article{chouldechova2018frontiers,
  title={The frontiers of fairness in machine learning},
  author={Chouldechova, Alexandra and Roth, Aaron},
  journal={arXiv preprint arXiv:1810.08810},
  year={2018}
}

@article{dal2017credit,
  title={Credit card fraud detection: a realistic modeling and a novel learning strategy},
  author={Dal Pozzolo, Andrea and Boracchi, Giacomo and Caelen, Olivier and Alippi, Cesare and Bontempi, Gianluca},
  journal={IEEE transactions on neural networks and learning systems},
  volume={29},
  number={8},
  pages={3784--3797},
  year={2017},
  publisher={IEEE}
}

@article{flores2022comparison,
  title={Comparison of predictive models with balanced classes using the SMOTE method for the forecast of student dropout in higher education},
  author={Flores, Vaneza and Heras, Stella and Julian, Vicente},
  journal={Electronics},
  volume={11},
  number={3},
  pages={457},
  year={2022},
  publisher={MDPI}
}

@article{ma2018eddi,
  title={Eddi: Efficient dynamic discovery of high-value information with partial vae},
  author={Ma, Chao and Tschiatschek, Sebastian and Palla, Konstantina and Hern{\'a}ndez-Lobato, Jos{\'e} Miguel and Nowozin, Sebastian and Zhang, Cheng},
  journal={arXiv preprint arXiv:1809.11142},
  year={2018}
}

@inproceedings{yoon2018gain,
  title={Gain: Missing data imputation using generative adversarial nets},
  author={Yoon, Jinsung and Jordon, James and Schaar, Mihaela},
  booktitle={International conference on machine learning},
  pages={5689--5698},
  year={2018},
  organization={PMLR}
}

@inproceedings{chang2022maskgit,
  title={Maskgit: Masked generative image transformer},
  author={Chang, Huiwen and Zhang, Han and Jiang, Lu and Liu, Ce and Freeman, William T},
  booktitle={Proceedings of the IEEE/CVF conference on computer vision and pattern recognition},
  pages={11315--11325},
  year={2022}
}

@article{bauer2024comprehensive,
  title={Comprehensive exploration of synthetic data generation: A survey},
  author={Bauer, Andr{\'e} and Trapp, Simon and Stenger, Michael and Leppich, Robert and Kounev, Samuel and Leznik, Mark and Chard, Kyle and Foster, Ian},
  journal={arXiv preprint arXiv:2401.02524},
  year={2024}
}

@article{guan2024zero,
  title={A zero auxiliary knowledge membership inference attack on aggregate location data},
  author={Guan, Vincent and Gu{\'e}pin, Florent and Cretu, Ana-Maria and de Montjoye, Yves-Alexandre},
  journal={arXiv preprint arXiv:2406.18671},
  year={2024}
}

@inproceedings{annamalai2024linear,
  title={A linear reconstruction approach for attribute inference attacks against synthetic data},
  author={Annamalai, Meenatchi Sundaram Muthu Selva and Gadotti, Andrea and Rocher, Luc},
  booktitle={33rd USENIX Security Symposium (USENIX Security 24)},
  pages={2351--2368},
  year={2024}
}

@article{mckenna2022aim,
  title={Aim: An adaptive and iterative mechanism for differentially private synthetic data},
  author={McKenna, Ryan and Mullins, Brett and Sheldon, Daniel and Miklau, Gerome},
  journal={arXiv preprint arXiv:2201.12677},
  year={2022}
}

@article{liu2021iterative,
  title={Iterative methods for private synthetic data: Unifying framework and new methods},
  author={Liu, Terrance and Vietri, Giuseppe and Wu, Steven Z},
  journal={Advances in Neural Information Processing Systems},
  volume={34},
  pages={690--702},
  year={2021}
}

@inproceedings{baak2022synthsonic,
  title={Synthsonic: fast, probabilistic modeling and synthesis of tabular data},
  author={Baak, Max and Brugman, Simon and Rojas, Ilan Fridman and Dalmeida, Lorraine and Urlus, Ralph EQ and Oger, Jean-Baptiste},
  booktitle={International Conference on Artificial Intelligence and Statistics},
  pages={4747--4763},
  year={2022},
  organization={PMLR}
}

@article{friedman1997bayesian,
  title={Bayesian network classifiers},
  author={Friedman, Nir and Geiger, Dan and Goldszmidt, Moises},
  journal={Machine learning},
  volume={29},
  number={2},
  pages={131--163},
  year={1997},
  publisher={Springer}
}

@article{chow1968approximating,
  title={Approximating discrete probability distributions with dependence trees},
  author={Chow, CKCN and Liu, Cong},
  journal={IEEE transactions on Information Theory},
  volume={14},
  number={3},
  pages={462--467},
  year={1968},
  publisher={IEEE}
}

@article{kinakh2024tabular,
  title={Tabular data generation using binary diffusion},
  author={Kinakh, Vitaliy and Voloshynovskiy, Slava},
  journal={arXiv preprint arXiv:2409.13882},
  year={2024}
}

@inproceedings{sklar1959fonctions,
  title={Fonctions de r{\'e}partition {\`a} n dimensions et leurs marges},
  author={Sklar, M},
  booktitle={Annales de l'ISUP},
  volume={8},
  number={3},
  pages={229--231},
  year={1959}
}

@article{hardt2012simple,
  title={A simple and practical algorithm for differentially private data release},
  author={Hardt, Moritz and Ligett, Katrina and McSherry, Frank},
  journal={Advances in neural information processing systems},
  volume={25},
  year={2012}
}

@inproceedings{nasr2019comprehensive,
  title={Comprehensive privacy analysis of deep learning: Passive and active white-box inference attacks against centralized and federated learning},
  author={Nasr, Milad and Shokri, Reza and Houmansadr, Amir},
  booktitle={2019 IEEE symposium on security and privacy (SP)},
  pages={739--753},
  year={2019},
  organization={IEEE}
}

@article{fonseca2023tabular,
  title={Tabular and latent space synthetic data generation: a literature review},
  author={Fonseca, Joao and Bacao, Fernando},
  journal={Journal of Big Data},
  volume={10},
  number={1},
  pages={115},
  year={2023},
  publisher={Springer}
}

@article{lautrup2024systematic,
  title={Systematic review of generative modelling tools and utility metrics for fully synthetic tabular data},
  author={Lautrup, Anton Danholt and Hyrup, Tobias and Zimek, Arthur and Schneider-Kamp, Peter},
  journal={ACM Computing Surveys},
  volume={57},
  number={4},
  pages={1--38},
  year={2024},
  publisher={ACM New York, NY}
}

@inproceedings{cormode2025synthetic,
  title={Synthetic tabular data: Methods, attacks and defenses},
  author={Cormode, Graham and Maddock, Samuel and Ullah, Enayat and Gade, Shripad},
  booktitle={Proceedings of the 31st ACM SIGKDD Conference on Knowledge Discovery and Data Mining V. 2},
  pages={5989--5998},
  year={2025}
}

@article{shi2025comprehensive,
  title={A comprehensive survey of synthetic tabular data generation},
  author={Shi, Ruxue and Wang, Yili and Du, Mengnan and Shen, Xu and Chang, Yi and Wang, Xin},
  journal={arXiv preprint arXiv:2504.16506},
  year={2025}
}

@article{umesh2025preserving,
  title={Preserving logical and functional dependencies in synthetic tabular data},
  author={Umesh, Chaithra and Schultz, Kristian and Mahendra, Manjunath and Bej, Saptarshi and Wolkenhauer, Olaf},
  journal={Pattern Recognition},
  volume={163},
  pages={111459},
  year={2025},
  publisher={Elsevier}
}

@article{umesh2026dependency,
  title={Dependency-aware synthetic tabular data generation},
  author={Umesh, Chaithra and Seegel-Schultz, Kristian and Mahendra, Manjunath and Bej, Saptarshi and Wolkenhauer, Olaf},
  journal={Pattern Recognition},
  pages={113819},
  year={2026},
  publisher={Elsevier}
}

@article{kim2025empirical,
  title={An empirical study on impact of label noise on synthetic tabular data generation},
  author={Kim, Jeonghoon and Huang, Chao and Liu, Xin},
  journal={Machine Learning},
  volume={114},
  number={4},
  pages={90},
  year={2025},
  publisher={Springer}
}

@inproceedings{d2025synthetic,
  title={Synthetic tabular data generation for imbalanced classification: The surprising effectiveness of an overlap class},
  author={D'souza, Annie and Swetha, M and Sarawagi, Sunita},
  booktitle={Proceedings of the AAAI Conference on Artificial Intelligence},
  volume={39},
  number={15},
  pages={16127--16134},
  year={2025}
}

@inproceedings{nguyen2024generating,
  title={Generating realistic tabular data with large language models},
  author={Nguyen, Dang and Gupta, Sunil and Do, Kien and Nguyen, Thin and Venkatesh, Svetha},
  booktitle={2024 IEEE International Conference on Data Mining (ICDM)},
  pages={330--339},
  year={2024},
  organization={IEEE}
}

@article{wang2024harmonic,
  title={Harmonic: Harnessing llms for tabular data synthesis and privacy protection},
  author={Wang, Yuxin and Feng, Duanyu and Dai, Yongfu and Chen, Zhengyu and Huang, Jimin and Ananiadou, Sophia and Xie, Qianqian and Wang, Hao},
  journal={Advances in Neural Information Processing Systems},
  volume={37},
  pages={100196--100212},
  year={2024}
}

@article{guan2026prism,
  title={Prism: Private Relational Data Synthesis with Language Models},
  author={Guan, Guohui and Ge, Chang},
  journal={Proceedings of the ACM on Management of Data},
  volume={4},
  number={3 (SIGMOD},
  pages={1--28},
  year={2026},
  publisher={ACM New York, NY, USA}
}

@inproceedings{an2025masked,
  title={Masked language modeling becomes conditional density estimation for tabular data synthesis},
  author={An, Seunghwan and Woo, Gyeongdong and Lim, Jaesung and Kim, ChangHyun and Hong, Sungchul and Jeon, Jong-June},
  booktitle={Proceedings of the AAAI Conference on Artificial Intelligence},
  volume={39},
  number={15},
  pages={15356--15364},
  year={2025}
}

@article{gardner2023benchmarking,
  title={Benchmarking distribution shift in tabular data with tableshift},
  author={Gardner, Josh and Popovic, Zoran and Schmidt, Ludwig},
  journal={Advances in Neural Information Processing Systems},
  volume={36},
  pages={53385--53432},
  year={2023}
}

@inproceedings{wang2024large,
  title={Large language models as data augmenters for cold-start item recommendation},
  author={Wang, Jianling and Lu, Haokai and Caverlee, James and Chi, Ed H and Chen, Minmin},
  booktitle={Companion proceedings of the ACM web conference 2024},
  pages={726--729},
  year={2024}
}

@article{beyazit2023inductive,
  title={An inductive bias for tabular deep learning},
  author={Beyazit, Ege and Kozaczuk, Jonathan and Li, Bo and Wallace, Vanessa and Fadlallah, Bilal},
  journal={Advances in Neural Information Processing Systems},
  volume={36},
  pages={43108--43135},
  year={2023}
}

@article{bickel1975sex,
  title={Sex Bias in Graduate Admissions: Data from Berkeley: Measuring bias is harder than is usually assumed, and the evidence is sometimes contrary to expectation.},
  author={Bickel, Peter J and Hammel, Eugene A and O'Connell, J William},
  journal={Science},
  volume={187},
  number={4175},
  pages={398--404},
  year={1975},
  publisher={American Association for the Advancement of Science}
}

@inproceedings{liu2025can,
  title={Can synthetic data be fair and private? A comparative study of synthetic data generation and fairness algorithms},
  author={Liu, Qinyi and Deho, Oscar and Vadiee, Farhad and Khalil, Mohammad and Joksimovic, Srecko and Siemens, George},
  booktitle={Proceedings of the 15th International Learning Analytics and Knowledge Conference},
  pages={591--600},
  year={2025}
}

@article{kwon2026generative,
  title={Do Generative Models Keep Time? A Time-Aware Evaluation of Synthetic Sequential Tabular Data},
  author={Kwon, Kiwan and Kim, Kangmin and Lee, Hojin and Jung, Yeseong and Kong, Hyeongwoo and Potluru, Vamsi K and Park, Saerom and Lee, Yongjae},
  journal={arXiv preprint arXiv:2607.15606},
  year={2026}
}

@article{jia2024tabular,
  title={A tabular data generation framework guided by downstream tasks optimization},
  author={Jia, Fengwei and Zhu, Hongli and Jia, Fengyuan and Ren, Xinyue and Chen, Siqi and Tan, Hongming and Chan, Wai Kin Victor},
  journal={Scientific Reports},
  volume={14},
  number={1},
  pages={15267},
  year={2024},
  publisher={Nature Publishing Group UK London}
}

@inproceedings{alismail2025survey,
  title={A survey of llm-based methods for synthetic data generation and the rise of agentic workflows},
  author={Alismail, Ahmad and Lanquillon, Carsten},
  booktitle={International conference on human-computer interaction},
  pages={119--135},
  year={2025},
  organization={Springer}
}

@inproceedings{shi2025tabdiff,
  title={Tabdiff: a mixed-type diffusion model for tabular data generation},
  author={Shi, Juntong and Xu, Minkai and Hua, Harper and Zhang, Hengrui and Ermon, Stefano and Leskovec, Jure},
  booktitle={International Conference on Learning Representations},
  volume={2025},
  pages={37353--37375},
  year={2025}
}

\pagebreak
\appendix
\renewcommand{\thesection}{\Alph{section}}
\renewcommand{\thesubsection}{\Alph{section}.\arabic{subsection}}
\renewcommand{\thetable}{\thesection.\arabic{table}}

\section*{Appendices}
\label{sec:appendix}
\section{Definitions in Literature}
\label{app:definitions}
To understand the nuances of synthetic data generation, it is useful to compare how different methods define the process. Table~\ref{tab:synthetic_data_definitions} contains examples of synthetic data generation definitions in prior literature, providing a comparative overview of these definitions and highlighting the mathematical formulation underlying synthetic data generation. Existing definitions emphasize the estimation of the original data distribution \( P(\mathbf{X}) \) and the generation of new synthetic data \( \mathbf{X_{\text{syn}}} \) by sampling from an estimated distribution \( \hat{P}(\mathbf{X}) \). However, different approaches prioritize distinct aspects. 
For instance, \raju{Bishop et al.} \cite{bishop2006pattern} and \raju{Goodfellow et al.} \cite{goodfellow2014generative} emphasize the need to ensure that \( \mathbf{X_{\text{syn}}} \) maintains the statistical properties and feature dependencies of the original dataset. GANs \cite{goodfellow2014generative} achieve this through adversarial learning, while VAEs ~\cite{kingma2013auto}, use latent variable models to structure the data.
\raju{Several flow based models \cite{kamthe2021copula,xu2019modeling,kim2022stasy}, learn to transform simple base distributions (e.g., Gaussian) into the data distribution. They handle continuous and categorical features, preserve statistical properties, and allow efficient sampling, complementing GANs and VAEs for tabular data generation.}
Kingma and Welling \cite{kingma2013auto} describe conditional synthetic data generation as sampling from \( P(\mathbf{X}\mid A = a) \), ensuring that the generated data aligns with specified conditions.  
Lastly, \raju{Reynolds et al.} \cite{reynolds2009gaussian} and \raju{Goodfellow et al.} \cite{goodfellow2014generative} discuss how privacy-focused synthetic data methods introduce noise into \( \hat{P}(\mathbf{X}) \) or the sampling process to prevent re-identification while maintaining data utility. 
\colorred{These perspectives inform the definition presented in
Section~\ref{sec:SDG}.}
\begin{longtable}[t]{|p{3.5cm}|p{9.5cm}|}
\caption{Examples of Synthetic Data Generation Definitions in prior literature} 
\label{tab:synthetic_data_definitions} \\

\hline
\textbf{Source} & \textbf{Definition} \\
\hline
\endfirsthead

\hline
\textbf{Source} & \textbf{Definition} \\
\hline
\endhead

\hline
\endfoot

\hline
\textbf{Bishop (2006)} \cite{bishop2006pattern} & 
\begin{minipage}[t]{\linewidth}
Discusses probabilistic modeling and density estimation, which serve as the foundation for synthetic data generation. The goal is to approximate the underlying distribution of real data $\mathbf{X}$ using a model $\hat{P}(\mathbf{X})$ and generate synthetic samples $Y$ by drawing from this estimated distribution:

\[
\mathbf{X}_{\text{syn}} \sim \hat{P}(\mathbf{X}).
\]

This approach includes parametric and non-parametric methods to learn $\hat{P}(\mathbf{X})$.
\end{minipage} \\ \hline

\textbf{Goodfellow et al. (2014)} \cite{goodfellow2014generative} & 
\begin{minipage}[t]{\linewidth}
Emphasizes the use of Generative Adversarial Networks (GANs), where a generator network \( G \) maps random noise \( Z \sim P(Z) \) to synthetic data \( X_{\text{syn}} \), and a discriminator network \( D \) evaluates its authenticity. The iterative training process refines the synthetic data distribution \( \hat{P}(X) \) to closely approximate the original data distribution \( P(X) \). This process is mathematically represented as:

\[
\mathbf{X}_{\text{syn}} \sim \hat{P}(\mathbf{X}).
\]

The adversarial training objective is given by:

\[
\min_{G} \max_{D} \mathbb{E}_{X \sim P(X)} [\log D(X)] + \mathbb{E}_{Z \sim P(Z)} [\log (1 - D(G(Z)))].
\]

Through this process, the generator learns to produce synthetic samples that are statistically similar to real data, ensuring that \( \hat{P}(\mathbf{X}) \approx P(\mathbf{X}) \).
\end{minipage} \\ \hline

\textbf{Kingma \& Welling (2013)} \cite{kingma2013auto} & 
\begin{minipage}[t]{\linewidth}
Introduces Variational Autoencoders (VAEs), which generate synthetic data by learning a probabilistic mapping between latent variables $\mathbf{Z}$ and observations $\mathbf{X}$. Given a prior $p(\mathbf{Z})$, synthetic samples are drawn as follows:

\[
\mathbf{Z} \sim \mathcal{N}(\mathbf{0}, \mathbf{I}), \quad \mathbf{X}_{\text{syn}} \sim p(\mathbf{X} \mid \mathbf{Z}).
\]

where $p(\mathbf{X} \mid \mathbf{Z})$ is a probability density function and $\mathbf{Z}$ is a latent variable. The objective is to maximize the evidence lower bound (ELBO) to approximate $P(\mathbf{X})$ via $p(\mathbf{X} \mid \mathbf{Z})$.
\end{minipage} \\ \hline

\textbf{Reynolds (2009)} \cite{reynolds2009gaussian} & 
\begin{minipage}[t]{\linewidth}
Discusses synthetic data generation using Gaussian Mixture Models (GMMs), where the original data's distribution is approximated by a mixture of Gaussians. The synthetic data is generated by sampling from this mixture model:

\[
P(\mathbf{X}) = \sum_{k=1}^{K} \pi_k \mathcal{N}(\mathbf{X} \mid \mu_k, \Sigma_k).
\]

where \( \pi_k \) are the mixture weights, and \( \mathcal{N}(\mathbf{X} \mid \mu_k, \Sigma_k) \) represents the Gaussian component with mean \( \mu_k \) and covariance \( \Sigma_k \). The synthetic data is then sampled as:

\[
\mathbf{X}_{\text{syn}} \sim \hat{P}(\mathbf{X}) = \sum_{k=1}^{K} \hat{\pi}_k \mathcal{N}(\mathbf{X} \mid \hat{\mu}_k, \hat{\Sigma}_k).
\]

where \( \hat{\pi}_k, \hat{\mu}_k, \) and \( \hat{\Sigma}_k \) are estimated from the original data distribution.
\end{minipage} \\ \hline

\textbf{Rabiner (1989)} \cite{rabiner1989tutorial} & 
\begin{minipage}[t]{\linewidth}
Synthetic data generation for sequential data can be achieved using Hidden Markov Models (HMMs), where sequences are modeled probabilistically as a series of hidden states and observed outputs. The synthetic sequences are generated by sampling from the estimated HMM parameters:

\[
P(\mathbf{X}, S) = P(S_1) \prod_{t=2}^{T} P(S_t \mid S_{t-1}) P(\mathbf{X}_t \mid S_t).
\]

where \( S_t \) is the hidden state at time \( t \), and \( \mathbf{X}_t \) is the observed output. The synthetic sequence is then generated as:

\[
\mathbf{X}_{\text{syn}} \sim \hat{P}(\mathbf{X}) = \hat{P}(S_1) \prod_{t=2}^{T} \hat{P}(S_t \mid S_{t-1}) \hat{P}(\mathbf{X}_t \mid S_t).
\]

where \( \hat{P}(S_t \mid S_{t-1}) \) and \( \hat{P}(\mathbf{X}_t \mid S_t) \) are the estimated transition and emission probabilities, respectively. This ensures that \( \hat{P}(\mathbf{X}) \) approximates \( P(\mathbf{X}) \), generating realistic sequential synthetic data.
\end{minipage}  
\\ \hline

\end{longtable}

\pagebreak

\section{Synthetic Data Generation} \label{sec:background}

\colorred{This section reviews the generative and statistical models
that underpin synthetic tabular data generation.}

\subsection{Generative Models}
\label{subsec:generative_models}
Generative models aim to learn the underlying distribution of a dataset and can generate new data points that resemble the training data. These models are particularly valuable in situations where acquiring real data is difficult, expensive, or raises privacy concerns \cite{goodfellow2014generative,kingma2013auto,jordon2018pate}. By capturing the essential features and patterns of the data, generative models can produce high-quality synthetic samples that maintain the statistical properties of the original dataset.

A \textit{generative model} \( G \) is a model that learns the underlying data distribution and generates new samples that resemble the original data. Formally, given a generative model \( G \), it produces synthetic data points  
\[
\mathbf{X}_{\text{syn}} = \{ \mathbf{x}'_i \in \mathbb{R}^p \mid i = 1, 2, \ldots, n \},
\]  
where each sample \( \mathbf{x}'_i \) is drawn from the learned distribution, i.e., \( \mathbf{x}'_i \sim G \).

\subsubsection{Autoencoders and Variational Autoencoders (VAEs)}

\begin{figure}[htpb]
    \centering
    \begin{subfigure}[b]{0.49\textwidth}
\centering\includegraphics[width=\linewidth]{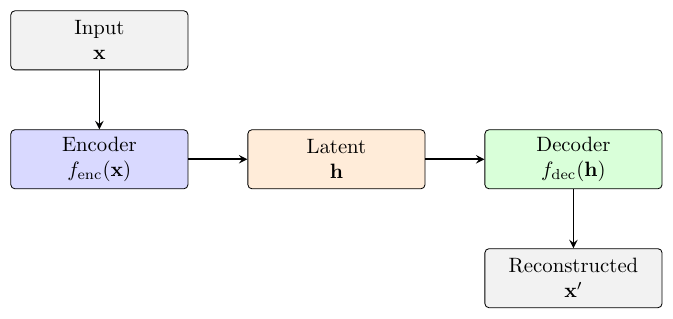}
        \caption{Autoencoder}
        \label{fig:a}
    \end{subfigure}
    \hfill
    \begin{subfigure}[b]{0.5\textwidth}
        \centering
        \includegraphics[width=\linewidth]{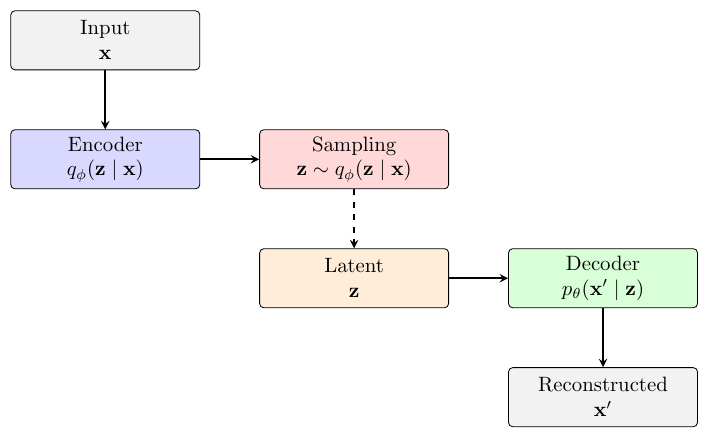}
        \caption{Variational Autoencoder}
        \label{fig:b}
    \end{subfigure}
    \caption{(a) A standard Autoencoder encodes the input into a deterministic latent representation \( \mathbf{h} \) before reconstructing it. (b) A Variational Autoencoder (VAE) introduces a probabilistic latent space, where the encoder maps the input into a distribution over possible latent representations.}
    \label{fig:ae_vae}
\end{figure}

Autoencoders \cite{kingma2013auto} are a type of neural network that encode data into a latent space and then reconstruct it back to the original input. The process can be mathematically represented as follows:  
\[
\mathbf{h} = f_{\text{enc}}(\mathbf{x}),
\]
\[
\mathbf{x}' = f_{\text{dec}}(\mathbf{h}),
\]
where \( f_{\text{enc}} \) and \( f_{\text{dec}} \) are the encoder and decoder functions, respectively. The encoder compresses the input data \( \mathbf{x} \) into a lower-dimensional representation \( \mathbf{h} \), while the decoder attempts to reconstruct the original data from this representation. Figure~\ref{fig:a} illustrates the structure of a standard Autoencoder (AE), where the latent code \( \mathbf{h} \) serves as a bottleneck that forces the model to learn compact feature representations.

Variational Autoencoders (VAEs) \cite{kingma2013auto} extend the concept of traditional autoencoders by introducing a probabilistic approach to the latent space. Instead of mapping inputs to a fixed latent code \( \mathbf{h} \), VAEs learn a distribution over latent variables \( \mathbf{z} \). As depicted in Figure~\ref{fig:b}, rather than directly outputting a single latent vector, the encoder produces parameters for a probability distribution:  
\[
q_{\phi}(\mathbf{z} \mid \mathbf{x}) = \mathcal{N}(\boldsymbol{\mu}, \boldsymbol{\sigma}^2\mathbf{I}),
\]  
where \( \boldsymbol{\mu} \) and \( \boldsymbol{\sigma}^2 \) represent the learned mean and variance of the distribution. Instead of using \( \mathbf{h} \) deterministically, VAEs sample \( \mathbf{z} \) from the encoded distribution:
\[
\mathbf{z} \sim \mathcal{N}(\boldsymbol{\mu}, \boldsymbol{\sigma}^2).
\]
The decoder then reconstructs the data by mapping \( \mathbf{z} \) back to the data space through a likelihood function:
\[
p_{\theta}(\mathbf{x}' \mid \mathbf{z}) \approx \mathcal{N}(f_{\text{dec}}(\mathbf{z}), \sigma^2 \mathbf{I}).
\]
The use of a probabilistic latent space allows VAEs to generate diverse samples by drawing from different points in the learned distribution, rather than relying on a single deterministic encoding as in standard Autoencoders.

The training objective of VAEs maximizes the Evidence Lower Bound (ELBO), which consists of two key terms:  
\begin{enumerate}
    \item \textbf{Reconstruction Loss} – Ensuring that \( \mathbf{x}' \) closely resembles the original input \( \mathbf{x} \).
    \item \textbf{Regularization Term} – Encouraging \( q_{\phi}(\mathbf{z} \mid \mathbf{x}) \) to be close to a prior distribution \( p(\mathbf{z}) \), typically a standard Gaussian:  
    \[
    p(\mathbf{z}) = \mathcal{N}(\mathbf{0}, \mathbf{I}).
    \]
\end{enumerate}
By enforcing this constraint, VAEs ensure that the latent space remains structured and meaningful, making them particularly effective for generative tasks such as image synthesis and anomaly detection.  

\subsubsection{Generative Adversarial Networks (GANs)}
\begin{figure}[htpb]
    \centering
    \centering\includegraphics[width=\linewidth]{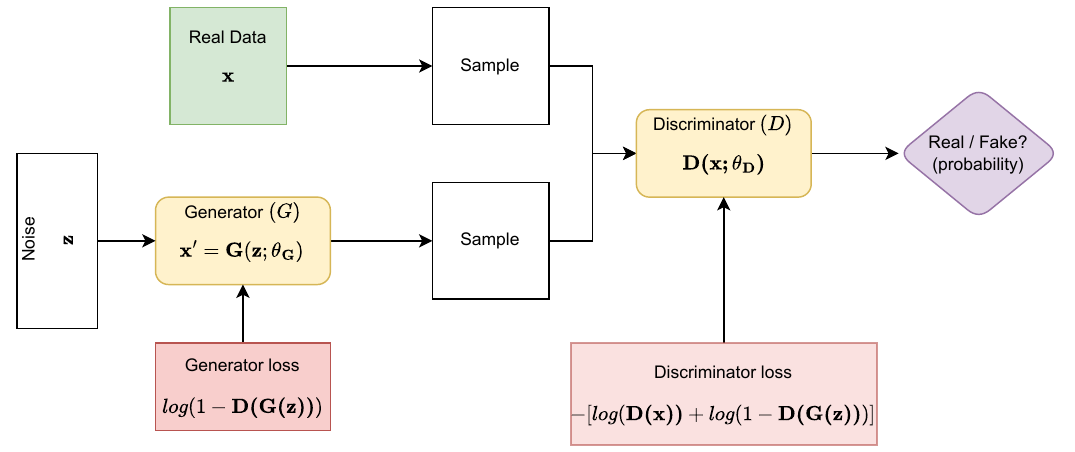}
    \caption{Architecture of a Generative Adversarial Network (GAN). The generator \( \mathbf{G} \) transforms noise \( \mathbf{z} \) into synthetic data \( \mathbf{x'} \), while the discriminator \( \mathbf{D} \) distinguishes between real data \( \mathbf{x} \) and generated data \( \mathbf{x'} \). \colorred{The generator minimizes the generated-sample term in Eq.~(\ref{eq:gan_g_logpbly}), whereas the discriminator maximizes the full adversarial objective in Eq.~(\ref{eq:gan_loss}).}}

    \label{fig:gen_gan}
\end{figure}

Generative Adversarial Networks (GANs) \cite{goodfellow2014generative} are a revolutionary framework in machine learning that consists of two neural networks: a generator \( \mathbf{G} \) and a discriminator \( \mathbf{D} \). The generator is responsible for creating synthetic data, while the discriminator evaluates the authenticity of the data, distinguishing between real data and the data generated by the generator. The relationship between the generator and the noise input can be expressed as:
\begin{equation}
\mathbf{x'} = \mathbf{G}(\mathbf{z}; \theta_{\mathbf{G}}),
\label{eq:gan_relinputnoise}
\end{equation}
where \( \mathbf{z} \) is a noise vector and \( \theta_{\mathbf{G}} \) are the generator parameters \cite{goodfellow2014generative}. The generator \( \mathbf{G} \) learns to transform noise \( \mathbf{z} \) into synthetic data \( \mathbf{x'} \) that resembles the real data \( \mathbf{X} \), while the discriminator \( \mathbf{D} \) tries to distinguish between real \( \mathbf{x} \) and generated \( \mathbf{x'} \). Figure \ref{fig:gen_gan} illustrates the architecture of a typical GAN, showing the interaction between the generator and discriminator during the learning process. The figure highlights the flow of data from the noise vector \( \mathbf{z} \) to the generator \( \mathbf{G} \), the generation of synthetic data \( \mathbf{x'} \), and the discriminator \( \mathbf{D} \)'s role in classifying real and generated data. GANs are trained using an adversarial objective to find an equilibrium:
\begin{equation}
\min_{\mathbf{G}} \max_{\mathbf{D}} \mathbb{E}_{\mathbf{x} \sim p_{\text{data}}(\mathbf{x})} [\log \mathbf{D}(\mathbf{x})] + \mathbb{E}_{\mathbf{z} \sim p_{\mathbf{z}}(\mathbf{z})} [\log (1 - \mathbf{D}(\mathbf{G}(\mathbf{z})))].
\label{eq:gan_loss}
\end{equation}
Here, 
\begin{equation}
\mathbb{E}_{\mathbf{x} \sim p_{\text{data}}(\mathbf{x})} [\log \mathbf{D}(\mathbf{x})]
\label{eq:gan_d_logpbly}
\end{equation}
represents the log-probability that the discriminator correctly classifies real data \( \mathbf{x} \), and 
\begin{equation}
    \mathbb{E}_{\mathbf{z} \sim p_{\mathbf{z}}(\mathbf{z})} [\log (1 - \mathbf{D}(\mathbf{G}(\mathbf{z})))]
    \label{eq:gan_g_logpbly}
\end{equation} represents the log-probability that the discriminator correctly classifies generated data \( \mathbf{G}(\mathbf{z}) \) \cite{goodfellow2014generative}. The adversarial training process is visually summarized in Figure \ref{fig:gen_gan}, which also includes the generator and discriminator losses.

The training process of Generative Adversarial Networks (GANs) involves an iterative optimization of two neural networks: the generator \( \mathbf{G} \) and the discriminator \( \mathbf{D} \). Initially, the parameters of both networks are randomly initialized. In each training iteration, a batch of real data is sampled from the training dataset, while a batch of noise vectors is drawn from a predefined prior distribution, such as a Gaussian distribution. The generator then transforms these noise vectors into synthetic data samples, denoted as Eq.(\ref{eq:gan_relinputnoise}). The discriminator is updated by optimizing a loss function that encourages it to correctly classify real data as authentic and generated data as synthetic Eq.(\ref{eq:gan_loss}). \colorred{Simultaneously, the generator minimizes the generated-sample term in Eq.~(\ref{eq:gan_g_logpbly}), encouraging \(\mathbf{G}\) to produce samples that \(\mathbf{D}\) classifies as real.} This adversarial process continues iteratively until convergence, where the generator produces synthetic data that closely resembles the real distribution while the discriminator struggles to distinguish between real and generated samples, as demonstrated in foundational studies on GANs \cite{goodfellow2014generative}. The interplay between the generator and discriminator is visually depicted in Figure \ref{fig:gen_gan}, which provides a clear representation of the GAN architecture and training process.
GANs have gained significant attention for their ability to generate high-quality synthetic data, making them a powerful tool in various applications, including image generation, video synthesis, and more. 

\subsubsection{Diffusion Models}
Diffusion models are a class of generative models that progressively add noise to data and then learn how to reverse this noising process to generate new samples. The process is defined by two main phases: the forward (noising) process and the reverse (denoising) process \cite{ho2020denoising}.

Let \( \mathbf{x}_0 \in \mathbb{R}^p \) be the original data point (e.g., an image or tabular data point). The forward process is a Markov chain that gradually adds Gaussian noise over a series of \( T \) steps. The forward process at step \( t \) is defined as:

\[
\mathbf{x}_t = \sqrt{\alpha_t} \mathbf{x}_{t-1} + \sqrt{1 - \alpha_t} \mathbf{\epsilon}_t, \quad t = 1, 2, \dots, T,
\]
where \( \mathbf{\epsilon}_t \sim \mathcal{N}(0, I) \) is standard Gaussian noise, and \( \alpha_t \) is a scaling factor that controls the amount of noise added at each step \cite{ho2020denoising}. The process continues until \( t = T \), at which point the data is completely noisy and approximately follows a standard normal distribution \( \mathbf{x}_T \sim \mathcal{N}(0, I) \).

The reverse process is the core of the diffusion model. Given the noisy data \( \mathbf{x}_T \), the goal is to reverse the noising process and recover data that resembles the original distribution. The reverse process is parameterized by a neural network and is given by:
\[
p_\theta(\mathbf{x}_{t-1} \mid \mathbf{x}_t) = \mathcal{N}(\mathbf{x}_{t-1}; \mu_\theta(\mathbf{x}_t, t), \Sigma_\theta(\mathbf{x}_t, t)),
\]
where \( \mu_\theta(\mathbf{x}_t, t) \) and \( \Sigma_\theta(\mathbf{x}_t, t) \) are the mean and variance predicted by the model at each step \( t \), and the neural network learns to reverse the forward process.

\begin{figure}[htpb]
    \centering
    \includegraphics[width=\linewidth]{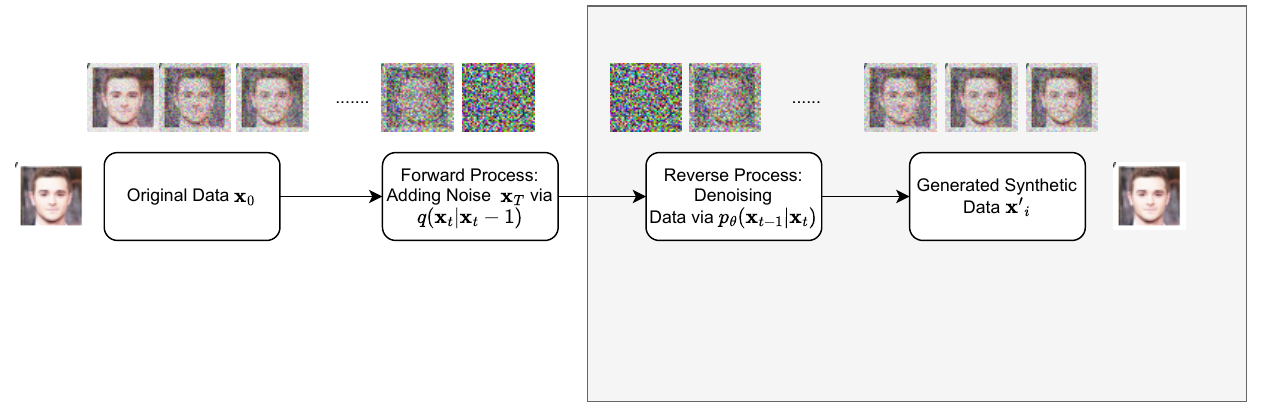}
    \caption{Illustration of the Diffusion Process: The forward process adds noise to the data progressively, while the reverse process denoises it to generate synthetic data. The process begins with original data \( \mathbf{x}_0 \) and ends with the generated data \( \mathbf{x'}_i \).}
    \label{fig:gen_diff}
\end{figure}

The objective of training a diffusion model is to minimize the Kullback-Leibler (KL) divergence between the true reverse process \( q(\mathbf{x}_{t-1} \mid \mathbf{x}_t, \mathbf{x}_0) \) and the learned reverse process \( p_\theta(\mathbf{x}_{t-1} \mid \mathbf{x}_t) \). \colorred{The corresponding loss is given in
Eq.~\ref{eq:diffusion_loss}:}
\begin{equation}
\mathcal{L} = \mathbb{E}_{q(\mathbf{x}_0, \mathbf{x}_1, \dots, \mathbf{x}_T)} \left[ \sum_{t=1}^T \text{KL}(q(\mathbf{x}_{t-1} \mid \mathbf{x}_t, \mathbf{x}_0) \| p_\theta(\mathbf{x}_{t-1} \mid \mathbf{x}_t)) \right],
\label{eq:diffusion_loss}
\end{equation}
where \( q(\mathbf{x}_{t-1} \mid \mathbf{x}_t, \mathbf{x}_0) \) is the true posterior distribution, and \( p_\theta(\mathbf{x}_{t-1} \mid \mathbf{x}_t) \) is the model’s learned reverse process \cite{ho2020denoising,song2020denoising}. 
Once the model is trained, synthetic data can be generated by starting from random noise \( \mathbf{x}_T \sim \mathcal{N}(0, I) \) and applying the learned reverse process to iteratively denoise the data:

\[
\mathbf{x'}_i = \hat{\mathbf{x}}_T \to \hat{\mathbf{x}}_{T-1} \to \dots \to \hat{\mathbf{x}}_1 \to \hat{\mathbf{x}}_0 = G(z),
\]
where \( G(z) \) denotes the generative process starting from random noise \( z \), and \( \hat{\mathbf{x}}_t \) represents the denoised data at each step. The final generated data point \( \mathbf{x'}_i \) approximates the original data distribution after applying the reverse diffusion steps \cite{song2020denoising}. This iterative process allows the model to learn to generate high-quality synthetic data that resembles the real data distribution. Figure~\ref{fig:gen_diff} illustrates the forward and reverse processes in diffusion models.

\subsubsection{Autoregressive Models}
Autoregressive generative models, initially developed for tasks such as language modeling \cite{radford2018improving}, have gained traction in the domain of synthetic data generation, particularly for structured tabular data. These models generate data sequentially, producing each feature conditioned on the previously generated ones. This formulation allows them to capture complex dependencies between features, which is essential in tabular data where relationships are often non-linear and cross-dimensional \cite{hoogeboom2021autoregressive}.

Formally, the generative process is defined as:
\[
\mathbf{x}'_i = G(\mathbf{x}_{<i}; \theta_G),
\]
where \( \mathbf{x}_{<i} = (\mathbf{x}'_1, \mathbf{x}'_2, \dots, \mathbf{x}'_{i-1}) \) denotes the sequence of previously generated features, \( \mathbf{x}'_i \) is the feature being generated at step \( i \), and \( \theta_G \) are the parameters of the generative model. The full synthetic data sample is then constructed as:
\[
\mathbf{x}' = (\mathbf{x}'_1, \mathbf{x}'_2, \dots, \mathbf{x}'_n).
\]

These models are typically trained using maximum likelihood estimation (MLE), optimizing the negative log-likelihood (NLL) over the training data:
\[
\mathcal{L} = -\sum_{i=1}^{n} \log P(\mathbf{x}_i \mid \mathbf{x}_{<i}; \theta_G),
\]
where \( P(\mathbf{x}_i \mid \mathbf{x}_{<i}; \theta_G) \) represents the conditional probability of the \( i \)-th feature given the previous ones. This objective encourages the model to accurately learn and replicate the statistical dependencies within the dataset.

This autoregressive formulation aligns naturally with the Transformer architecture \cite{vaswani2017attention}, which relies on self-attention mechanisms to model dependencies across input positions. By applying causal (masked) self-attention, Transformers ensure that the prediction at position \( i \) is conditioned only on positions \( <i \), thereby maintaining the autoregressive property. This makes them especially well-suited for tasks like synthetic tabular data generation, where feature dependencies can be highly non-linear and sparse \cite{huang2020tabtransformer}. A Transformer-based autoregressive model for synthetic tabular data generation typically incorporates the following core components:

\noindent\textbf{Input Embedding}: Each feature, whether categorical or numerical, is mapped into a high-dimensional space using embeddings. For numerical features, this is achieved through a linear transformation:
\[
\mathbf{E(x)} = W_e \mathbf{x},
\]
where \( W_e \) is the embedding weight matrix. For categorical features, separate embeddings are learned:
\[
\mathbf{E_{\text{cat}}(x)} = W_c \mathbf{x}.
\]
\noindent\textbf{Positional Encoding}: Since Transformers do not inherently encode the order of features, positional encodings are added to retain information about the sequence of features. The positional encoding for the \( i \)-th feature is computed as:
\[
\mathbf{PE(x)_i} = \sin\left(\frac{i}{10000^{2j/d}}\right), \quad \mathbf{PE(x)_{i+1}} = \cos\left(\frac{i}{10000^{2j/d}}\right),
\]
where \( d \) is the dimensionality of the embeddings.

\noindent\textbf{Multi-Head Self-Attention}: The self-attention mechanism captures dependencies between features by computing queries (\( \mathbf{Q} \)), keys (\( \mathbf{K} \)), and values (\( \mathbf{V} \)):
\[
\mathbf{Q} = W_Q \mathbf{E(x)}, \quad \mathbf{K} = W_K \mathbf{E(x)}, \quad \mathbf{V} = W_V \mathbf{E(x)}.
\]
The attention scores are computed as:
\begin{equation}
\text{Attention}(\mathbf{Q}, \mathbf{K}, \mathbf{V}) = \text{softmax} \left( \frac{\mathbf{QK}^T}{\sqrt{d_k}} \right) \mathbf{V},
\label{eq:selfattention}
\end{equation}
where \( d_k \) is the dimensionality of the keys. Multi-head attention extends this by computing multiple attention scores in parallel and concatenating the results.

\noindent\textbf{Feedforward Network}: After the self-attention layer, each feature representation is passed through a feedforward neural network:
\[
\mathbf{x'} = \sigma(W_2 \text{ReLU}(W_1 \mathbf{x})),
\]
where \( W_1 \) and \( W_2 \) are learnable weight matrices, and \( \sigma \) is a non-linear activation function.

\begin{figure}[htpb]
\centering
\includegraphics[width=0.7\linewidth]{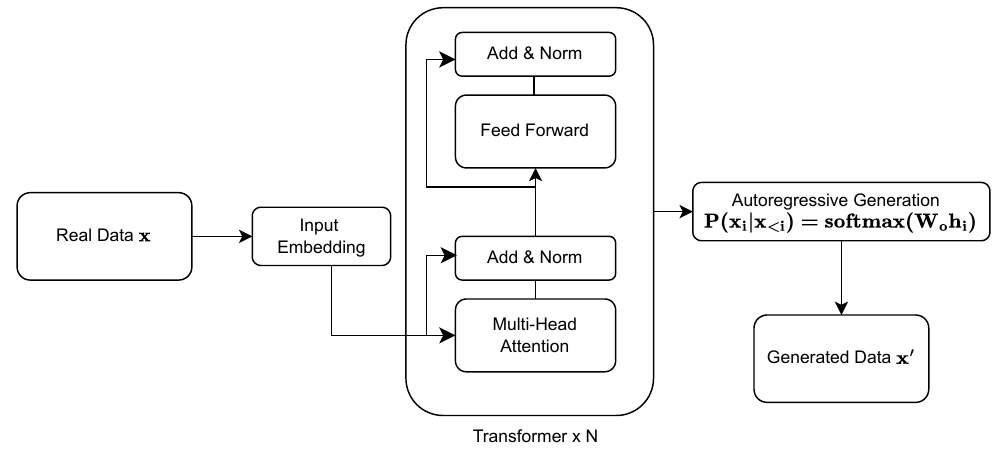}
    \caption{Architecture of a Transformer-based Synthetic Data Generator. The model processes input features through embeddings and positional encodings, applies self-attention to capture dependencies, and generates synthetic samples autoregressively.}
    \label{fig:gen_transformer}
\end{figure}

\noindent\textbf{Autoregressive Generation}: The final step involves generating each feature sequentially. The probability distribution for the \( i \)-th feature is computed as:
\[
P(\mathbf{x_i} | \mathbf{x_{<i}}) = \text{softmax}(W_o h_i),
\]
where \( h_i \) is the hidden representation of the \( i \)-th feature, and \( W_o \) is the output weight matrix. This integrated autoregressive design enables Transformer models to generate synthetic tabular data feature-by-feature, capturing both linear and non-linear dependencies in the order defined during training. Refer to Figure~\ref{fig:gen_transformer} for a visual representation of the autoregressive Transformer-based generation process.

\subsubsection{Masked Generative Models}
Masked generative models are a class of models that learn to model the data distribution by predicting missing or masked features conditioned on the observed ones. These models iteratively mask and reconstruct portions of the data, enabling them to capture complex feature dependencies in high-dimensional data such as tabular datasets \cite{ma2018eddi, yoon2018gain}.

Let \( \mathbf{x} = (x_1, x_2, \dots, x_p) \in \mathbb{R}^p \) denote a data sample with \( p \) features. A random binary mask \( \mathbf{m} \in \{0, 1\}^p \) indicates which features are observed (\( m_j = 1 \)) and which are masked (\( m_j = 0 \)). The model is trained to predict the masked features \( \mathbf{x}_\text{mask} \) given the observed ones \( \mathbf{x}_\text{obs} \), by maximizing the conditional likelihood:
\[
\log p_\theta(\mathbf{x}_\text{mask} \mid \mathbf{x}_\text{obs}),
\]
where \( \theta \) represents the model parameters.

During training, the model learns to infer missing values across different masking patterns by minimizing the reconstruction loss:
\[
\mathcal{L} = \mathbb{E}_{\mathbf{x}, \mathbf{m}} \left[ \| \mathbf{x}_\text{mask} - f_\theta(\mathbf{x}_\text{obs}, \mathbf{m}) \|^2 \right],
\]
where \( f_\theta \) is a neural network that predicts the masked entries. In practice, the masking pattern is varied across training samples to encourage the model to learn dependencies among all features \cite{ma2018eddi}. Once trained, synthetic data can be generated in an autoregressive or iterative fashion. For instance, one may begin with an empty or partially filled vector \( \hat{\mathbf{x}} \), and iteratively sample unobserved features conditioned on previously generated ones:
\[
\hat{x}_j \sim p_\theta(x_j \mid \hat{x}_1, \dots, \hat{x}_{j-1}),
\]
where the feature ordering can be fixed or learned dynamically. Alternatively, masked models can perform Gibbs sampling-like generation by repeatedly masking and imputing values over several iterations until convergence \cite{yoon2018gain}.

\begin{figure}[htpb]
    \centering
    \includegraphics[width=\linewidth]{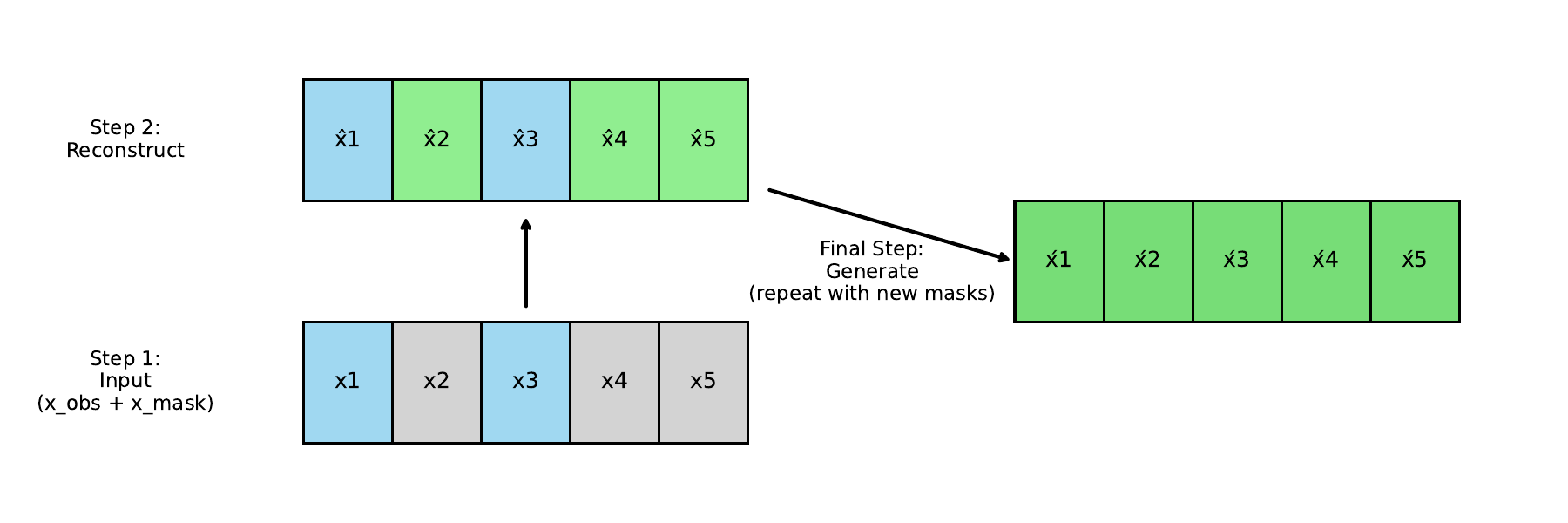}
    \caption{Illustration of the masked generative modeling process. A binary mask separates observed (blue) and masked (gray) features. The model is trained to predict the masked values (shown in green) conditioned on the observed ones using a reconstruction loss. During generation, synthetic samples are produced by iteratively imputing masked features, starting from an empty or partially observed vector, and repeating the process with new masks until a complete sample is formed.
}
    \label{fig:gen_masked}
\end{figure}

These models are particularly suited for tabular data because they can handle mixed data types, missingness, and heterogeneous feature correlations. Notable examples include GAIN (Generative Adversarial Imputation Nets) \cite{yoon2018gain} and more recent approaches such as MICE-style transformers like MaskGIT \cite{chang2022maskgit}. Figure~\ref{fig:gen_masked} shows an illustration of the masked generation process.

\subsection{Statistical Models}
Statistical models aim to capture the statistical properties and dependencies within data by approximating the distributions and relationships observed in real datasets. These models assume a specific form for the data distribution, characterized by a finite set of parameters, and focus on estimating these parameters to define the distribution. By leveraging such assumptions, they provide a structured and efficient approach to modeling data, which is particularly valuable in synthetic data generation. Controlling the distribution and dependencies of generated data is crucial, making these models well-suited for this purpose. Commonly used statistical models in synthetic data generation include Bayesian Networks (BNs) \cite{pearl2014probabilistic}, Gaussian Mixture Models (GMMs) \cite{bishop2006pattern, reynolds2009gaussian}, and Hidden Markov Models (HMMs) \cite{rabiner1989tutorial}. These models enable efficient parameter estimation and sampling, facilitating the generation of large amounts of synthetic data while preserving statistical properties such as correlations and distributions.

\subsubsection{Bayesian Networks}
Bayesian Networks (BNs) \cite{pearl2014probabilistic} are powerful probabilistic models used for modeling dependencies among a set of variables, making them useful for generating synthetic data. These networks represent conditional dependencies through a directed acyclic graph (DAG) \cite{bishop2006pattern}, where nodes represent random variables and edges represent conditional dependencies between these variables.

Let \( \mathbf{X = \{X_1, X_2, \dots, X_n\}} \) represent a set of real-world variables (data features). A Bayesian network defines the joint distribution \( P(\mathbf{X}) \) of these variables by decomposing it into a product of conditional distributions based on the structure of the network:
\begin{equation}
\mathbf{P(X) = \prod_{i=1}^{n} P(\mathbf{X}_i \mid \text{Pa}(\mathbf{X}_i))}
\label{eq:bayesian}
\end{equation}
where \( \text{Pa}(\mathbf{X}_i) \) represents the set of parent variables for node \( \mathbf{X}_i \) in the DAG.

To generate synthetic data \( \mathbf{X_{\text{syn}}} \) using a Bayesian network, the process typically involves the following steps.

\noindent\textbf{Parameter Estimation:} First, a Bayesian network is learned from the real data \( \mathbf{X} \) using parameter estimation techniques. This involves estimating the conditional probability distributions (CPDs) \( \mathbf{P(X_i} \mid \text{Pa}(\mathbf{X}_i)) \) for all nodes based on the observed data.

\begin{figure}[htpb]
    \centering
    \includegraphics[width=0.7\linewidth]{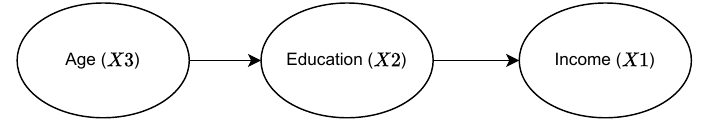}
    \caption{A simple Bayesian network showing the dependencies between age (\( \mathbf{X}_3 \)), education (\( \mathbf{X}_2 \)), and income (\( \mathbf{X}_1 \)).}
    \label{fig:BNs}
\end{figure}

For example, consider a network (Figure~\ref{fig:BNs}) with three variables: income (\( \mathbf{X}_1 \)), education (\( \mathbf{X}_2 \)), and age (\( \mathbf{X}_3 \)). The relationships between these variables can be captured using CPDs such as
\[
\mathbf{
X_1 \mid X_2 \sim \mathcal{N}(\mu_1, \sigma_1^2)},
\]
where income \( \mathbf{X}_1 \) is conditionally dependent on education \( \mathbf{X}_2 \), with a Gaussian distribution. Similarly,
\[
\mathbf{
X_2 \mid X_3 \sim \mathcal{N}(\mu_2, \sigma_2^2)},
\]
where education \( \mathbf{X}_2 \) is conditionally dependent on age \( \mathbf{X}_3 \), also modeled using a Gaussian distribution. Finally,
\[
\mathbf{
X_3 \sim \text{Uniform}(a, b)},
\]
where age \( \mathbf{X}_3 \) is assumed to follow a uniform distribution. These CPDs are estimated from the observed data, allowing the structure of the Bayesian network to reflect the dependencies between the variables.

\noindent\textbf{Sampling:} Once the parameters of the Bayesian network are estimated, synthetic data \( \mathbf{X_{\text{syn}}} \) can be generated by sampling from the conditional distributions of each node \( \mathbf{X}_i \). The process starts with the root nodes (nodes without parents). For instance, sample \( \mathbf{X'}_3 \) from \( P(\mathbf{X}_3) \), the age distribution. Then, given the sampled value \( \mathbf{X'}_3 \), sample \( \mathbf{X'}_2 \) from the conditional distribution \( P(\mathbf{X}_2 \mid \mathbf{X'}_3) \), which represents the education distribution conditioned on the sampled age. Finally, given \( \mathbf{X'}_2 \), sample \( \mathbf{X'}_1 \) from \( P(\mathbf{X}_1 \mid \mathbf{X'}_2) \), the income distribution conditioned on the sampled education:
\[
\mathbf{
X'_3 \sim P(X_3)},
\]
\[
\mathbf{X'_2 \sim P(X_2 \mid X'_3)},
\]
\[
\mathbf{X'_1 \sim P(X_1 \mid X'_2)}.
\]

\noindent\textbf{Iterative Sampling:} This sampling process is repeated for all nodes in the network, following the structure of the directed acyclic graph (DAG). Each time the synthetic data is generated, the dependencies between the variables (age, education, and income) are preserved, ensuring that the synthetic dataset reflects the statistical relationships in the real data.
After repeating this process for multiple instances, a synthetic dataset \(\mathbf{X_{\text{syn}}} = \{\mathbf{X'_1, X'_2, X'_3\}}\) would be generated, where each synthetic value corresponds to a feature in the real data \( \mathbf{X = \{X_1, X_2, X_3\} }\).

\subsubsection{Gaussian Mixture Models (GMM)}

Gaussian Mixture Models (GMMs) \cite{bishop2006pattern} are probabilistic models that represent a mixture of multiple Gaussian distributions. These models are widely used to model complex data that can be assumed to originate from several distinct Gaussian distributions. Each component in a GMM has its own mean and variance, and the data is assumed to be generated from a weighted sum of these Gaussian distributions.

The joint probability distribution \( P(\mathbf{X}) \) of a dataset \( \mathbf{X = \{X_1, X_2, \dots, X_n}\} \) under a GMM is represented as a weighted sum of the individual Gaussian distributions:
\[
\mathbf{P(X)} = \sum_{k=1}^{K} \pi_k \mathcal{N}(\mathbf{X} \mid \mu_k, \sigma_k^2),
\]
where \( \pi_k \) is the weight of the \(k\)-th Gaussian component, representing the proportion of the dataset generated by this component. \( \mathcal{N}(\mathbf{X} \mid \mu_k, \sigma_k^2) \) is the Gaussian distribution with mean \( \mu_k \) and variance \( \sigma_k^2 \). \( K \) is the number of Gaussian components in the mixture \cite{reynolds2009gaussian}.

\begin{figure}[htpb]
    \centering
    \begin{tikzpicture}
        \begin{axis}[domain=-4:4, width=10cm, height=6cm, 
                     axis lines=middle, enlarge y limits, xlabel=$\mathbf{X}$, ylabel={$\mathbf{P(X)}$}, 
                     title={Gaussian Mixture Model}, 
                     legend style={at={(0.5,-0.2)}, anchor=north, legend columns=2}]
            \addplot [blue, thick, samples=100, smooth] {0.4 * exp(-0.5 * x^2)};
            \addlegendentry{Gaussian 1}

            \addplot [red, thick, samples=100, smooth] {0.6 * exp(-0.5 * (x-2)^2)};
            \addlegendentry{Gaussian 2}

            \addplot [green, thick, samples=100, smooth] {0.4 * exp(-0.5 * x^2) + 0.6 * exp(-0.5 * (x-2)^2)};
            \addlegendentry{Mixture of Gaussians}
        \end{axis}
    \end{tikzpicture}
    \caption{A simple Gaussian Mixture Model with two Gaussian components. Data is modeled as a mixture of two Gaussian distributions. The blue curve represents the first Gaussian component with mean 0 and variance 1, while the red curve represents the second Gaussian component, centered at 2 with a variance of 1. The green curve represents the mixture of these two Gaussian components, with the blue component contributing 40\% and the red component contributing 60\% to the overall distribution.}
    \label{fig:GMM}
\end{figure}

GMMs are particularly advantageous because they do not require prior knowledge of the cluster assignments for the data points. Instead, they learn the parameters—means, covariances, and mixing coefficients—directly from the data $\mathbf{X}$. This flexibility allows GMMs to adapt to the inherent structure of the data, making them suitable for various applications, including clustering, density estimation, and anomaly detection.
By generating synthetic data 
$\mathbf{X}'$ that matches the statistical distribution of the original data 
$\mathbf{X}$, GMMs provide a robust framework for understanding and analyzing complex datasets. This capability is especially useful in scenarios where the data may not conform to a single Gaussian distribution, enabling the model to capture a richer representation of the underlying phenomena.

\noindent\textbf{Parameter Estimation:} To estimate the parameters of a GMM (the means \( \mu_k \), variances \( \sigma_k^2 \), and mixture weights \( \pi_k \) for each component), the Expectation-Maximization (EM) algorithm \cite{dempster1977maximum} is typically used. The EM algorithm iterates between two steps: the E-step and the M-step.

\noindent E-step: In this step, estimate the responsibilities \( \gamma_{ik} \), which represent the probability that data point \( \mathbf{X}_i \) was generated by the \( k \)-th Gaussian component. The responsibilities are calculated using the current estimates of the Gaussian parameters:
\[
\gamma_{ik} = \frac{\pi_k \mathcal{N}(\mathbf{X}_i \mid \mu_k, \sigma_k^2)}{\sum_{k=1}^{K} \pi_k \mathcal{N}(\mathbf{X}_i \mid \mu_k, \sigma_k^2)}.
\]
This formula calculates the posterior probability that the \( k \)-th component is responsible for generating the data point \( \mathbf{X}_i \).

\noindent M-step: In this step, we update the parameters of the GMM using the responsibilities \( \gamma_{ik} \). The updated parameters are the ones that maximize the likelihood of the data, given the responsibilities:
\[
\mu_k = \frac{\sum_{i=1}^n \gamma_{ik} \mathbf{X}_i}{\sum_{i=1}^n \gamma_{ik}}, \quad \sigma_k^2 = \frac{\sum_{i=1}^n \gamma_{ik} (\mathbf{X}_i - \mu_k)^2}{\sum_{i=1}^n \gamma_{ik}}, \quad \pi_k = \frac{1}{n} \sum_{i=1}^n \gamma_{ik}.
\]
These equations update the means, variances, and mixture weights based on the responsibilities computed in the E-step.

Once the parameters of the GMM are estimated, synthetic data can be generated by sampling from the mixture of Gaussian components. For each synthetic data point, we choose a component \( k \) with probability \( \pi_k \), and then sample a value from the corresponding Gaussian distribution \( \mathcal{N}(\mathbf{X} \mid \mu_k, \sigma_k^2) \). This can be mathematically represented as:
\[
\mathbf{X}_{\text{syn}} \sim \mathcal{N}(\mathbf{X} \mid \mu_k, \sigma_k^2) \quad \text{where} \quad k \sim \text{Categorical}(\pi_1, \pi_2, \dots,\pi_K).
\]
This process involves selecting a Gaussian component according to its weight \( \pi_k \) and then sampling a synthetic data point from the chosen Gaussian.

Suppose we want to generate synthetic data for income, education, and age, modeled using a GMM. In this case, we assume the variables age (\( \mathbf{X}_3 \)), education (\( \mathbf{X}_2 \)), and income (\( \mathbf{X}_1 \)) are modeled by a GMM as in Figure~\ref{fig:GMM}.
The income distribution \( \mathbf{X}_1 \) is modeled as a mixture of two Gaussians. The first Gaussian component (blue curve) might represent lower-income individuals, while the second component (red curve) represents higher-income individuals. The education distribution \( \mathbf{X}_2 \), conditioned on age \( \mathbf{X}_3 \), can also be modeled as a mixture of Gaussians, where different age groups are more likely to have different educational backgrounds. The age distribution \( \mathbf{X}_3 \) can be modeled as a Gaussian distribution, representing a population with a certain age distribution.

By estimating the parameters of the GMM for each of these variables and sampling from the estimated distributions, we can generate synthetic data that closely resembles the real-world data. This ensures that the relationships between the variables, as defined by the mixture of Gaussians, are preserved in the synthetic dataset.
\subsubsection{Hidden Markov Models (HMM)}
Hidden Markov Models (HMMs) \cite{rabiner1989tutorial} are probabilistic models that represent a system as a Markov process with hidden states. These models are widely used for sequential data modeling, where observations are generated by underlying hidden states that follow a Markov chain.

The joint probability distribution of a sequence of observations \( \mathbf{X = \{X_1, X_2, \dots, X_n\} }\) under an HMM is given by:
\[
\mathbf{P(X) = \sum_{z_1, \dots, z_n} P(z_1) \prod_{i=2}^{n} P(z_i \mid z_{i-1}) P(X_i \mid z_i)},
\] 
where \( \mathbf{z_i} \) represents the hidden state at time \( i \), \( \mathbf{P(z_1)} \) is the initial state probability, $\mathbf{P(z_i \mid z_{i-1})}$ is the transition probability between states, \( \mathbf{P(X_i \mid z_i)} \) is the emission probability that generates an observation \( \mathbf{X_i} \) given the state \( \mathbf{z_i} \) \cite{bishop2006pattern}.

\begin{figure}[htpb]
    \centering
    \includegraphics{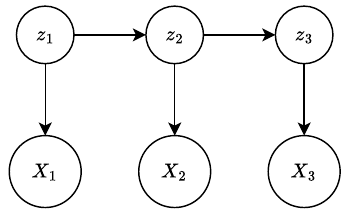}
    \caption{A simple Hidden Markov Model where hidden states \( \mathbf{z} \) generate observations \( \mathbf{X} \).}
    \label{fig:Hmm}
\end{figure}

The structure and flow of an HMM are depicted in Figure \ref{fig:Hmm}. In this diagram, each hidden state \( \mathbf{z_i} \) generates an observation \( \mathbf{X_i} \), and the transitions between the hidden states follow a Markov process, as indicated by the arrows.

The sequence of hidden states \( \mathbf{z_1, z_2, \dots, z_n }\) generates the corresponding sequence of observations \(\mathbf{ X_1, X_2, \dots, X_n} \). The model starts with an initial hidden state \( \mathbf{z_1} \), and then the hidden state at each time step is determined by the transition probabilities \( \mathbf{P(z_i \mid z_{i-1}) }\), while the corresponding observation \( \mathbf{X_i }\) is generated based on the emission probabilities \( \mathbf{P(X_i \mid z_i) }\).

\noindent \textbf{Parameter Estimation}
The parameters of an HMM (transition probabilities, emission probabilities, and initial state probabilities) are typically estimated using the Expectation-Maximization (EM) algorithm \cite{dempster1977maximum}, specifically the Baum-Welch algorithm \cite{baum1970maximization}.

\noindent\underline{1) E-step:} In this step, we estimate the posterior probabilities of the hidden states given the observations using the Forward-Backward algorithm:
\[
\mathbf{\gamma_{ik} = \frac{P(z_i = k \mid X)}{\sum_{j} P(z_i = j \mid X)}}.
\]
This represents the probability that the hidden state at time \( i \) is \( k \) given the observations.

\noindent\underline{2) M-step:} In this step, we update the transition probabilities, emission probabilities, and initial probabilities based on the expected values computed in the E-step:
\[
\mathbf{P(z_i = k \mid z_{i-1} = j) = \frac{\sum_{i=1}^{n} \gamma_{i-1,j} P(z_i = k \mid z_{i-1} = j)}{\sum_{i=1}^{n} \gamma_{i-1,j}}}.
\]

Once the parameters of the HMM are estimated using the EM algorithm, synthetic data can be generated by sampling from the learned model. The synthetic data sequence \( \mathbf{X_{\text{syn}} = \{X'_1, X'_2, X'_3\}} \) is generated as follows:
\begin{enumerate}
    \item Sample the initial state \( \mathbf{z_1} \) from \( \mathbf{P(z_1)} \).
    \item For each time step \( i \):
    \begin{itemize}
        \item Sample \( \mathbf{z_i} \) from \( \mathbf{P(z_i \mid z_{i-1})} \).
        \item Sample \( \mathbf{X'_i} \) from \( \mathbf{P(X'_i \mid z_i)} \).
\end{itemize}
\end{enumerate}
This process leverages the estimated parameters from the EM algorithm to ensure that the generated sequences maintain the statistical dependencies of the original data. Suppose we want to generate synthetic data for a speech recognition task. The hidden states represent phonemes, and the observations are audio feature vectors. By estimating the parameters of the HMM, we can generate realistic synthetic speech sequences that follow the learned probabilistic patterns.

Hidden Markov Models (HMMs) are widely used for modeling sequential data, with applications in speech recognition, natural language processing, and time-series analysis. Using the EM algorithm, we can estimate HMM parameters and generate synthetic data sequences \( \mathbf{X_{\text{syn}} } \) that closely resemble real-world sequences.

\subsubsection{Hybrid Probabilistic Approaches.}
While copula-based methods are effective in preserving marginal distributions and Bayesian Networks (BNs) provide interpretable dependency structures, each approach in isolation is limited in capturing complex non-linear interactions across features. This has motivated hybrid approaches that combine structured probabilistic modeling with discriminative refinement.

A representative example is \textbf{Synthsonic} \cite{baak2022synthsonic}, which decomposes tabular data generation into sequential modeling stages. Given tabular data $\mathbf{X}$, each feature is first transformed into a copula space using its empirical cumulative distribution function (CDF), i.e., $U_j = F_j(X_j)$, resulting in uniformly distributed marginals. This transformation enables the separation of marginal distributions from dependency structure.

A tree-structured Bayesian Network is then learned over the transformed variables, modeling the joint distribution as:
\begin{equation*}
P(\mathbf{U}) = \prod_{i=1}^{d} P(U_i \mid \text{Pa}(U_i)),
\end{equation*}
where $\text{Pa}(U_i)$ denotes the parent set of feature $U_i$, typically obtained using Chow--Liu or Tree-Augmented Naive Bayes (TAN) structures. Synthetic samples are generated via forward sampling and mapped back to the original space through the inverse transformation. While the BN effectively captures dominant pairwise dependencies, it is limited in representing higher-order and non-linear interactions. To address this, Synthsonic incorporates a calibrated discriminative model that refines the generated samples, correcting residual discrepancies between real and synthetic distributions.

From a taxonomy perspective, Synthsonic models feature dependencies through explicit BN factorization while preserving statistical properties via copula-based marginal transformation, with residual higher-order interactions handled through discriminative correction. However, the reliance on tree-structured dependencies limits its ability to fully capture complex multi-way relationships, and performance may degrade in highly imbalanced data settings. In contrast to GAN- and Transformer-based models that learn dependencies implicitly in a unified framework, Synthsonic separates marginal modeling, dependency learning, and residual correction into distinct stages, providing improved interpretability and computational efficiency.

\subsubsection{Large Language Models for Tabular Data Generation}
\colorred{Large language models generate tabular data by converting each row into a token sequence containing column names and values. For a serialization \(w_{1:T}\), an autoregressive LLM represents the row probability as \(p_{\theta}(w_{1:T})=\prod_{t=1}^{T}p_{\theta}(w_t\mid w_{<t})\). Because tabular columns do not generally possess a natural left-to-right order, the serialization and column-order strategy affect which conditional relationships are encountered during training and sampling. Numerical values are also represented as tokens rather than as continuous quantities, making numerical precision and invalid-value handling dependent on the tokenizer and the decoding procedure.} \colorred{GReaT \cite{borisov2022language} fine-tunes an autoregressive language model on textual representations of rows and randomizes feature order to reduce dependence on a fixed column sequence. It can begin generation from a prompt containing selected column--value pairs, thereby supporting conditional synthesis. Pred-LLM \cite{nguyen2024generating} extends this approach with a permutation strategy during fine-tuning, feature-conditional sampling, and a separate prompting step for target-label generation. Its design specifically addresses the loss of feature--target correlation that can occur when labels are generated as ordinary tokens in an unconstrained sequence.} \colorred{HARMONIC \cite{wang2024harmonic} constructs instruction-fine-tuning examples using nearest-neighbour rows so that the LLM learns local inter-row relationships and table format. It also introduces DLT for testing leakage from an LLM-based generator. Prism \cite{guan2026prism} addresses a different requirement: formal privacy for relational-data synthesis. It performs differentially private aggregation of teacher-LLM outputs before training a student generator. Accordingly, HARMONIC's distance- and attack-based measurements must not be interpreted as equivalent to Prism's explicit differential-privacy accounting.}

\colorred{The use of a language-modeling objective does not automatically make a method an LLM-based generator. MaCoDE \cite{an2025masked}, for example, is more appropriately classified as a masked generative model because its defining mechanism is orderless masked conditional-density estimation. In the taxonomy, GReaT, Pred-LLM, HARMONIC, and Prism should therefore be placed in an explicit LLM column, while MaCoDE should remain in the masked-model column. Across these methods, unresolved limitations include serialization sensitivity, inefficient token-level representation of high-precision numerical values, invalid or out-of-range outputs, computational cost, and the absence of formal privacy unless the complete synthesis mechanism is accompanied by auditable privacy accounting.}

\subsection{Data Types}
\label{app:data_types}
Synthetic data generation spans various data types, each with unique structures, properties, and challenges. While synthetic data techniques apply to a broad range of domains, this survey emphasizes tabular data due to its structured nature and widespread use in critical applications such as finance \cite{assefa2020generating}, healthcare \cite{hernandez2022synthetic}, and risk assessment \cite{dal2017credit}. Unlike unstructured data types, tabular data consists of well-defined attributes, often mixing numerical and categorical variables, which necessitates generative models capable of accurately capturing feature dependencies, distributions, and correlations.

Beyond tabular data, synthetic data generation extends to unstructured and semi-structured formats, including time series, images, text, graphs, audio, and three-dimensional data.
Time-series data introduce the complexity of sequential dependencies, where models must preserve trends and temporal relationships \cite{box2015time}. This is particularly important in applications such as financial forecasting and healthcare monitoring, where synthetic sequences should retain meaningful patterns. Image data, on the other hand, presents challenges in high-dimensional space, requiring models to capture intricate spatial features. Synthetic image generation has seen significant advancements with deep learning techniques, particularly GANs \cite{goodfellow2014generative} and diffusion models \cite{kotelnikov2023tabddpm}, which have been applied in areas such as medical imaging, security, and artistic content generation.

Text generation relies on language models to produce syntactically and semantically coherent sequences, making it a distinct challenge in natural language processing. The rise of transformer-based architectures such as GPT \cite{radford2018improving} and BERT \cite{devlin2018bert} has enabled more sophisticated synthetic text applications, including document synthesis and conversational AI. Graph-based data requires maintaining connectivity and structural properties, often tackled with generative graph neural networks \cite{scarselli2008graph}. Audio and 3D data synthesis further expand the landscape, supporting applications in speech synthesis, augmented reality, and virtual simulations.
\section{Taxonomy}
Table~\ref{tab:taxonomy} summarizes the reviewed methods by
model family and generation objective, covering feature
dependencies, statistical preservation, privacy, and
conditioning on specific attributes.
\begin{table}[!htbp]
\centering
\small

\caption{Overview of Synthetic Data Generation Methods for Tabular Data.
This table presents a comprehensive overview of various synthetic data
generation models for tabular data, elucidating how each method addresses
critical aspects, including feature dependencies \textbf{(FDs)},
preservation of statistical properties \textbf{(PSPs)}, conditioning on
specific attributes \textbf{(CSAs)}, and privacy preservation
\textbf{(PP)}. The methods are categorized into Traditional
\colorred{(Tr)} and Modern approaches including
\colorred{generative adversarial networks (GAN), Diffusion (Df),}
Autoregressive models (AR)\colorred{, Masked models (Mk), and
large language models (LLM)}. Traditional methods include techniques
like Autoencoders (AEs), Variational Autoencoders (VAEs), SMOTE,
Gaussian Mixture Models (GMMs), Markov Chains, Bayesian Networks,
and other statistical models, which rely on probabilistic or heuristic
methods for data generation and feature modeling. The symbols in the
table represent the following: a single tick mark (\(\checkmark\))
signifies that the method provides an enhanced or specialized treatment,
such as more robust handling of that characteristic.
An (\(\times\)) signifies that the method does not explicitly address
the characteristic.}
\label{tab:taxonomy}

\begin{tabularx}{\linewidth}{
|>{\raggedright\arraybackslash}X|
*{10}{c|}}
\hline
\rowcolor{tabheadcolor}
& \multicolumn{4}{c|}{\textbf{Characteristics}}
& \multicolumn{6}{c|}{\textbf{Methods and Models}} \\
\cline{2-11}
\rowcolor{tabheadcolor}
\multirow{-2}{*}{\textbf{Approaches}} &
\textbf{FDs} &
\textbf{PSPs} &
\textbf{CSAs} &
\textbf{PP} &
\textbf{Tr} &
\textbf{GAN} &
\textbf{Df} &
\textbf{AR} &
\textbf{Mk} &
\textbf{LLM} \\
\hline


\colorred{GReaT \cite{borisov2022language}} &
\colorred{\(\checkmark\)} & &
\colorred{\(\checkmark\)} &
\colorred{\(\times\)} &
& & & & & \colorred{\(\checkmark\)} \\
\hline

\colorred{Pred-LLM \cite{nguyen2024generating}} &
\colorred{\(\checkmark\)} & &
\colorred{\(\checkmark\)} &
\colorred{\(\times\)} &
& & & & & \colorred{\(\checkmark\)} \\
\hline

\colorred{HARMONIC \cite{wang2024harmonic}} &
& \colorred{\(\checkmark\)} &
\colorred{\(\times\)} &
\colorred{\(\checkmark\)} &
& & & & & \colorred{\(\checkmark\)} \\
\hline

\colorred{Prism \cite{guan2026prism}} &
& & \colorred{\(\times\)} &
\colorred{\(\checkmark\)} &
& & & & & \colorred{\(\checkmark\)} \\
\hline


TabDDPM \cite{kotelnikov2023tabddpm} &
\(\checkmark\) & & & \(\times\) &
& & \(\checkmark\) & & \(\checkmark\) & \\
\hline

STaSy \cite{kim2022stasy} &
\(\checkmark\) & \(\checkmark\) & \(\times\) & \(\times\) &
& & \(\checkmark\) & & & \\
\hline

TABSYN \cite{zhang2023mixed} &
\(\checkmark\) & \(\checkmark\) & \(\times\) & &
\(\checkmark\) & & \(\checkmark\) & \(\checkmark\) & & \\
\hline

Jolicoeur et al., 2024 \cite{jolicoeur2024generating} &
& & & \(\times\) &
& & \(\checkmark\) & & & \\
\hline

\colorred{TabDiff \cite{shi2025tabdiff}} &
\colorred{\(\checkmark\)} &
\colorred{\(\checkmark\)} &
\colorred{\(\checkmark\)} &
\colorred{\(\times\)} &
& & \colorred{\(\checkmark\)} & & & \\
\hline

Binary Diffusion \cite{kinakh2024tabular} &
\(\checkmark\) & \(\checkmark\) & \(\checkmark\) & \(\times\) &
& & \(\checkmark\) & & & \\
\hline


CTGAN \cite{xu2019modeling} &
\(\checkmark\) & \(\times\) & \(\checkmark\) & \(\times\) &
& \(\checkmark\) & & & & \\
\hline

FCT-GAN \cite{zhao2022fct} &
\(\checkmark\) & & & \(\times\) &
& \(\checkmark\) & & & & \\
\hline

CTAB-GAN \cite{zhao2021ctab} &
\(\checkmark\) & \(\checkmark\) & \(\checkmark\) & \(\times\) &
& \(\checkmark\) & & & & \\
\hline

PATE-GAN \cite{jordon2018pate} &
& & & \(\checkmark\) &
& \(\checkmark\) & & & & \\
\hline

IT-GAN \cite{lee2021invertible} &
& & \(\times\) & \(\checkmark\) &
& \(\checkmark\) & & & & \\
\hline

TabFairGAN \cite{rajabi2022tabfairgan} &
& & & \(\times\) &
& \(\checkmark\) & & & & \\
\hline

Park et al., 2018 \cite{park2018data} &
& & & \(\times\) &
& \(\checkmark\) & & & & \\
\hline

FinGAN \cite{kate2022fingan} &
\(\times\) & \(\times\) & \(\times\) & \(\times\) &
& \(\checkmark\) & & & & \\
\hline

medGAN \cite{choi2017generating} &
& \(\checkmark\) & & \(\times\) &
& \(\checkmark\) & & & & \\
\hline

Che et al., 2017 \cite{che2017boosting} &
& \(\checkmark\) & & &
& \(\checkmark\) & & & & \\
\hline

Abay et al., 2022 \cite{abay2019privacy} &
\(\times\) & & \(\times\) & \(\checkmark\) &
\(\checkmark\) & \(\checkmark\) & & & & \\
\hline

Acs et al., 2018 \cite{acs2018differentially} &
\(\times\) & & \(\times\) & \(\checkmark\) &
\(\checkmark\) & \(\checkmark\) & & & & \\
\hline

DP-CTGAN \cite{fang2022dp} &
\(\checkmark\) & & \(\checkmark\) & \(\checkmark\) &
& \(\checkmark\) & & & & \\
\hline

Avi{\~n}{\'o} et al., 2018 \cite{avino2018generating} &
& & \(\times\) & \(\times\) &
& \(\checkmark\) & & & & \\
\hline


TabTransformer \cite{huang2020tabtransformer} &
\(\checkmark\) & & & \(\times\) &
& & & \(\checkmark\) & & \\
\hline

TabMT \cite{gulati2023tabmt} &
\(\checkmark\) & & & \(\times\) &
& & & \(\checkmark\) & \(\checkmark\) & \\
\hline


\colorred{MaCoDE \cite{an2025masked}} &
\colorred{\(\checkmark\)} &
\colorred{\(\checkmark\)} &
\colorred{\(\checkmark\)} &
\colorred{\(\times\)} &
& & & & \colorred{\(\checkmark\)} & \\
\hline


Han et al., 2019 \cite{han2005borderline} &
\(\times\) & \(\times\) & \(\times\) & \(\times\) &
\(\checkmark\) & & & & & \\
\hline

He et al., 2019 \cite{he2008adasyn} &
\(\times\) & \(\times\) & \(\times\) & \(\times\) &
\(\checkmark\) & & & & & \\
\hline

PrivBayes \cite{zhang2017privbayes} &
& \(\checkmark\) & & \(\checkmark\) &
\(\checkmark\) & & & & & \\
\hline

DataSynthesizer \cite{ping2017datasynthesizer} &
& & \(\checkmark\) & \(\checkmark\) &
\(\checkmark\) & & & & & \\
\hline

Copula Flows \cite{kamthe2021copula} &
\(\checkmark\) & \(\checkmark\) & & \(\times\) &
\(\checkmark\) & & & & & \\
\hline

EHR-Safe \cite{yoon2023ehr} &
\(\checkmark\) & & & \(\checkmark\) &
\(\checkmark\) & & & & & \\
\hline

McKenna et al., 2022 \cite{mckenna2022aim} &
& & \(\times\) & \(\checkmark\) &
\(\checkmark\) & & & & & \\
\hline

Liu et al., 2021 \cite{liu2021iterative} &
& & \(\times\) & \(\checkmark\) &
\(\checkmark\) & & & & & \\
\hline

Synthsonic \cite{baak2022synthsonic} &
\(\checkmark\) & \(\checkmark\) & \(\times\) & \(\times\) &
\(\checkmark\) & & & & & \\
\hline

\end{tabularx}
\end{table}
\section{Evaluation: Metrics for the existing models}

To summarize the evaluation metrics used in prior literature, it is crucial to assess synthetic data generation models across key aspects such as data similarity, privacy preservation, and machine learning utility. These evaluations typically focus on categories like feature dependency, statistical similarity, and privacy preservation, alongside performance on machine learning tasks such as classification and regression. The metrics employed vary across models, reflecting their specific design objectives and intended use cases.

\paragraph{Feature Dependency} is one of the key considerations when evaluating synthetic data. It is essential to determine how well the synthetic data preserves the relationships between features found in the original dataset. 
Feature dependency preservation is evaluated by comparing the pairwise correlations between the features in the synthetic data and the original data. This is typically done using the Pearson correlation coefficient for continuous features, the uncertainty coefficient for categorical features, and the correlation ratio for mixed-type data. The Pearson correlation between two continuous variables \(\mathbf{x_1}\) and \(\mathbf{x_2}\) is calculated as:
\begin{equation*}
r_{\mathbf{x}_1,\mathbf{x}_2}
=
\frac{\operatorname{Cov}(\mathbf{x}_1,\mathbf{x}_2)}
{\sigma_{\mathbf{x}_1}\sigma_{\mathbf{x}_2}},
\end{equation*}
where \(\operatorname{Cov}(\mathbf{x}_1,\mathbf{x}_2)\) represents the covariance between \(\mathbf{x_1}\) and \(\mathbf{x_2}\), and \(\sigma_{\mathbf{x_1}}\) and \(\sigma_{\mathbf{x_2}}\) are the standard deviations of \(\mathbf{x_1}\) and \(\mathbf{x_2}\), respectively. The closer this value is to 1 or -1, the stronger the linear relationship between the two variables.

\paragraph{Statistical Similarity} is another important evaluation criterion in synthetic data generation. 
Various divergence measures quantify the difference between the distributions of real and synthetic data. Two key metrics for this evaluation are the Jensen-Shannon Divergence (JSD) \cite{lin1991divergence} and the Wasserstein Distance (WD) \cite{ramdas2017wasserstein}. The JSD is used to measure the difference between the probability distributions of categorical features in real and synthetic data. \colorred{The Jensen--Shannon divergence between distributions
\(P\) and \(Q\) is given in Eq.~\ref{eq:JSD}:}
\begin{equation}
\text{JSD}(P, Q) = \frac{1}{2} \left( D_{\text{KL}}(P \parallel M) + D_{\text{KL}}(Q \parallel M) \right), 
\label{eq:JSD}
\end{equation}
where \(M = \frac{1}{2}(P + Q)\), and \(D_{\text{KL}}\) is the Kullback-Leibler divergence.

The Wasserstein Distance is often used for continuous variables, as it is numerically more stable than the JSD, especially when there is no overlap between the synthetic and original data distributions. The Wasserstein Distance between two distributions \(P\) and \(Q\) is defined as:
\begin{equation*}
W(P, Q) = \inf_{\gamma \in \Gamma(P, Q)} \mathbb{E}_{(x, y) \sim \gamma} \left[ \|x - y\| \right], 
\end{equation*}
where \(\Gamma(P, Q)\) is the set of all possible couplings of \(P\) and \(Q\), and \(\|x - y\|\) represents the distance between two points in the sample space.

\paragraph{Privacy Preservation}
can be evaluated using metrics such as the \textit{privacy loss} and the \textit{privacy-utility trade-off}. Differential privacy ensures that an individual's contribution to the dataset remains indistinguishable, even if the synthetic data is used in downstream tasks \cite{dwork2016calibrating}. Formally, a mechanism \( M \) satisfies \(\epsilon\)-differential privacy if, for all neighboring datasets \( D_1 \) and \( D_2 \) (differing by at most one individual), and for all outputs \( S \subseteq \text{Range}(M) \):
\[
\frac{\Pr[M(D_1) \in S]}{\Pr[M(D_2) \in S]} \leq e^\epsilon.
\]
Here, \(\epsilon\) is the \textit{privacy loss}, which quantifies the strength of the privacy guarantee \cite{dwork2014algorithmic}. Lower values of \(\epsilon\) correspond to stronger privacy, as they limit the influence of any single individual on the output. The privacy loss is typically controlled by adding noise to the model's outputs using mechanisms such as the \textit{Laplace noise} or \textit{Gaussian noise} \cite{dwork2016calibrating}. For example, the Laplace mechanism adds noise drawn from the Laplace distribution:

\[
M(D) = f(D) + \text{Laplace}\left(0, \frac{\Delta f}{\epsilon}\right),
\]
where \(\Delta f\) is the sensitivity of the query \( f \), defined as the maximum change in the output when one individual is added or removed from the dataset \cite{dwork2014algorithmic}. Similarly, the Gaussian mechanism adds noise drawn from a Gaussian distribution:

\[
M(D) = f(D) + \mathcal{N}\left(0, \sigma^2\right),
\]
where \(\sigma\) is scaled based on \(\epsilon\) and the sensitivity \(\Delta f\) \cite{dwork2014algorithmic}.
The \textit{privacy budget} \(\epsilon\) represents the total allowable privacy loss across multiple queries or iterations \cite{dwork2016calibrating}. A smaller \(\epsilon\) provides stronger privacy guarantees but may reduce the utility of the output due to increased noise. This trade-off is critical in applications where both privacy and data usefulness are important. PATE-GAN enforces strict privacy constraints while maintaining a high level of utility, making it a suitable model for privacy-sensitive applications. To evaluate utility, metrics such as the \textit{Area Under the Receiver Operating Characteristics Curve (AUROC)} and the \textit{Area Under the Precision-Recall Curve (AUPRC)} are often used \cite{jordon2018pate}. AUROC measures how well a model distinguishes between classes, while AUPRC is particularly useful for evaluating performance on imbalanced datasets \cite{jordon2018pate}. These metrics help quantify the effectiveness of synthetic data in training privacy-preserving models while maintaining high utility.

\paragraph{Machine Learning Utility} measures the effectiveness of synthetic data for downstream tasks like classification and regression. 
ML utility is often evaluated by training machine learning classifiers or regressors on the synthetic data and comparing their performance to that of models trained on the original data. Metrics such as accuracy, F1-score, and AUC are used for classification, while for regression tasks, mean absolute percentage error (MAPE), explained variance score (EVS), and R² score are used. These metrics help determine whether the synthetic data retains the predictive power of the original data. For example, the F1-score, which is a harmonic mean of precision and recall, is calculated as:

\begin{equation*}  
F_1 = 2 \cdot \frac{\text{precision} \cdot \text{recall}}{\text{precision} + \text{recall}},
\end{equation*}
where precision and recall are defined as:

\begin{equation*}  
\text{precision} = \frac{TP}{TP + FP}, \quad \text{recall} = \frac{TP}{TP + FN},
\end{equation*}
with \(TP\), \(FP\), and \(FN\) representing the true positives, false positives, and false negatives, respectively. The AUC measures the area under the ROC curve, which plots the true positive rate against the false positive rate at various threshold settings. In regression tasks, the R² score quantifies the proportion of variance in the target variable that is predictable from the features:

\begin{equation*}
    R^2 = 1 - \frac{\sum_{i=1}^{n} (y_i - \hat{y}_i)^2}{\sum_{i=1}^{n} (y_i - \bar{y})^2},
\end{equation*}
where \(y_i\) represents the true values, \(\hat{y}_i\) are the predicted values, and \(\bar{y}\) is the mean of the true values. Similarly, MAPE is defined as:
\begin{equation*}
    MAPE = \frac{1}{n} \sum_{i=1}^{n} \left| \frac{y_i - \hat{y}_i}{y_i} \right| \times 100\% ,
\end{equation*}
which calculates the average absolute percentage error between the predicted and true values.

Overall, these findings underscore the importance of selecting the right model based on the specific requirements of the task, whether that be for better privacy preservation, statistical similarity, or machine learning utility, as no one model outperforms others across all categories.
Table \ref{tab:eval_metrics_comparison} provides the summary of the evaluation metrics for the models discussed.

\begin{table*}[htpb]
\centering
\caption{Summary of Evaluation Metrics for Synthetic Data Generation Models}
\label{tab:eval_metrics_comparison}
\begin{tabular}{|>{\raggedright\arraybackslash}p{5cm}|>{\raggedright\arraybackslash}p{8cm}|}
\hline
\textbf{Metric} & \textbf{Method} \\ \hline
\textbf{Feature Dependency} & 
Pearson Correlation (Continuous), Uncertainty Coefficient (Categorical), Correlation Ratio (Mixed) \\ \hline
\textbf{Statistical Similarity} & 
Jensen-Shannon Divergence (JSD), Wasserstein Distance (WD), Area Under the Receiver Operating Characteristics Curv (AUROC), Area Under the
Precision-Recall Curve (AUPRC). 
\\ \hline
\textbf{Privacy Preservation} & Model Inversion attacks, Membership-inference attacks,
Privacy Loss (\(\epsilon\)), Privacy-Utility Trade-off (\(\epsilon\) vs. AUROC/AUPRC)\\ \hline
\textbf{Machine Learning Utility} & 
Accuracy, F1-score, AUC, MAPE, EVS, R² (Regression),AUROC, AUPRC.
\\ \hline
\end{tabular}
\end{table*}
\section{Model Architectures \& Plots}
\renewcommand{\thefigure}{A.\arabic{figure}}
\setcounter{figure}{0}
\begin{figure*}[htpb]
\centering\includegraphics[width=0.7\linewidth]{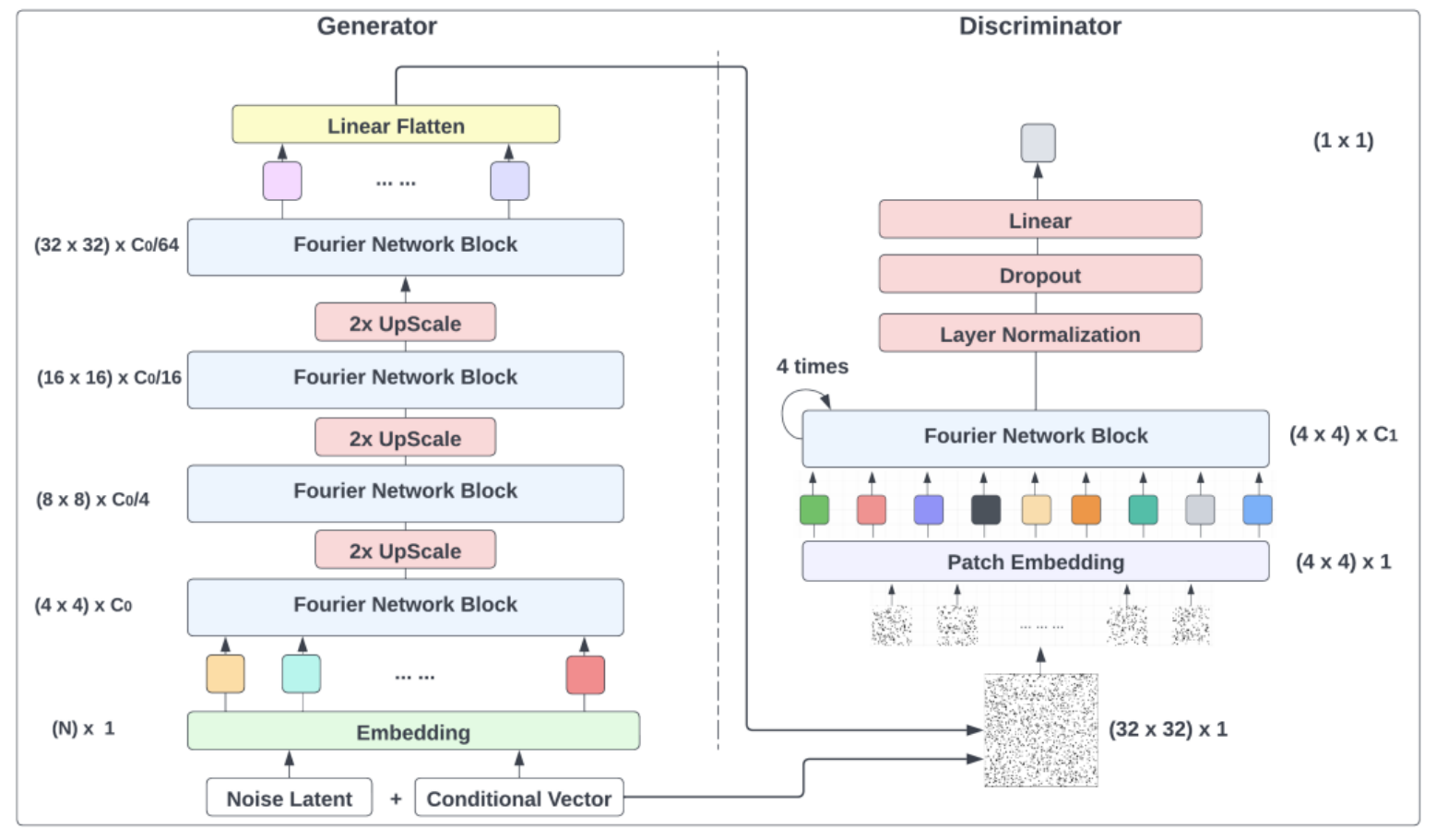}
    \Description{ Structure of FCT-GAN. \cite{zhao2022fct}}
    \caption{ Structure of FCT-GAN. \cite{zhao2022fct}}
    \label{fig:fct_gan}
\end{figure*}
\begin{figure}[t] 
    \centering
    \includegraphics[width=0.7\linewidth]{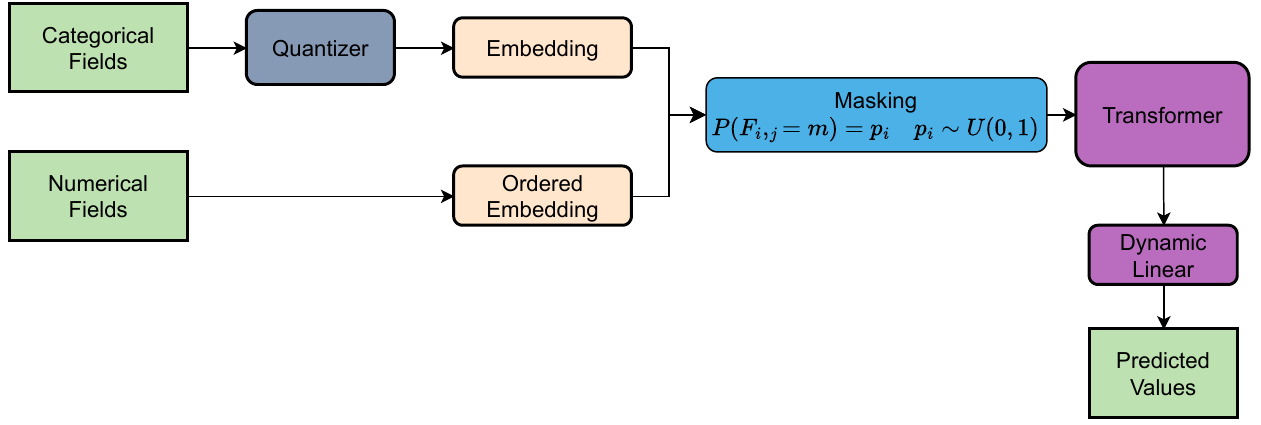}
    \caption{Diagram of TabMT. \( m \) is the mask token, and \( p_i \) is the masking probability of the \( i \)-th row. \cite{gulati2023tabmt}}
    \Description{Diagram of TabMT. \( m \) is the mask token, and \( p_i \) is the masking probability of the \( i \)-th row. \cite{gulati2023tabmt}}
    \label{fig:tabmt}
\end{figure}
\begin{figure*}[t] 
    \centering
    \includegraphics[width=\linewidth]{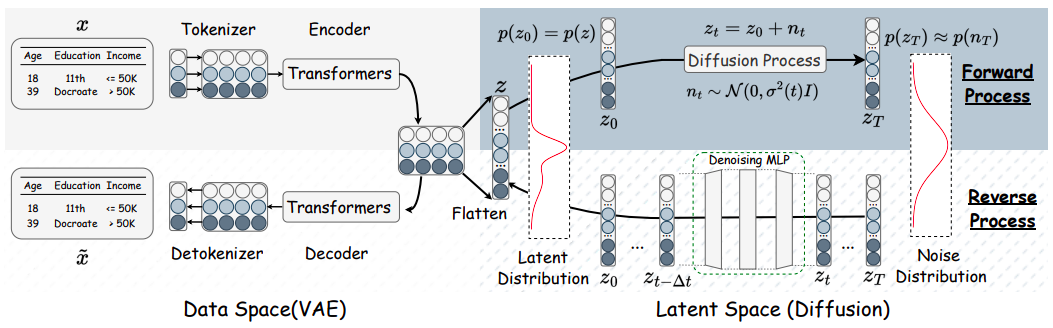}
    \Description{An overview of the proposed TABSYN. Each row of tabular data \(\mathbf{x}\) is mapped to a latent space \(z\) via a column-wise tokenizer and an encoder. A diffusion process \(z_0 \rightarrow z_T\) is applied in the latent space. Synthesis \(z_T \rightarrow z_0\) starts from the base distribution \(p(z_T)\) and generates samples \(z_0\) in the latent space through a reverse process. These samples are then mapped from the latent space \(z\) to the data space \(\tilde{\mathbf{x}}\) using a decoder and a detokenizer \cite{zhang2023mixed}.}
    \caption{An overview of the proposed TABSYN. Each row of tabular data \(\mathbf{x}\) is mapped to a latent space \(z\) via a column-wise tokenizer and an encoder. A diffusion process \(z_0 \rightarrow z_T\) is applied in the latent space. Synthesis \(z_T \rightarrow z_0\) starts from the base distribution \(p(z_T)\) and generates samples \(z_0\) in the latent space through a reverse process. These samples are then mapped from the latent space \(z\) to the data space \(\tilde{\mathbf{x}}\) using a decoder and a detokenizer \cite{zhang2023mixed}.}
    \label{fig:tabsyn}
\end{figure*}
\begin{figure}[t] 
\centering\includegraphics[width=0.45\linewidth]{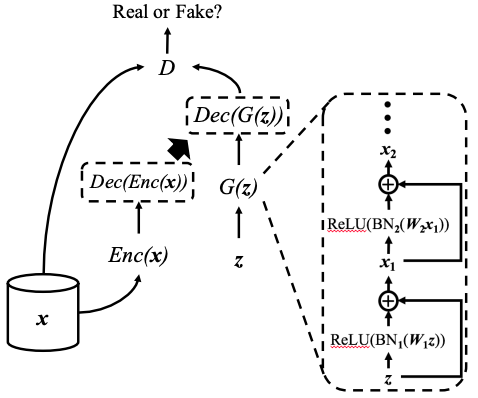}
    \Description{Architecture of medGAN: The discrete $\mathbf{x}$ comes from the source EHR data, $\mathbf{z}$ is the random prior for the generator $G$; $G$ is a feedforward network with shortcut connections (right-hand side figure); An autoencoder (i.e., the encoder $\text{Enc}$ and decoder $\text{Dec}$) is learned from $\mathbf{x}$; The same decoder $\text{Dec}$ is used after the generator $G$ to construct the discrete output. The discriminator $D$ tries to differentiate real input $\mathbf{x}$ and discrete synthetic output $\text{Dec}(G(\mathbf{z}))$ \cite{choi2017generating}.}
    \caption{Architecture of medGAN: The discrete $\mathbf{x}$ comes from the source EHR data, $\mathbf{z}$ is the random prior for the generator $G$; $G$ is a feedforward network with shortcut connections (right-hand side figure); An autoencoder (i.e., the encoder $\text{Enc}$ and decoder $\text{Dec}$) is learned from $\mathbf{x}$; The same decoder $\text{Dec}$ is used after the generator $G$ to construct the discrete output. The discriminator $D$ tries to differentiate real input $\mathbf{x}$ and discrete synthetic output $\text{Dec}(G(\mathbf{z}))$ \cite{choi2017generating}.}
    \label{fig:medGAN}
\end{figure}
\begin{figure}[htpb]
    \centering
    \includegraphics[width=0.7\linewidth]{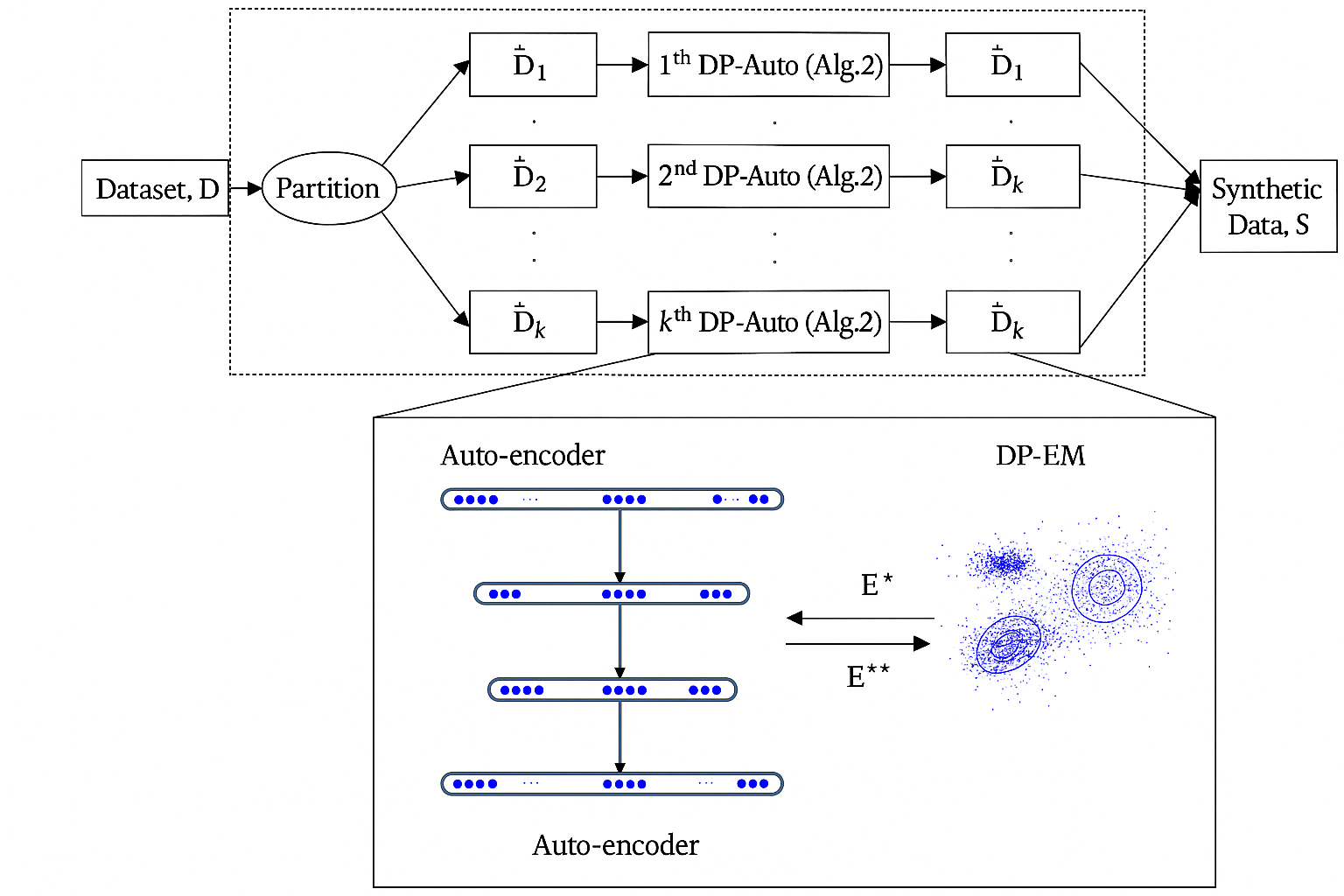}
    \Description{Differentially private synthetic data generation}
    \caption{Differentially private synthetic data generation \(DP-SYN\) \cite{abay2019privacy}.}
    \label{fig:dpsyn}
\end{figure} 
\begin{figure}[t!] 
    \centering
    \centering\includegraphics[width=0.7\linewidth]{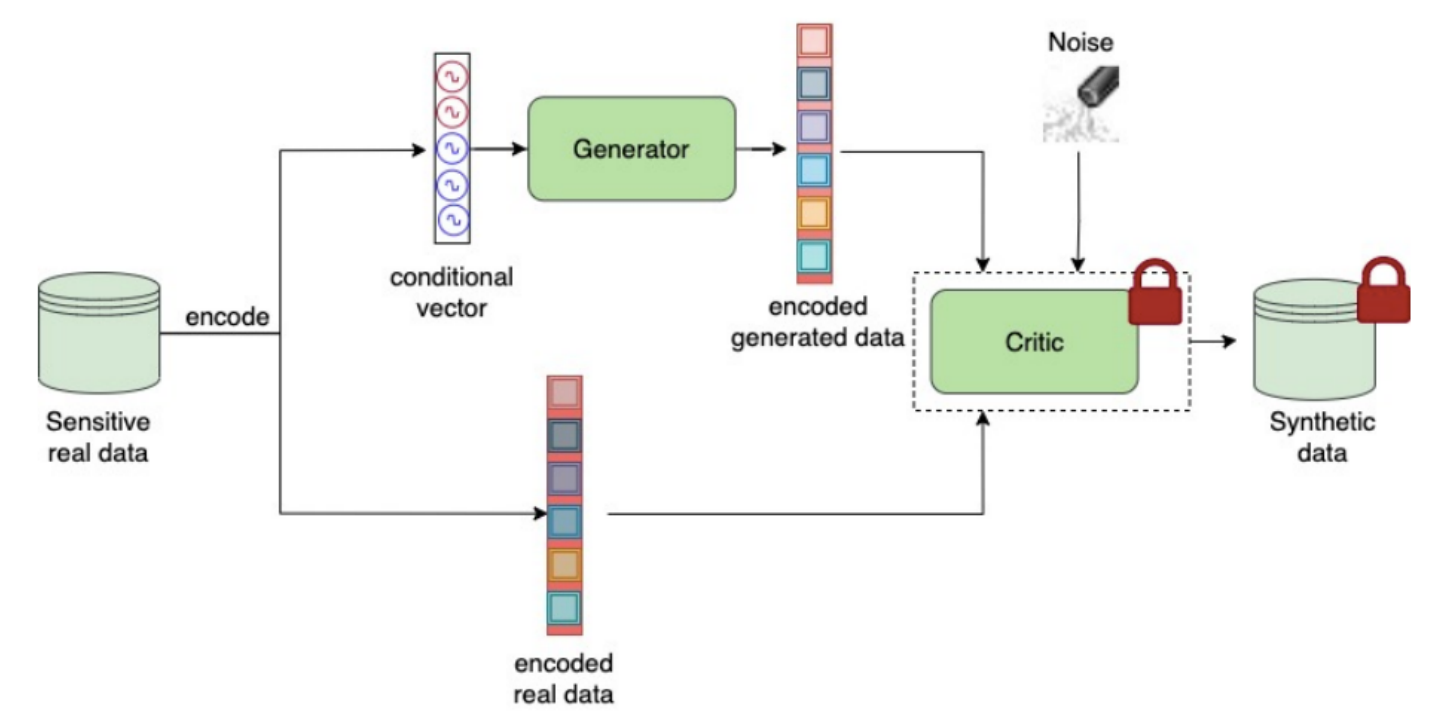}
    \caption{DP-CTGAN. Sensitive training data is fed into a conditional generator to generate samples which can evenly cover
all possible discrete values. At the same time, random perturbation is added to the critic to enforce privacy protection \cite{fang2022dp}.}
    \label{fig:dpctgan}
\end{figure}
\begin{figure}[t!]    \centering\includegraphics[width=0.7\linewidth]{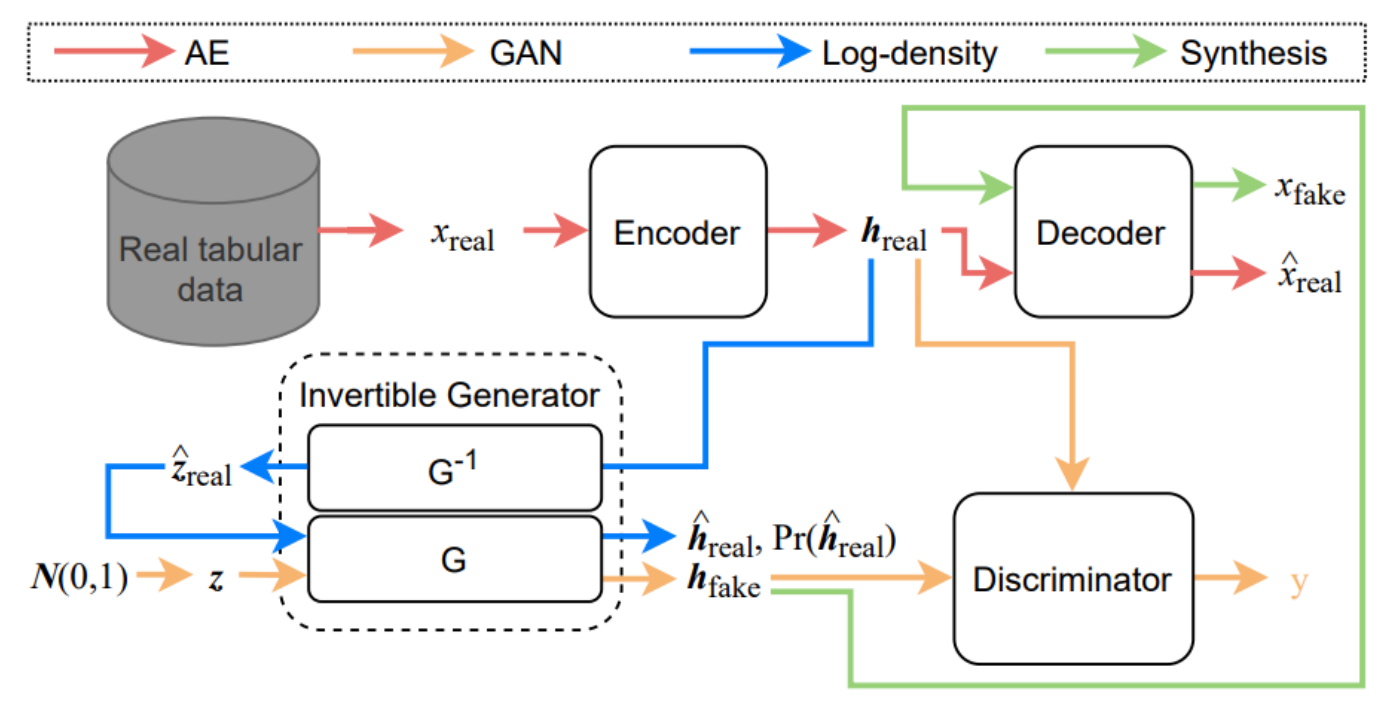}
    \caption{The overall architecture of IT-GAN. Each edge color means
a certain type of data path \cite{lee2021invertible}.}
    \label{fig:itgan}
\end{figure}

\begin{figure}[htpb]
    \centering
    
    \begin{subfigure}[t]{0.48\textwidth}
        \centering
        \includegraphics[width=0.8\linewidth]{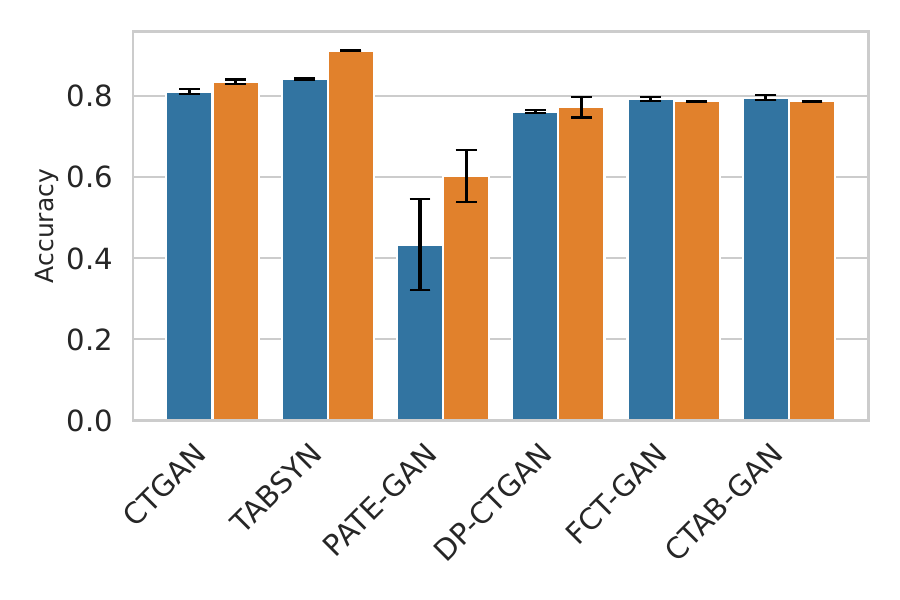}
        \vspace{-0.2cm}
        \caption{Model Inversion Attack - Accuracy}
        \label{fig:mia_acc}
    \end{subfigure}
    \hfill
    \begin{subfigure}[t]{0.48\textwidth}
        \centering
        \includegraphics[width=0.8\linewidth]{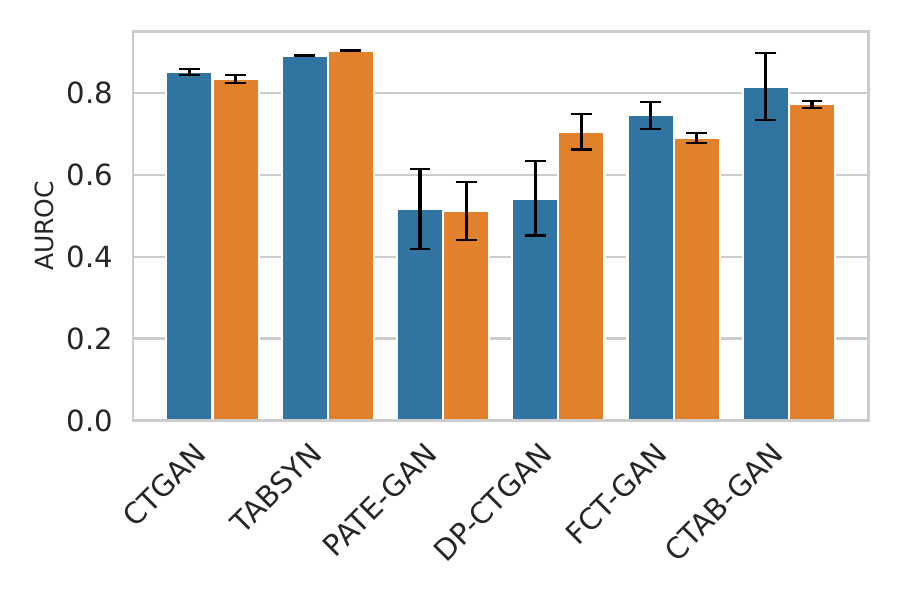}
        \vspace{-0.2cm}
        \caption{Model Inversion Attack - AUROC}
        \label{fig:mia_auroc}
    \end{subfigure}
    \begin{subfigure}[t]{0.48\textwidth}
        \centering
        \includegraphics[width=0.8\linewidth]{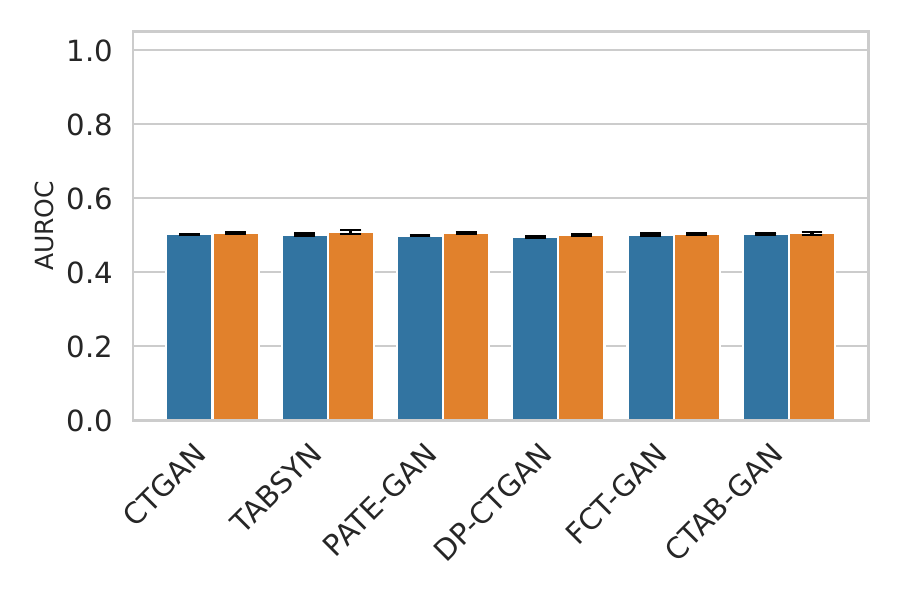}
        \vspace{-0.2cm}
        \caption{Membership Inference Attack - AUROC}
        \label{fig:minf_auroc}
    \end{subfigure}
    \vspace{0.1cm}
    \begin{subfigure}{\textwidth}
        \centering
        \includegraphics[width=0.4\textwidth]{plots/dataset_legend.pdf}
    \end{subfigure}
    \Description{Comparison of Privacy metrics by model and dataset.}
    \vspace{-0.7cm}
    \caption{Comparison of Privacy metrics by model and dataset.}
    \label{fig:pmc}
\end{figure}
\begin{figure}[htpb]
    \centering
    
    \begin{subfigure}[t]{0.5\textwidth}
        \centering
        \includegraphics[width=0.8\linewidth]{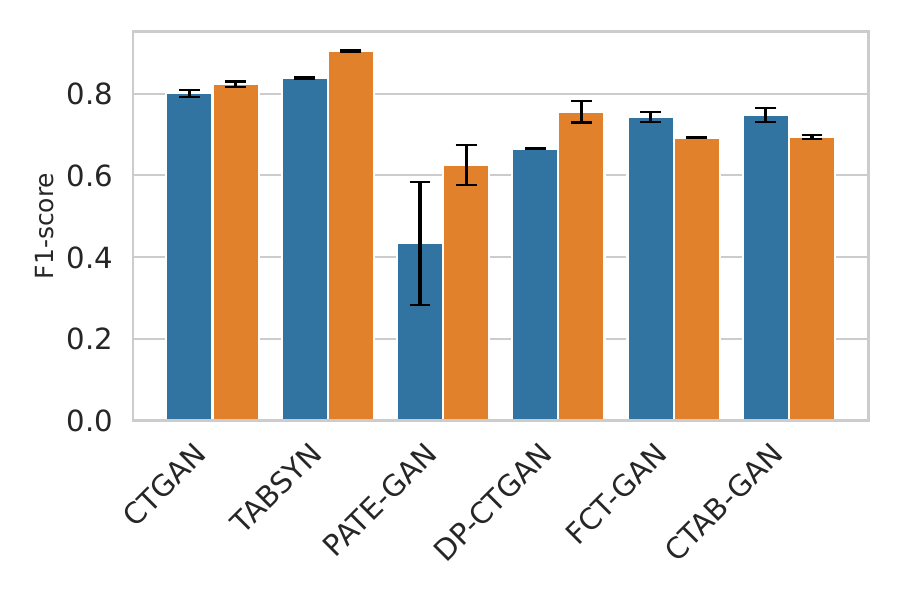}
        \vspace{-0.2cm}
        \caption{F1-Score}
        \label{fig:f1score}
    \end{subfigure}
    \hfill
    \begin{subfigure}[t]{0.49\textwidth}
        \centering
        \includegraphics[width=0.8\linewidth]{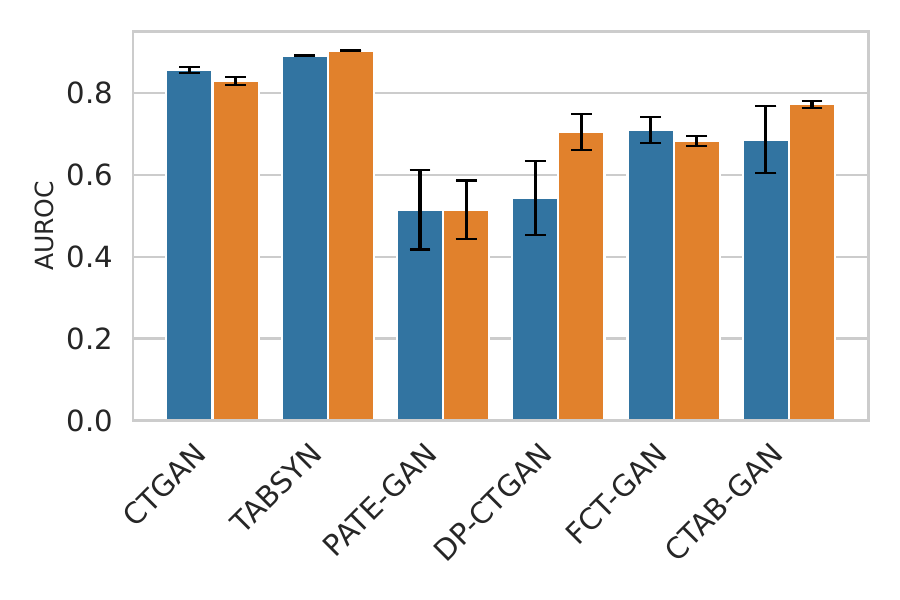}
        \vspace{-0.2cm}
        \caption{AUROC}
        \label{fig:auroc}
    \end{subfigure}
    \begin{subfigure}[t]{0.5\textwidth}
        \centering        \includegraphics[width=0.8\linewidth]{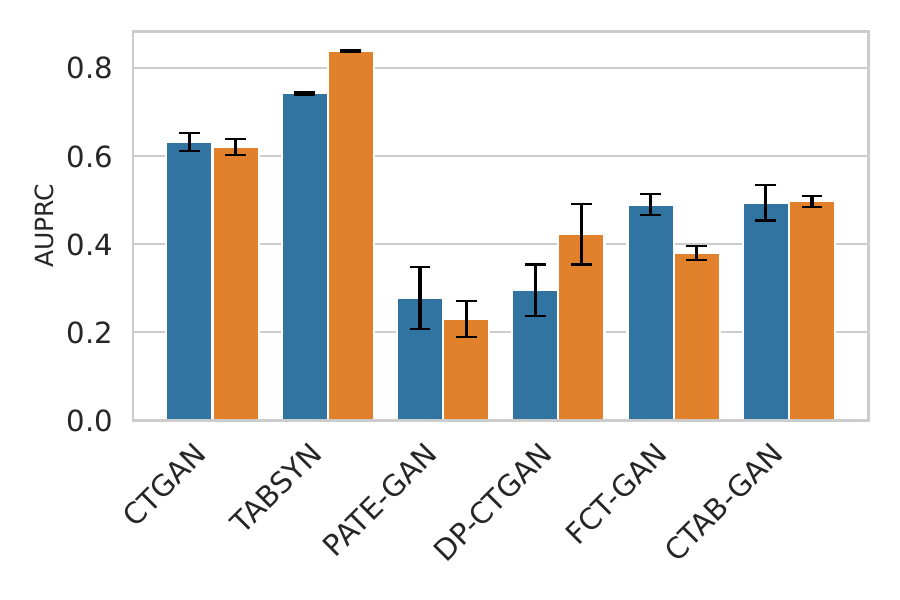}
        \vspace{-0.2cm}
        \caption{AUPRC}
        \label{fig:auprc}
    \end{subfigure}
    \hfill
    \begin{subfigure}[t]{0.49\textwidth}
        \centering
        \includegraphics[width=0.8\linewidth]{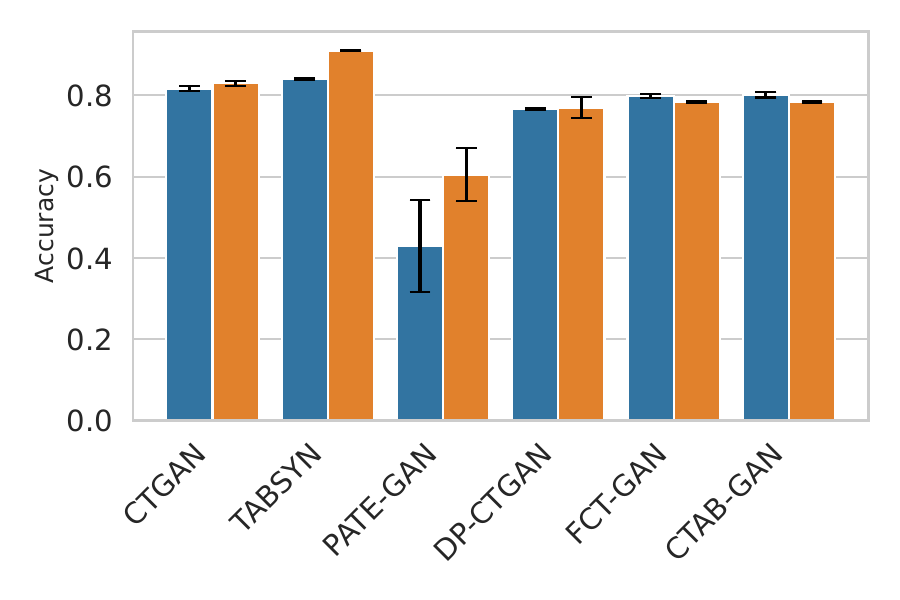}
        \vspace{-0.2cm}
        \caption{Accuracy}
        \label{fig:Accuracy}
    \end{subfigure}
    
    \vspace{0.1cm}
    \begin{subfigure}{\textwidth}
        \centering
        \includegraphics[width=0.4\textwidth]{plots/dataset_legend.pdf}
    \end{subfigure}
    \Description{Machine Learning Utility Metrics by model and dataset.}
    \vspace{-0.7cm}
    \caption{Machine Learning Utility Metrics by model and dataset.}
    \label{fig:mlu}
    \vspace{-0.8em}
\end{figure}

\end{document}